\documentclass{article}

\PassOptionsToPackage{numbers, compress}{natbib}

\usepackage{mmll}
\usepackage{booktabs,tabularx}
\usepackage{caption}

\makeatletter
\newenvironment{breakablealgorithm}
{%
  \par\addvspace{\intextsep}%
  \begingroup
  \renewcommand{\caption}[2][]{%
    \refstepcounter{algorithm}%
    \if\relax\detokenize{##1}\relax
      \addcontentsline{loa}{algorithm}%
        {\protect\numberline{\thealgorithm}##2}%
    \else
      \addcontentsline{loa}{algorithm}%
        {\protect\numberline{\thealgorithm}##1}%
    \fi
    \hrule height .8pt\relax
    \nobreak\kern 2pt
    {\raggedright\interlinepenalty\@M\noindent
      \textbf{\fname@algorithm~\thealgorithm}\space ##2\par}%
    \nobreak\kern 2pt
    \hrule height .4pt\relax
    \nobreak\kern 2pt
  }%
}
{%
  \par\nobreak\kern 2pt
  \hrule height .4pt\relax
  \endgroup
  \par\addvspace{\intextsep}%
}
\makeatother

 \usepackage[main, final]{neurips_2026}

\usepackage[utf8]{inputenc} % allow utf-8 input
\usepackage[T1]{fontenc}    % use 8-bit T1 fonts
\usepackage{url}            % simple URL typesetting
\usepackage{booktabs}       % professional-quality tables
\usepackage{amsfonts}       % blackboard math symbols
\usepackage{nicefrac}       % compact symbols for 1/2, etc.
\usepackage{microtype}      % microtypography
\usepackage{xcolor}         % colors
\usepackage{wrapfig}
\title{High Probability Risk Control for \\Online Policy Learning}

\author{%
  \textbf{Yihong Guo,\quad Drew Prinster,\quad  Suchi Saria,\quad Anqi Liu}\\[3pt]
  Johns Hopkins University\\[3pt]
  {\small
    \texttt{\{yguo80,aprinst1,ssaria1,aliu.cs\}@jhu.edu}
  }
}

\hypersetup{
  pdftitle={High Probability Risk Control for Online Policy Learning},
  pdfauthor={Yihong Guo, Drew Prinster, Suchi Saria, Anqi Liu}
}

\makeatletter
\providecommand*{\theHALG@line}{}
\renewcommand*{\theHALG@line}{\thealgorithm.\arabic{ALG@line}}
\makeatother

\begin{document}

\maketitle

\begin{abstract}
We study risk control for online policy learning, where an agent adaptively collects data and aims to learn a new policy whose risk stays below a tolerance threshold at every round with high probability. Such policy updates and data collection induce feedback covariate shift (FCS): observed data depend on history through evolving policies and the data-collection process. FCS brings two substantive problems for risk control: 1) efficient risk evaluation for each candidate policy and 2) valid policy selection with high-probability risk control. To address these, we propose a permutation-based weighted risk evaluator with weights efficiently approximated by the Sinkhorn algorithm and select policies whose estimated risk upper confidence bounds (UCB) fall below the target threshold, with the UCB obtained through a proposed Gaussian bootstrap method.  Theoretically, we prove that our risk evaluator has a no-larger variance upper bound than that of the standard importance-weighted evaluator. Our policy control method also yields an asymptotically valid simultaneous risk bound and provides asymptotic high-probability risk control for the selected policy. Empirically, we evaluate our method on the online contextual bandit problem, safe question answering benchmark, and Medical Claim Filtering tasks. Our method maintains the target risk level, achieves lower risk than baseline methods, has lower variance in risk estimation, and achieves reliable policy selection with high-probability risk control. Further ablation studies support our theoretical claims.
\end{abstract}

\section{Introduction}

Machine learning (ML) / artificial intelligence (AI) systems can be applied to sequential decision-making in high-stakes settings such as healthcare \citep{habehh2021machine} and autonomous driving \citep{guo2026correctionplanner}, but deployed models should adhere to constraints such as safety, fairness, and the law. In other words, responsible model deployment often requires controlling the risk of such constraint violations.

We study risk control for online policy learning with adaptive data collection and the policy learning process. Across sequential rounds of deployment, the goal is to repeatedly select a learned policy (i.e., a distribution over actions) to deploy whose risk is below a prescribed threshold. 
At each round, the agent updates a policy using historical data, calibrates its risk with tuning parameters, deploys the calibrated policy, and then incorporates the newly collected data into the next update. 

While \citet{prinster2026conformal} developed a framework for this policy control problem (building on ideas from conformal risk control \citep{angelopoulos2024conformal}) with guarantees that hold \textit{in expectation} over the risk of constraint violations, here we focus on a stronger desideratum of risk control with an (asymptotic) \textit{high-probability} guarantee: the risk of the selected deployment policy is below a tolerance threshold with high probability. 
This introduces at least two key challenges for high-probability risk control: (1) efficient risk \textit{evaluation} for each candidate policy and (2) policy \textit{selection} while maintaining valid high-probability guarantees. That is, at each round, one requires not only a valid estimate of each candidate policy's risk, but also a valid procedure for choosing a policy to deploy without selection bias corrupting the decision \citep{prinster2026conformal}. Our paper focuses on the two aspects of high-probability risk control.

Efficient risk evaluation is difficult in online policy learning because adaptive data collection induces feedback covariate shift \citep{fannjiang2022conformal, prinster2024conformal}: the policy and the observed data distribution change over time as a function of the history during the online learning process. Under FCS, standard importance-weighted approaches
may be invalid because they do not account for sequential dependencies; moreover, such estimators can suffer from high variance when policies diverge and encounter extreme values, which is common when the policy changes quickly. To address this, we extend the permutation-based weighted method for CP in \citet{prinster2024conformal}  to risk evaluation under FCS and theoretically show that the proposed evaluator has no greater variance upper bound than the standard importance-weighted evaluator. However, such permutation marginal weights are intractable (factorial complexity). Therefore, we formulate permutation-based weight computation as a variational solution to a constrained optimization problem that can be solved efficiently using the Sinkhorn algorithm \citep{sinkhorn1967concerning}. 

A second challenge is high-probability risk control. 
While existing works such as conformal policy control (CPC) \citep{prinster2026conformal} or CRC \citep{angelopoulos2024conformal} only guarantee that the expected risk is below the threshold,
we propose a Gaussian multiplier bootstrap for policy control on top of the risk estimator to achieve an asymptotic high-probability guarantee, which provides an empirical risk upper bound and does not require i.i.d. data. Specifically, our method estimates the worst-case inflation of the risk estimate over all $\lambda$, yielding a simultaneous upper bound over the candidate set. Selecting policies with UCBs below the threshold ensures that all selected policies satisfy the risk threshold with asymptotic high probability.

\section{Preliminaries and Related Work}
\label{sec:setup}

\subsection{Online Policy Learning and Risk Control}
\label{sec:online-policy-calibration}

At each round $t$, a learner observes a context $X_t\in\mathcal X$,
selects an action $A_t\in\mathcal A$, and receives a reward
$R_t\in\mathbb R$ and a bounded loss $L_t\in[0,B]$.
Let $Z_t=(X_t,A_t,R_t,L_t)$ and
$Z_{1:t}=(Z_1,\ldots,Z_t)$ denote the observation and ordered history.
Using this history, the learner updates a base policy
$\pi_\theta^{t+1}$.
Rather than deploying the policy directly, we select from a
candidate family
$\{\pi_{\theta,\lambda}^{t+1}:\lambda\in\Lambda\}$,
where $\lambda$ adjusts the policy's action distribution, such as temperature, and thereby
its reward--risk trade-off.
The goal is to select
$\widehat\lambda_{t+1}$ such that
$ 
    \ell_{t+1}(\widehat\lambda_{t+1})
    :=
    \mathbb E_{\pi_{\theta,\widehat\lambda_{t+1}}^{t+1}}
    \left[
        L_{t+1}(X_{t+1},A_{t+1})\mid Z_{1:t},X_{t+1}
    \right]
    \leq \alpha,
$
where $\alpha$ is the risk tolerance.
We assume that the candidate set \(\Lambda\) is finite. We also assume access to a known fallback policy \(\pi_{\mathrm{safe}}\) whose deployment risk is at most \(\alpha\). 
A marginal guarantee controls expected risk below the tolerance threshold at every round:  $\mathbb E[\ell_{t+1}(\widehat\lambda_{t+1})] \leq \alpha$.
The high-probability objective instead controls the probability of
exceeding the tolerance every round:
$\mathbb P(
        \ell_{t+1}(\widehat\lambda_{t+1})>\alpha
    )
\leq \delta,$ where $\delta\in(0,1)$ is the allowed failure probability.
\subsection{Feedback Covariate Shift and Conformal Risk/Policy Control}
\label{sec:online-crc-baseline}

\textbf{Feedback Covariate Shift.}
Online policy updates create a challenge for risk control at every round:
The policy deployed at each round depends on past observations,
while also determining the distribution of future collected
data. This adaptive data-collection process induces feedback covariate shift and produces dependent, generally non-exchangeable observations.
It differs from non-adaptive covariate shift or non-adaptive policy shift, where the sampling
policies are independent of the observed data, and there is no such sequential dependency among data observations.
Building on weighted conformal prediction under fixed covariate
shift \citep{tibshirani2019conformal}, subsequent work studies
single-round \citep{fannjiang2022conformal} and multi-round FCS
\citep{nair2023randomization,prinster2024conformal}. Sample-wise importance ratios account for differences between
behavior and target policies, but do not explicitly account for such sequential data dependency and feedback covariate shift. 
Permutation-based weighting accounts for the
joint sampling mechanism, including its dependence on the
observation history, but direct computation becomes intractable
as the history grows \cite{prinster2024conformal}.

\textbf{Conformal Risk Control.}
Conformal risk control (CRC)
\citep{angelopoulos2024conformal} extends conformal prediction
\citep{vovk2005algorithmic} 
to control the expectation of any bounded loss function $L_i(\lambda)$ that is monotonically parameterized by some tuning parameter, $\lambda$.
For example, CRC recovers conformal prediction for the miscoverage loss $L_i(\lambda):=\mathbf{1}\{Y_i\not\in\hat{C}_{\lambda}(X_i)\}$, where $\lambda$ controls the size of a prediction set $\hat{C}_{\lambda}(X_i)$ to reduce the risk of not including $Y_i$.
Or, $\lambda$ could be a classification threshold that monotonically reduces the false negative rate.
Related approaches, including \citet{bates2021distribution}
and learn-then-test (LTT) \citep{angelopoulos2021learn},
provide high-probability risk control for non-monotonic losses under i.i.d.\ data.
However, these guarantees do not directly apply to the adaptive
policy learning setting with feedback covariate shift.

\textbf{Conformal Policy Control (CPC).}
CPC \citep{prinster2026conformal} is most related to our work.
Rather than assuming a loss $L_i(\lambda)$ that can be directly tuned by a parameter $\lambda$, 
CPC accommodates losses $L_i$ that admit no tuning parameter at all, by instead having $\lambda$ index a family of \textit{policies}, $\{\pi_{\theta,\lambda}^{t+1}:\lambda\in\Lambda\}$.
For example, the constraint violation indicator loss $L_i:=\mathbf{1}\{A_i\not\in \mathcal{S}\}$, where $\mathcal{S}$ is a set of safe or allowable actions, is not a function of any tuning parameter $\lambda$; CPC controls the expectation of such losses by instead tuning the \textit{action distribution} that the expectation is taken over.
Its conservative
selection rule,
$\widehat\lambda_{\mathrm{CPC}}
=
\inf\{
\lambda_0\in\Lambda:
\sup_{\substack{\lambda\in\Lambda,\lambda\geq\lambda_0}}
[
\sum_{i=1}^{t}p_i(\lambda)L_i
+
p_{\max}(\lambda)B
]
\leq\alpha
\},$
handles non-monotonicity in the policy search, 
where $p_i(\lambda)$ are normalized permutation weights that account for the FCS (to be introduced in Section \ref{sec:perm_eval})
and $p_{\max}(\lambda)$ is the worst-case test-point weight
used in the correction. With exact permutation weights, CPC guarantees risk control in expectation: $\mathbb E[\ell_{t+1}(\widehat\lambda_{t+1})]\leq \alpha$.

\textbf{Addressing Limitations of CPC: Efficient Weight Estimation and High-Probability Risk Control.} CPC, however, still has at least two limitations, which our paper aims to address: First, CPC's guarantee only applies for exact permutation weights, and \citet{prinster2026conformal} do not theoretically characterize any relation of their practical estimates (based on mixture weights) to the exact permutation weights. Our first contribution addresses this gap by formulating permutation-weight computation as a constrained optimization problem that can be efficiently estimated via the Sinkhorn algorithm, and shows that the resulting risk evaluator has no greater variance upper bound than the standard importance-weight method. Second, the marginal expectation form of CPC's guarantee does not necessarily guarantee risk control for a particular calibration set, as in the desideratum, due to potentially high variance. To address this limitation, we focus on obtaining an asymptotic \textit{high probability} guarantee, which says: $\limsup_{t\to\infty}
\mathbb P\!(
\ell_{t+1}(\widehat\lambda_{t+1})>\alpha
)
\le \delta$. Note that we mainly focus on the adaptive policy learning setting as in CPC, but our risk control component can apply to non-adaptive covariate shift settings.

Additional related work is provided in
Appendix~\ref{app:related_work}.

\section{Method}
\label{sec:overview}

In this section, we introduce our main algorithm. We first introduce a permutation-based weighted risk estimator to evaluate next-round risk that accounts for FCS, extending the permutation-based weighted method for CP in \cite{prinster2024conformal} to risk estimation. To reduce the computational cost of computing permutation weights exactly, we introduce an efficient estimation method with a constrained optimization problem. With this efficient risk evaluator, our method, \textbf{R}isk \textbf{C}ontrol for Policy \textbf{L}earning in \textbf{E}xpectation (\textbf{RCL-E}), can achieve risk control in expectation under a monotonic risk function. Then, we introduce our policy selection method through a Gaussian multiplier bootstrap method to obtain an asymptotic high-probability risk control guarantee, namely \textbf{R}isk \textbf{C}ontrol for Policy \textbf{L}earning with \textbf{H}igh \textbf{P}robability (\textbf{RCL-HP}). We summarize our algorithm in Algorithm \ref{alg:perm_crc_short}.

\subsection{Efficient Evaluation of Candidate Policy Risks}
\label{sec:perm_eval}

\subsubsection{Preliminaries for Permutation-Based Risk Estimation under Feedback Covariate Shift}

Adapting the key idea from \citet{prinster2024conformal} to our setting, we account for FCS by analyzing the risk conditionally on the \textit{bag} (or multiset) of calibration \textit{and test data}, $E_z:=\{Z_{1:t+1}\}$, and drawing permutations from the bag with probabilities given by 
the sequence of historical policies, $\pi_{\theta, \lambda}^{1:t+1}:=(\pi_{\theta, \lambda_1}^1, ..., \pi_{\theta, \lambda_{t+1}}^{t+1})$, letting $\lambda=\lambda_{t+1}$. 
Importantly, note that the bag $E_z$ conveys the information that the random variable $L_{t+1}$ will take a value from the observed values $\{l_{1:t+1}\}$, but we do not know \textit{which} value (see \citet{prinster2024conformal} for details). The bag-conditional risk is then
\begin{align}
\textstyle
\label{eq:bag_conditional_risk}
    \ell(\lambda)
:=\mathbb{E}_{\pi_{\theta, \lambda}^{1:t+1}}\big[L_{t+1}\mid E_z\big] = \sum_{i=1}^{t+1}l_{i}\cdot \mathbb{P}_{\pi_{\theta, \lambda}^{1:t+1}}\big(L_{t+1}=l_{i}\mid E_z\big),
\end{align}
where $\mathbb{P}_{\pi_{\theta, \lambda}^{1:t+1}}\big(L_{t+1}=l_{i}\mid E_z\big)$ can be interpreted as the probability of drawing a sequence of values from $E_z$ such that the test loss, $L_{t+1}$, is observed to be the value $l_i$. As described in \citet{prinster2024conformal} (which explains analysis from \citet{tibshirani2019conformal}), these probability weights can be expanded via the law of total probability over permutations, $\sigma:[t+1]\rightarrow [t+1]$, as
\begin{equation}\label{eq:permutation_weights_def}
    p_i(\lambda):=\mathbb{P}_{\pi_{\theta, \lambda}^{1:t+1}}\big(L_{t+1}=l_{i}\mid E_z\big) = \frac{\sum_{\sigma:\sigma(t+1)=i}\pi(z_{\sigma(1)}, ..., z_{\sigma(t+1)})}{\sum_{\sigma}\pi(z_{\sigma(1)}, ..., z_{\sigma(t+1)})},
\end{equation}
where $\pi$ denotes the joint distribution induced by the policies $\pi_{\theta, \lambda}^{1:t+1}$ for more concise notation. 

Together, Eq. \eqref{eq:bag_conditional_risk} and Eq. \eqref{eq:permutation_weights_def} thus clarify how permutation-based weights offer a natural and valid, though computationally intimidating, target for risk evaluation under the candidate policy, $\pi_{\theta, \lambda}^{t+1}$. These weights account for the adaptive data collection process and stand in contrast to standard sample-wise importance weights. Moreover, in our setting we assume that the context distribution is i.i.d., so the context distribution cancels out in  Eq. \eqref{eq:permutation_weights_def} (see Appendix \ref{app:permutation_marginal}) and the marginal permutation weights $p_i(\lambda)$ are a function of the historical context-dependent policies, $\{\pi_{\theta, \lambda}^{i}(a \mid x)\}$. 

As described in \citet{prinster2026conformal}, however, a key difference from the CP setting in \citet{prinster2024conformal} is that here we wish to estimate the risk of the policy $\pi_{\theta, \lambda}^{t+1}$ \textit{before sampling the next action}, $A_{t+1}$. 
Accordingly, we simulate a hypothetical test sample $Z_{t+1}$ and conservatively assume it receives the worst-case loss, $L_{t+1}=B$. That is, we conservatively estimate the expected risk as
\begin{equation}\label{eq:bag_conditional_risk_conservative_est}\textstyle
\widehat\ell(\lambda):= \sum_{i=1}^t l_{i}\cdot p_i(\lambda) + B\cdot p_{t+1}(\lambda). 
\end{equation}

\subsubsection{Efficiently Approximating the Permutation Weight}
\label{sec:sinkhorn}

The exact permutation marginal (Eq. \eqref{eq:permutation_weights_def}) easily becomes intractable as the number of rounds grows. \citet{prinster2024conformal} proposed a surrogate approximation that considers only the permutation of the most recent position assignment rather than the full permutation assignment. Instead, we approximate the permutation marginals via entropy-relaxed variational inference over doubly stochastic matrices, whose rows and columns each sum to 1. This leads to a tractable doubly-stochastic assignment matrix computed via Sinkhorn iterations \citep{sinkhorn1967concerning,linderman2018reparameterizing,mena2019sinkhorn}. 

To efficiently obtain the permutation marginal $p_i(\lambda)$, we use an entropy-regularized marginal approximation. We denote the policy matrix $J\in\mathbb R_+^{(t+1)\times(t+1)}$, where $J_{k,i}:=\pi_k(A_i\mid X_i)$. Each element $J_{k, i}$ represents the density that policy $k$ places on the data point $i$, irrespective of the other data points.
We further let $Q^\star$ be the permutation marginal: each entry $Q^\star_{k, i}$ represents the probability that sample $i$ occupies position $k$ in a permutation drawn from the bag, $E_z$. The following Theorem states that the permutation marginal weight $Q^\star$ is a weight matrix close to the original policy matrix $J$, subject to the permutation constraint that each data point is mapped uniquely to one position.

\begin{theorem}[Sinkhorn approximation of permutation marginal weights]
\label{thm:sinkhorn}
We approximate the permutation marginal weights by the entropy-relaxed surrogate:
\begin{align}
\label{eq: sinkhorn_objective}
Q^\star
=
\arg\min_{Q\in\mathcal B_{t+1}}
\mathrm{KL}(Q\|J),
\qquad
\mathcal B_{t+1}
=
\left\{
Q\in\mathbb R_+^{(t+1)\times(t+1)}:
Q\mathbf 1=\mathbf 1,\;
Q^\top \mathbf 1=\mathbf 1
\right\}.
\end{align}
Then $Q^\star$'s last row provides an
approximation to the permutation marginal weights:
\begin{align*}
\textstyle
p_i(\lambda)\approx Q^\star_{t+1,i},\qquad i=1,\dots,t+1.
\end{align*}
\end{theorem}

\begin{remark}
Theorem~\ref{thm:sinkhorn} provides a tractable surrogate for the exact permutation
marginal. Instead of summing over exponentially many feasible assignments, we relax
the exact assignment to an entropic assignment marginal and solve a convex KL
projection problem. The constraint set $\mathcal B_{t+1}$ preserves the global
one-to-one structure of a permutation, where $Q_{i,j}$ represents the probability of data point $j$ being assigned at position $i$. Because each data point is assigned to one position and each position is assigned one data point, the row and column sums of $Q$ are 1. The optimizer can be computed by matrix
scaling, and its last row $Q^\star_{t+1, i}$ serves as the approximate permutation
weight for evaluating the target policy at the final round. The derivations are in Appendix~\ref{app:sinkhorn_proof}. The optimization problem can be solved efficiently via Sinkhorn/IPFP \citep{sinkhorn1967concerning} through iterative matrix scaling, as described in Appendix~\ref{app: sinkhorn_optimization} and Algorithm~\ref{alg:sinkhorn}. 
\end{remark}
\textbf{Sources of Approximation.} The matrix \(Q^\star\) approximates the permutation marginal because the optimization problem is an entropy-relaxed surrogate of the exact marginal inference objective. The exact objective has an intractable entropy term: the entropy of the permutation. We replace the intractable entropy of the full permutation distribution by an entrywise entropy relaxation:
$H(\sigma)
\approx
H(Q)
=
-\sum_{ij} Q_{ij}\log Q_{ij}$.
In Appendix \ref{app:additional-approximation}, we empirically evaluate the approximation error of the Sinkhorn approximation and show it is small. We also provide a theoretical analysis of how the error affects later risk estimation and guarantees in Appendix \ref{app:How approximation error affects the guarantee}. 

\textbf{Interpretation of the Permutation-based Estimator.} We provide how we can calculate the exact permutation marginal in Appendix~\ref{app:permutation_marginal} following \citet{prinster2024conformal}. We now further interpret the permutation-based risk evaluator in the following propositions.  

\begin{proposition}[Reduction to SNIPS under standard covariate shift]
\label{prop:snips}
Suppose all historical data are collected by a fixed behavior policy $\pi_b$,
and the next-round risk is evaluated under the target policy
$\pi_{\theta,\lambda}^{t+1}$.
Then the exact permutation marginal weight under the fixed policy sequence
reduces to the Self-Normalized Importance Sampling (SNIPS) weight:
$p_i(\lambda)
=
\frac{
\pi_{\theta,\lambda}^{t+1}(A_i\mid X_i)
/\pi_b(A_i\mid X_i)
}{
\sum_{j=1}^{t+1}
\pi_{\theta,\lambda}^{t+1}(A_j\mid X_j)
/\pi_b(A_j\mid X_j)
}$.
\end{proposition}

\begin{proposition}[Permutation weight as a global importance weight]
\label{prop:global_iw}
Under the fixed policy sequence
$\pi_{\theta,\lambda}^{1},\ldots,\pi_{\theta,\lambda}^{t+1}$, the exact permutation marginal weight can be written as
$p_i(\lambda)
=
\frac{
\pi_{\theta,\lambda}^{t+1}(A_i\mid X_i)\,C_{i,t+1}
}{
\sum_{k=1}^{t+1}
\pi_{\theta,\lambda}^{k}(A_i\mid X_i)\,C_{i,k}
}$,
where $C_{i,k}:= \sum_{\sigma:\,\sigma(k)=i}
\prod_{\substack{h=1, h\neq k}}^{t+1}
\pi_{\theta,\lambda}^{h}
(A_{\sigma(h)}\mid X_{\sigma(h)})$ denotes the total contribution of all feasible assignments
in which sample $i$ occupies position $k$.
\end{proposition}

\begin{remark}
Proposition~\ref{prop:snips} shows that the permutation weights reduce to the SNIPS in the covariate shift setting, and Proposition~\ref{prop:global_iw} shows their connection to importance sampling under a fixed policy sequence. Classic sample-wise importance weights are local, depending only on the ratio between the candidate policy and the behavior policy for the realized sample. In contrast, permutation weights are global, normalized weights: each weight depends not only on the corresponding policy likelihood but also on the entire observed data bag and all feasible assignments. This is what allows the estimator to account for the sequential dependence and potentially provide a smoother importance weight.
\end{remark}

\subsection{Bootstrap Policy Control for Asymptotic High-Probability Guarantees}
\label{sec:selection}

\textbf{Risk Control in Expectation under Monotonicity.} With our efficient risk evaluator, we propose a CPC/CRC-style risk control when the risk is monotonic. For the selected $\hat{\lambda} = \inf\{\lambda:\widehat \ell(\lambda)\le\alpha\}$, our method \textbf{RCL-E} has: 
   $ \mathbb{E}[\ell(\hat{\lambda})] \leq \alpha$,
where the risk is controlled in expectation. Under monotonicity, \textbf{RCL-E} can be viewed as CPC \citep{prinster2026conformal} with our efficient Sinkhorn-based weight estimation.

\textbf{High Probability Risk Control.}  However, when the risk function is non-monotonic, or we want to achieve high-probability risk control, selecting $\lambda$ through $\hat{\lambda} = \inf\{\lambda:\widehat \ell(\lambda)\le\alpha\}$ does not provide the desired guarantee. We now introduce our policy selection method for high-probability risk control, namely \textbf{RCL-HP}. To guarantee high-probability risk control after this policy selection step, it is not enough to ensure high-probability risk control of each fixed candidate separately. Instead, we need to ensure that no unsafe candidate enters the estimated feasible set:
$
\mathbb P(
\exists \lambda\in \widehat\Lambda_{\mathrm{feas}}
:\ell(\lambda)>\alpha
)
\le \delta$.

We construct a \emph{simultaneous} uncertainty correction over the entire candidate set $\Lambda$. The correction uses the largest risk underestimation error,
$T
=
\sup_{\lambda\in\Lambda}
\{
\ell(\lambda)-\widehat \ell(\lambda)
\}$.
Let $q_{1-\delta}$ be the $(1-\delta)$-quantile of T. Then $\ell(\lambda)+q_{1-\delta}$ is a simultaneous upper confidence bound over $\Lambda$. Define the feasible set as:
$\widehat\Lambda_{\mathrm{feas}}
=
\{\lambda\in\Lambda:\widehat\ell(\lambda) + q_{1-\delta} \le \alpha\}$.

\begin{theorem}[High probability risk control with true $q_{1-\delta}$]
\label{thm:fwer}
With $q_{1-\delta}$ as the true value of the $(1-\delta)$-quantile of $T$, and $\widehat\Lambda_{\mathrm{feas}}
=
\{\lambda\in\Lambda:\widehat\ell(\lambda) + q_{1-\delta} \le \alpha\}$, we have $\mathbb{P}(
\exists \lambda\in \widehat\Lambda_{\mathrm{feas}}
:\ell(\lambda)>\alpha
)
\le \delta$.
\end{theorem}

\begin{remark}
Theorem~\ref{thm:fwer} shows that if the true quantile $q_{1-\delta}$ were available, then the policy selection rule achieves high-probability risk control. The $q_{1-\delta}$ acts as a uniform safety margin that accounts for the worst underestimation error over the entire candidate family, rather than the error of a single fixed candidate. This lets us control the policy and select a policy from the candidate set regardless of the non-monotonic loss. Consequently, the policy selected from $\widehat\Lambda_{\mathrm{feas}}$ satisfies the risk constraint with high probability at least $1-\delta$. The proof is given in Appendix D. This result assumes access to the true $(1-\delta)$-quantile of the maximal underestimation error. In practice, this quantile is unknown. We approximate its distribution using a Gaussian multiplier bootstrap, establish bootstrap consistency in Theorem \ref{thm:main_boot}, and obtain the asymptotic guarantee in Corollary \ref{cor:main_fwer}.
\end{remark}

\textbf{Gaussian Multiplier Bootstrap to Approximate $q_{1-\delta}$}.
A direct resampling bootstrap is inappropriate because the data are collected adaptively rather than i.i.d., and resampling raw observations would distort the sequential data-collection mechanism.
Instead, we apply a Gaussian multiplier bootstrap \citep{chernozhukov2013gaussian}.
Let $\widehat \ell(\lambda)$ denote the base risk estimator, and let $\widehat\phi_i(\lambda)$ denote the contribution of sample $Z_i$ to the linearized estimation error at parameter $\lambda$:
    $\textstyle
    \widehat\phi_i(\lambda)
=
p_i(\lambda)\bigl(L_i-\widehat \ell(\lambda)\bigr)$.
Given independent multipliers $\zeta_i\overset{\mathrm{i.i.d.}}{\sim}N(0,1)$, we define the bootstrap process
\begin{align*}
\textstyle
\widehat G_{t+1}^*(\lambda)
=
\sum_{i=1}^{t+1}
\zeta_i\,\widehat\phi_i(\lambda).
\end{align*}

For each bootstrap draw, we compute the maximal fluctuation
$T^*
=
\sup_{\lambda\in\Lambda}
\widehat G_{t+1}^*(\lambda),
$
and repeat this procedure to obtain an empirical approximation to the distribution of the maximal risk-estimation error. Then, using this distribution, we can obtain the $1-\delta$ quantile of $T$ and select a policy. However, since we approximate it with high probability, we can only obtain an asymptotic high-probability risk control with bootstrap consistency rather than a finite-sample guarantee. 

\section{Theoretical Analysis}
\label{sec:theory_main}
In Theorem~\ref{thm:varenv-comparison}, we show that our method achieves a variance upper bound no larger than that of the standard importance weight estimator. In Theorem~\ref {thm:main_boot} and Corollary~\ref{cor:main_fwer}, we show the asymptotic high-probability risk control of our policy selection method. 

\begin{proposition}[Weight upper bounds]
\label{prop:max_weight_compare}
Let $n=t+1$, $m=|\mathcal A|<\infty$, and
$M_\epsilon=[1-(m-1)\epsilon]/\epsilon$.
Assume all historical policies $\pi_{\theta,\lambda_i}^{i}$ and
the candidate $\pi_{\theta,\lambda}^{t+1}$ satisfy
$\pi(a\mid x)\ge\epsilon$ for all $(x,a)$, with
$0<\epsilon\le1/m$.
For the positive, normalized permutation weights $p$, assume
$\Delta_p:=\log(\max_i p_i/\min_i p_i)\le
\Delta_J:=\log(\max_{k,i}J_{k,i}/\min_{k,i}J_{k,i})$, i,e., the log range of the weight matrix does not increase after the permutation.
For the importance weight, set $\omega_i=\pi_{\theta,\lambda}^{t+1}(A_i\mid X_i)/
\pi_{\theta,\lambda_i}^{i}(A_i\mid X_i)$ for $i\le t$, and
$\omega_n=1$ for the candidate policy. Then
\begin{align}
 \max_i p_i
 \le \frac{M_\epsilon}{M_\epsilon+n-1}
 =:a_{\mathrm{perm}},
 \label{eq:varenv-perm-coeff} \qquad
 \max_i\frac{\omega_i}{n}
 \le \frac{M_\epsilon}{n}
 =:a_{\mathrm{IPS}}.
 \end{align}
Hence $a_{\mathrm{perm}}\le a_{\mathrm{IPS}}$.
\end{proposition}

\begin{remark}
A smaller log-range $\Delta_p\le\Delta_J$ indicates more uniform weight.
Under $\Delta_p\le\Delta_J$,
Proposition~\ref{prop:max_weight_compare} gives a no-larger upper bound for permutation weight than
for IPS, namely $a_{\mathrm{perm}}\le a_{\mathrm{IPS}}$.
We empirically show that $\Delta_Q^{(r)} \le \Delta_J$ likely holds in Appendix \ref{app:additional-log-range}, where $\Delta_Q^{(r)}$ is the log range of the last row of $Q$.
\end{remark}
\begin{theorem}[Variance upper bounds]
\label{thm:varenv-comparison}
Under Proposition~\ref{prop:max_weight_compare}, for the bounded losses, put $\widetilde L_i=L_i$ for $i\le t$
and $\widetilde L_n=B$.
Assume
$\operatorname{Var}(\widetilde L_i)\le\sigma_L^2$
for all $i$ and
$|\operatorname{Cov}(\widetilde L_i,\widetilde L_j)|
\le C_{\mathrm{mix}}\sigma_L^2\rho^{|i-j|}$ for $i\ne j$,
where $C_{\mathrm{mix}}>0$ and $0<\rho<1$.
Let $\widehat\ell_{\mathrm{perm}}=\sum_{i=1}^n p_i\widetilde L_i$,
$\widehat\ell_{\mathrm{IPS}}=n^{-1}\sum_{i=1}^n\omega_i\widetilde L_i$,
and $\Gamma_n=\sigma_L^2[n+2C_{\mathrm{mix}}
\sum_{k=1}^{n-1}(n-k)\rho^k]$.
Then,
\begin{equation}
\label{eq:perm_smaller_var}
\begin{aligned}
 \operatorname{Var}(\widehat\ell_{\mathrm{perm}})
 &\le U_{\mathrm{perm}}:=\Gamma_n a_{\mathrm{perm}}^2, \qquad
 \operatorname{Var}(\widehat\ell_{\mathrm{IPS}})
 &\le U_{\mathrm{IPS}}:=\Gamma_n a_{\mathrm{IPS}}^2.
\end{aligned}
\end{equation}
Moreover, $U_{\mathrm{perm}}\le U_{\mathrm{IPS}}$.
\end{theorem}
\begin{remark}
The variance decomposition gives
$\operatorname{Var}(\widehat\ell)
\le\Gamma_n c_{\max}^2$,
where $c_{\max}=\max_i c_i$ and
$c_i=p_i$ for permutation weighting or $c_i=\omega_i/n$ for IPS.
Thus, the weight bounds control variance and
covariance contributions to the upper bound.
Proposition~\ref{prop:max_weight_compare} yields
$a_{\mathrm{perm}}\le a_{\mathrm{IPS}}$, implying
$U_{\mathrm{perm}}\le U_{\mathrm{IPS}}$. 
Proofs of Proposition~\ref{prop:max_weight_compare} and
Theorem~\ref{thm:varenv-comparison} are given in
Appendices~\ref{app:proof_of_smalle_weight}
and~\ref{app:proof_perm_smaller_var}, respectively.
We separately report empirical variance comparisons
with SNIPS in Section~\ref{sec:exp}.
\end{remark}

\begin{theorem}[Bootstrap validity in high probability risk control]
\label{thm:main_boot}
Under Assumption~\ref{ass:contextual-primitive},
consider a fixed finite candidate family $\Lambda_{t+1}$
and define
\begin{align*}
\textstyle
     T_{t+1}
 =\sup_{\lambda\in\Lambda_{t+1}}
 \bigl\{\ell_{t+1}(\lambda)
       -\widehat\ell_{t+1}(\lambda)\bigr\}.
\end{align*}

For each $\lambda_j\in\Lambda_{t+1}$, define
\begin{align*}
\textstyle
     \widehat\phi_i(\lambda_j)
 =p_i(\lambda_j)
  \bigl\{\widetilde L_i-\widehat\ell_{t+1}(\lambda_j)\bigr\},
 \qquad
 \widehat G_{t+1}^{*}(\lambda_j)
 =\sum_{i=1}^{t+1}\zeta_i\widehat\phi_i(\lambda_j),
\end{align*}
where $\widetilde L_i=L_i$ for $i\le t$,
$\widetilde L_{t+1}=B$ for the test point, and
$\zeta_i\overset{\mathrm{i.i.d.}}{\sim}N(0,1)$ are
independent of the data and shared across candidates.
Define
\begin{align*}
\textstyle
     T_{t+1}^{*}
 =\sup_{\lambda\in\Lambda_{t+1}}
   \widehat G_{t+1}^{*}(\lambda).
\end{align*}

For fixed $\delta\in(0,1)$, let $q_{1-\delta,t+1}^{*}$
be the $(1-\delta)$-quantile of $T_{t+1}^{*}$
given $\mathcal D_t$.
Then, as $t\to\infty$,
\begin{align*}
\textstyle
 \sup_{u\in\mathbb R}
 \left|
  \mathbb P(T_{t+1}\le u)
  -\mathbb P(T_{t+1}^{*}\le u\mid\mathcal D_t)
 \right|
 \xrightarrow{p}0.
\end{align*}
Moreover,
\[
 \mathbb P\!\left(
  \ell_{t+1}(\lambda)
  \le \widehat\ell_{t+1}(\lambda)+q_{1-\delta,t+1}^{*},
  \ \text{for all }\lambda\in\Lambda_{t+1}
 \right)
 \longrightarrow 1-\delta.
\]
\end{theorem}

\begin{remark}
Theorem~\ref{thm:main_boot} justifies that the bootstrap
approximation of the error can asymptotically represent the true underestimation error of the risk estimate. So we can obtain a valid $q_{1-\delta}$ quantile for the error, leading to the asymptotically high-probability guarantee in the following Corollary \ref{cor:main_fwer}. 
The assumption includes a fixed finite family of
candidate policies, uniformly positive action probabilities,
and adaptive logging policies satisfying the finite-dimensional
affine structure and stabilization conditions
in Assumption \ref{ass:contextual-primitive}.
\end{remark}
\noindent\textbf{Proof Sketch:} We first derive a first-order linear expansion of the estimation error under Assumption~\ref{ass:contextual-primitive} in terms of the policy sequence. Then, we show that for each candidate policy, the ground-truth estimation error and the bootstrap-estimated error converge asymptotically to the same Gaussian distribution. Then, the ground-truth and the estimated maximum error among all candidate policies follow the same distribution, and the bootstrap quantile is a valid approximation of the true error. The complete proof is given in Appendix \ref{app:proof of main thm}. Appendix \ref{app: Empirical Evaluation of the Assumption in Theory} provides empirical and theoretical diagnostics and verification of the Lemma and Assumptions.

\begin{corollary}[Asymptotic High Probability Risk Control]
\label{cor:main_fwer}
Under the assumptions of Theorem~\ref{thm:main_boot}, define the feasible set
$
\widehat{\Lambda}_{t+1,\mathrm{feas}}
:=
\{
\lambda\in\Lambda_{t+1}:
\widehat \ell_{t+1}(\lambda)+q_{1-\delta,t+1}^*\le \alpha
\}.
$
If the selected parameter \(\widehat\lambda_{t+1}\) is restricted to
\(\widehat{\Lambda}_{t+1,\mathrm{feas}}\) or fallback to default safe policy $\pi_\text{safe}$, then
\begin{align*}
\textstyle
\limsup_{t\to\infty}
\mathbb P\!(
\ell_{t+1}(\widehat\lambda_{t+1})>\alpha
)
\le \delta.
\end{align*}
Moreover,
$\liminf_{t\to\infty}
\mathbb P\!(
\ell_{t+1}(\lambda)\le \alpha
\text{ for all } \lambda\in\widehat{\Lambda}_{t+1,\mathrm{feas}}
)
\ge 1-\delta$.
\end{corollary}

\begin{remark}
Corollary~\ref{cor:main_fwer} converts the simultaneous bootstrap UCB into a high-probability
guarantee: after screening by the corrected risk estimate, any chosen
$\widehat\lambda_{t+1}$ is asymptotically safe up to error probability $\delta$.
The final statement corresponds to controlling the family-wise error of
selecting at least one candidate that violates any risk constraint.
\end{remark}

\section{Experiments}
\label{sec:exp}
In this section, we evaluate the empirical risk control performance on different online learning applications in Section \ref{sec:mainresults} and  \ref{sec:medicalQA}. We show the benefit of our permutation-based weighted risk evaluator in Section \ref{sec:riskestimatecomp}. We compare with high-probability control methods and show that it effectively accounts for feedback covariate shift while maintaining high-probability risk control without being overly conservative in Appendix \ref{sec:highprobcomp}.  We also provide additional results on 
permutation weight approximation error and bootstrap validity diagnostics in Appendices~\ref{app:additional-approximation}, \ref{app: Empirical Evaluation of the Assumption in Theory}, respectively. The code is available at:
\url{https://github.com/guoyihonggyh/High-Probability-Risk-Control}

\subsection{Risk Control in Expectation and High Probability Risk Control} 
\label{sec:mainresults}
We briefly introduce the baselines, datasets, tasks, risk definition, as well as the policy and update rules. Appendix \ref{app:baseline} details the baselines, and Appendix~\ref{app: dataset} details the task settings. 

\textbf{Baselines:} 
Online CRC (\textbf{O-CRC}) \citep{angelopoulos2024conformal} extends CRC under the distribution shift setting to online learning and uses importance weighting to account for the policy shift. \textbf{RollingRC} \citep{feldman2022achieving} utilizes the adaptive conformal inference (ACI) \cite{gibbs2021adaptive} to adapt to the distribution shift. Conformal Policy Control (\textbf{CPC}) \citep{prinster2026conformal} uses a similar permutation-weighting idea as ours but provides a control method for a non-monotonic risk function. Note that CPC in our implementation exactly matches RCL-E in the monotonic risk-function setting and uses its proposed control method in the non-monotonic risk-function setting. We also compare against the permutation weight estimation method originally proposed in \cite{prinster2024conformal}. Details are in Appendix~\ref{app:baseline}. We do not compare against Learn-then-Test (LTT) here because it requires i.i.d. data, but we discuss and compare it with our RCL-HP in a later section. 

\textbf{Datasets and Tasks:} We conduct experiments on two online contextual bandit tasks, breast cancer diagnosis \cite{street1993nuclear} and stock investment \cite{thomas2023yahoo}, and one LLM safe question answering task with SafetyBench \cite{zhang2024safetybench}. For breast cancer diagnosis, the policy controls the false negative rate of breast cancer diagnosis. For stock investment, the policy controls the probability of selecting the worst action below the threshold.  For SafetyBench Question Answering, the policy controls the error rate by refusing to answer questions. 
The breast cancer diagnosis is a monotonic risk setting; the remaining two tasks are non-monotonic risk settings. We summarize how the $\lambda$ affects the policy, which risk is controlled, and whether the risk function is monotonic in Table
\ref{tab:experiment_metrics}. 

\textbf{Data Collection and Policy Update:}
We initialize with $m_0$ offline samples split into training and calibration sets. Each round collects $m$ new samples, splits them equally between the two sets, and updates the policy using training data until the data budget is reached. We set $(m_0,m)=(50,20)$ for Breast Cancer Diagnosis and $(300,50)$ for Stock Investment.
For safe question answering, we use SFT checkpoints from the base model as a fixed policy sequence to induce larger policy changes and potentially stronger covariate shift. Figure~\ref{fig:safetybench-training} shows checkpoint validation accuracy. Calibration starts with $m_0=512$ and collects $m=128$ samples per round, using data from the preceding step without training data or policy updates. This non-adaptive sequence simulates online policy optimization but does not constitute feedback covariate shift. We adopt it to evaluate risk estimation and high-probability risk control under potentially larger shifts than standard online learning.

\begin{remark} Our work is primarily motivated by feedback covariate shift induced by online policy updates, with deployment risk controlled by adjusting the policy's action distribution. Our experiments also consider non-adaptive covariate shift and risk-control settings. In the non-adaptive setting, the policy sequence is independent of the observed data. Our SafetyBench experiment follows this setting, with no feedback-driven policy updates, and such a setting can create a larger distribution shift than standard online policy learning with feedback covariate shift in this dataset. MedicalQA, discussed in Section~\ref{sec:medicalQA}, illustrates the risk-control framework described in CRC \cite{angelopoulos2024conformal}. Table~\ref{tab:experiment_settings} summarizes the control objectives and distribution-shift settings across our experiments. Appendix~\ref{app:comparisonofdefinition} further clarifies the distinctions between risk control and policy control, and between feedback covariate shift and non-adaptive covariate shift.
\end{remark}

\begin{figure*}[t]
\centering
\setlength{\tabcolsep}{2pt}

\begin{tabular}{
    @{}
    *{3}{>{\centering\arraybackslash}m{0.32\textwidth}}
    @{}
}

Breast cancer &
Stock investment &
SafetyBench
\\[0.3em]

\includegraphics[width=\linewidth]{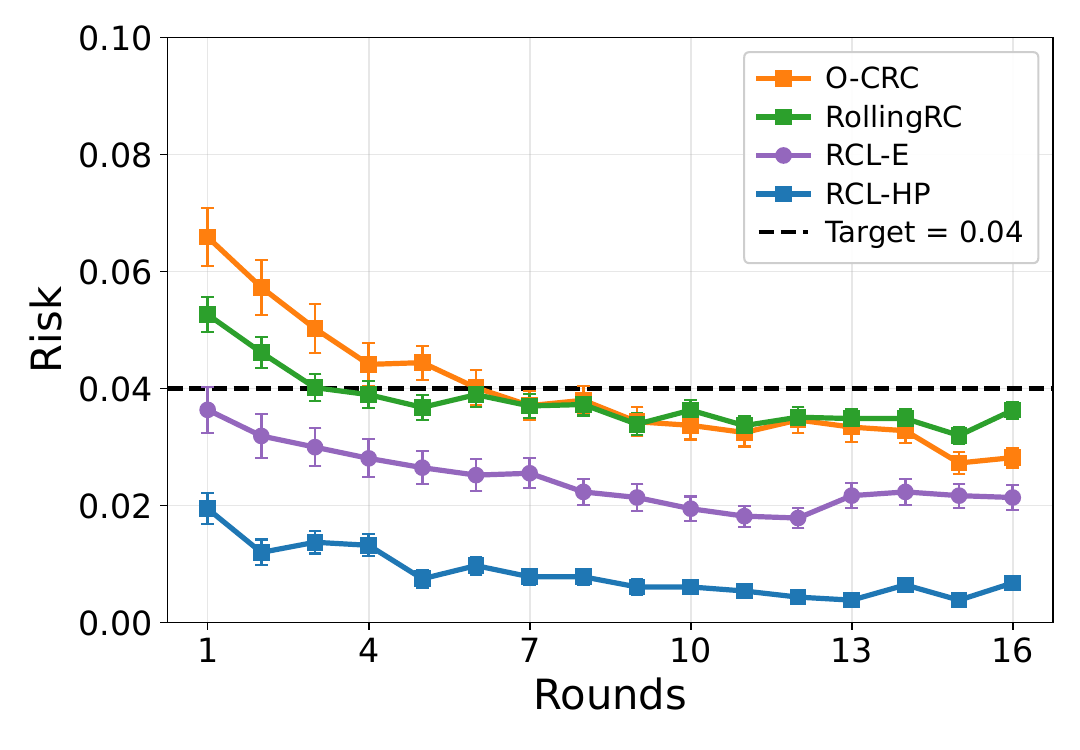}
&
\includegraphics[width=\linewidth]{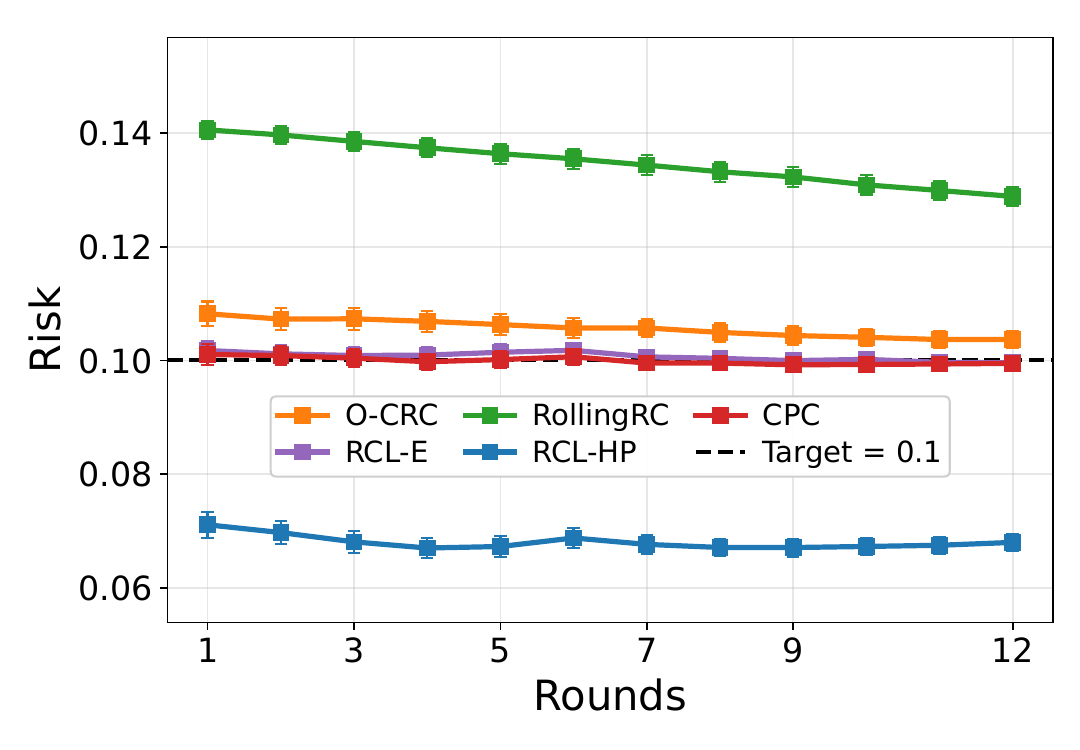}
&
\includegraphics[width=\linewidth]{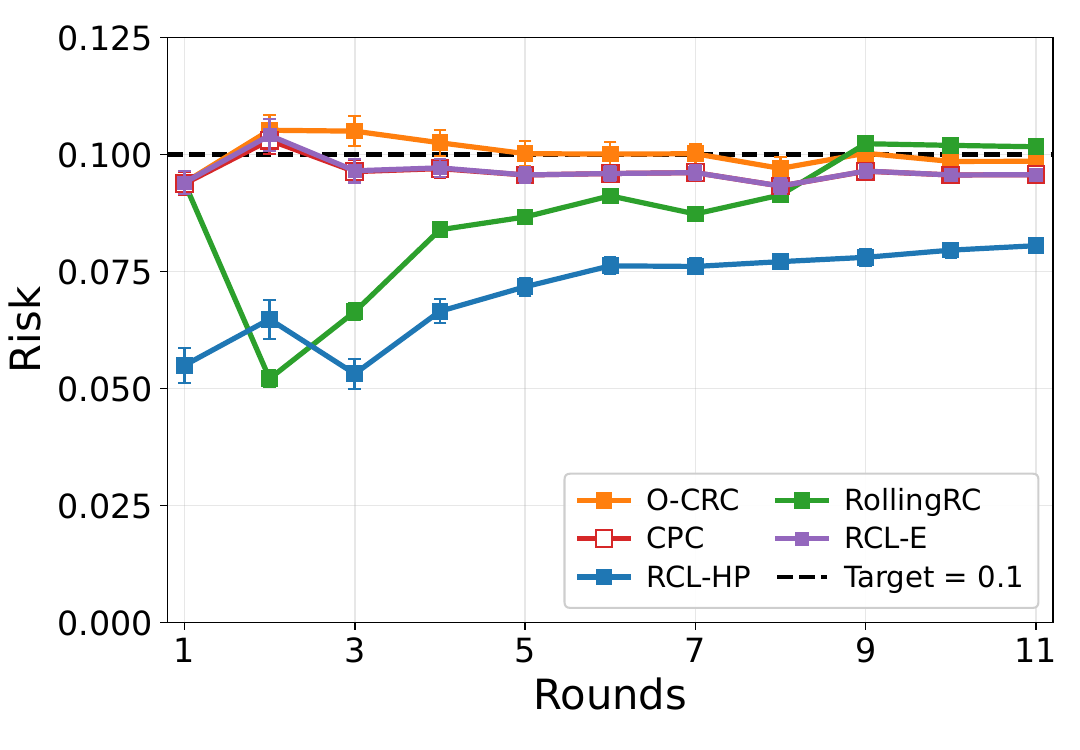}
\\[0pt]
\\[0.3em]

\includegraphics[width=\linewidth]{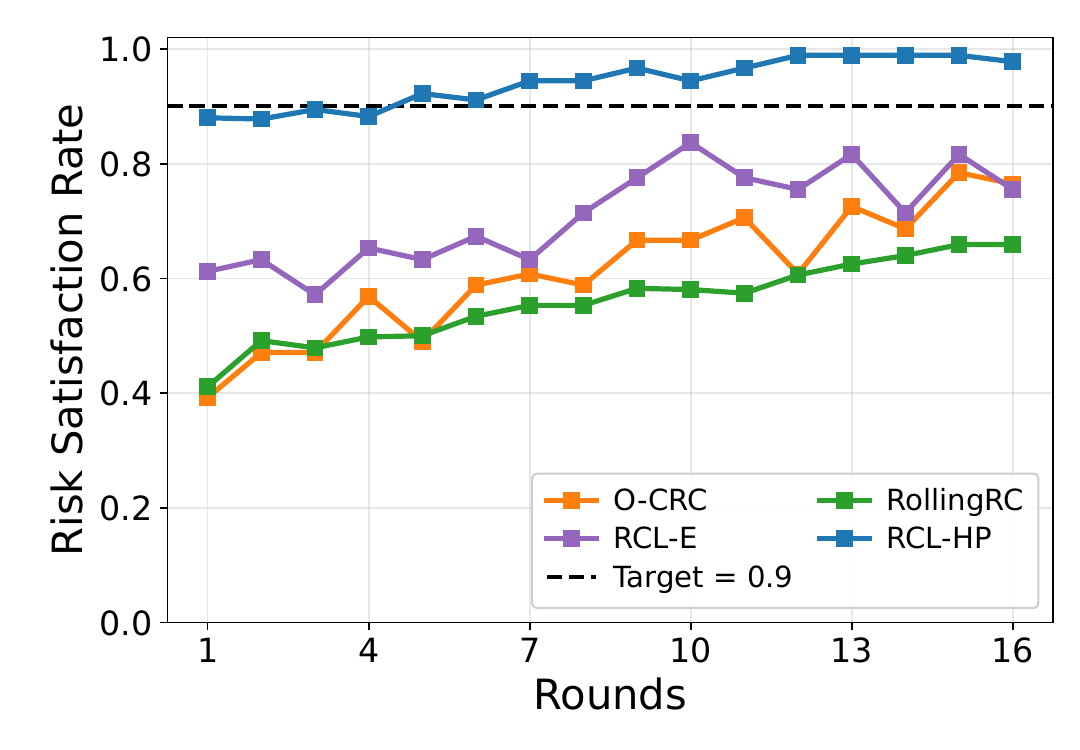}
&
\includegraphics[width=\linewidth]{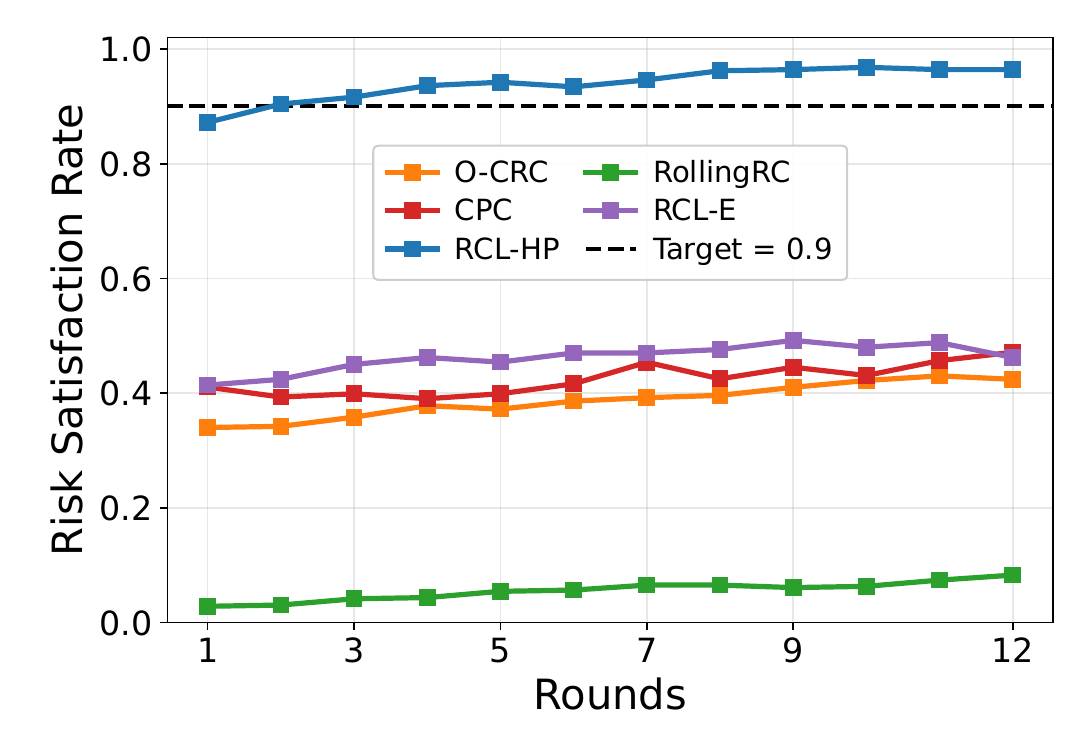}
&
\includegraphics[width=\linewidth]{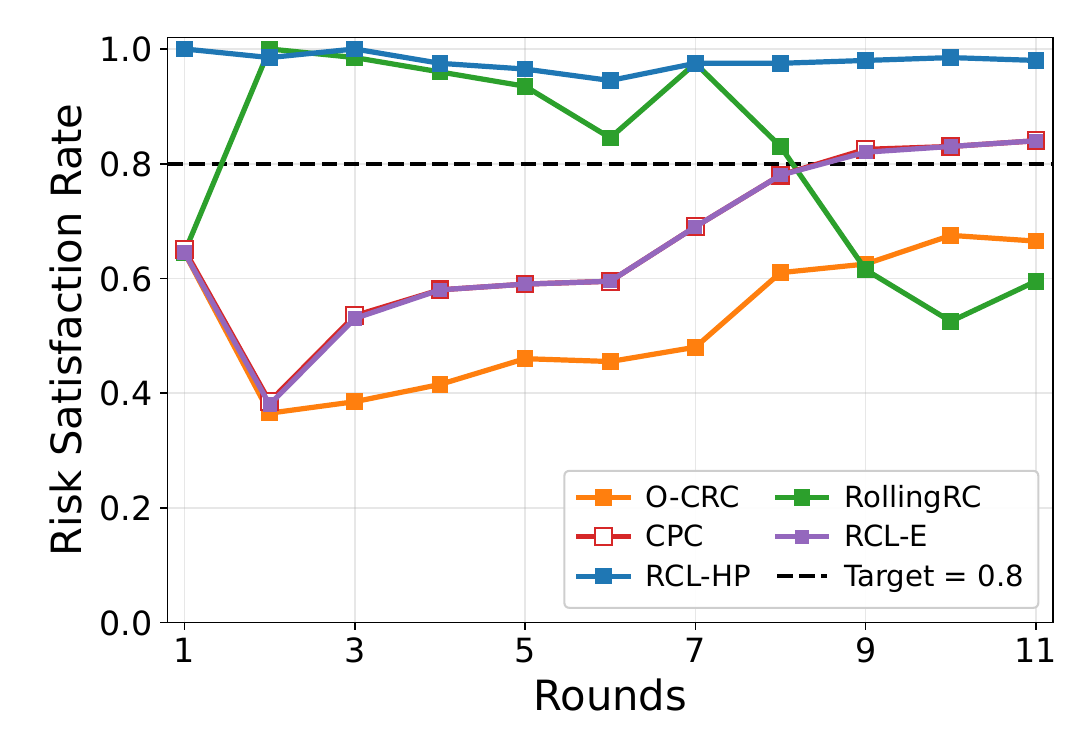}
\\[0pt]

\end{tabular}

\caption{
Comparison across breast cancer diagnosis, stock investment,
and SafetyBench. We report risk and empirical risk satisfaction on two rows;
each column corresponds to one task.
See Figure~\ref{fig:reward} for reward.
Risk for breast cancer, stock investment, and SafetyBench
is measured by FNR, the probability of selecting the worst
investment, and the error rate among answered questions,
respectively.
Reward represents task-specific utility: accuracy for breast cancer,
investment income for stock investment, and cumulative correct
answers for SafetyBench.
Results are based on 500 runs for breast cancer and stock investment,
and 200 calibration histories for SafetyBench.
Breast cancer uses $\alpha=0.04$ and $\delta=0.1$.
Stock investment uses $\alpha=0.1$ and $\delta=0.1$.
SafetyBench uses $\alpha=0.1$ and $\delta=0.2$ for RCL-HP.
RCL-E achieves expected risk control in the monotonic risk setting
(breast cancer), and RCL-HP achieves empirical high-probability
control across the three tasks and outperforms baselines.
}
\label{fig:main}
\end{figure*}

\textbf{Results:} Figure \ref{fig:main} shows the main results. We first evaluate our method, RCL-E risk control in expectation, on a monotonic risk function with breast cancer. We do not report the CPC in breast cancer as it is the same as RCL-E. We show that, compared with the baseline that controls risk in expectation, our method maintains risk below the threshold from the start with a small dataset, while the baselines require more data to meet the constraint. This demonstrates the effectiveness of our permutation-based risk estimator. We then evaluate our method, RCL-HP high-probability risk control, on three datasets. Across the three datasets, RCL-HP achieves the lowest risk. In terms of risk satisfaction, our method empirically achieves high-probability risk control.

\begin{figure*}[t]
\centering
\captionsetup[subfigure]{
    labelfont=normalfont,
    textfont=normalfont,
    justification=centering,
    skip=2pt
}

\begin{subfigure}[t]{0.32\textwidth}
    \centering
    \includegraphics[width=\linewidth]{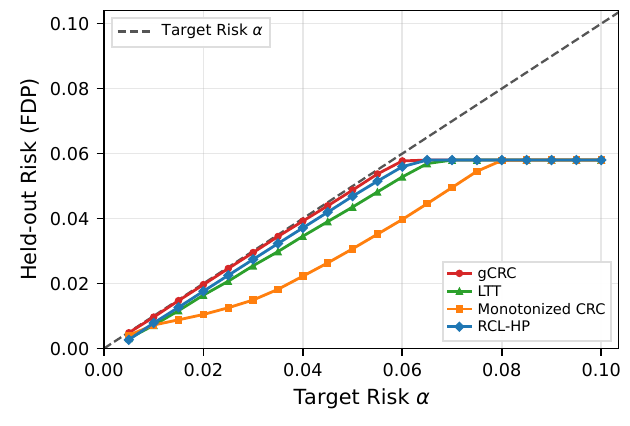}
    \caption{Risk}
    \label{fig:medical_qa_risk}
\end{subfigure}
\hfill
\begin{subfigure}[t]{0.32\textwidth}
    \centering
    \includegraphics[width=\linewidth]{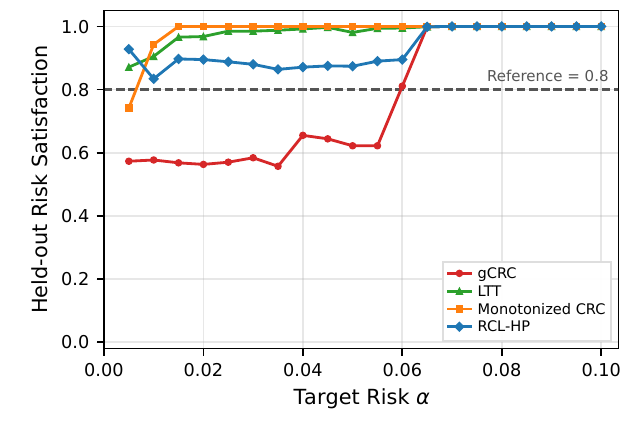}
    \caption{Risk satisfaction}
    \label{fig:medical_qa_satisfaction}
\end{subfigure}
\hfill
\begin{subfigure}[t]{0.32\textwidth}
    \centering
    \includegraphics[width=\linewidth]{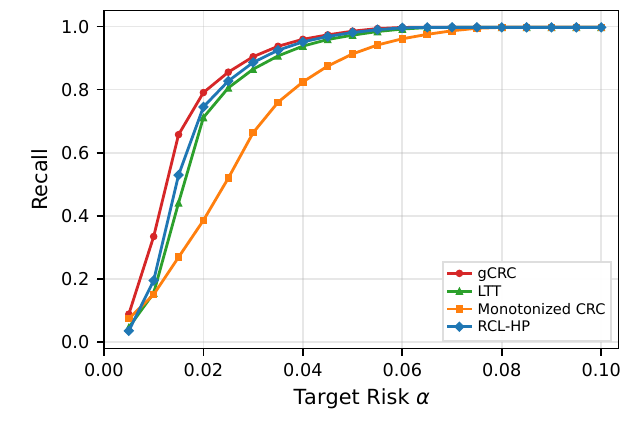}
    \caption{Reward}
    \label{fig:medical_qa_recall}
\end{subfigure}

\caption{
Comparison of medical QA across target risk levels $\alpha$.
Risk is measured by mean held-out false discovery proportion (FDP),
and reward by supported-claim recall.
Results are based on 1,000 random calibration/test splits.
The satisfaction rate is the fraction of splits with mean
held-out FDP at most $\alpha$.
Both LTT and RCL-HP use $\delta=0.2$.
}
\label{fig:medical_qa}
\vspace{-20pt}
\end{figure*}

\begin{figure}[t]
    \begin{subfigure}[b]{0.31\textwidth}
        \centering
        \includegraphics[width=\textwidth]{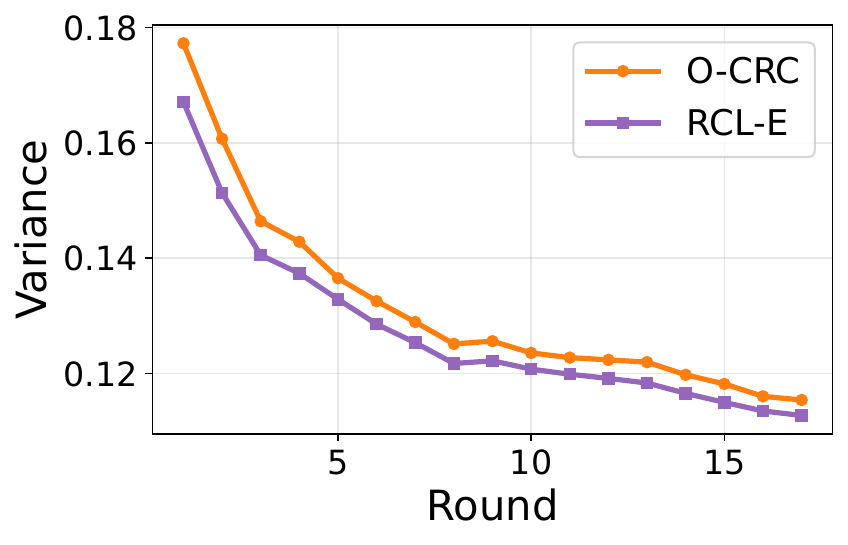}
        \caption{Variance}
    \end{subfigure}
    \hfill
    \begin{subfigure}[b]{0.3\textwidth}
        \centering
        \includegraphics[width=\textwidth]{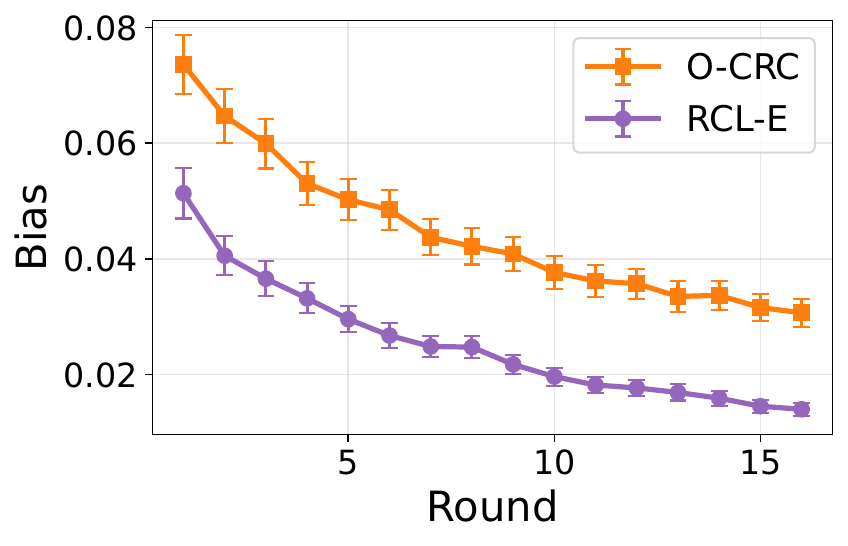}
        \caption{Bias}
    \end{subfigure}
\hfill
     \begin{subfigure}[b]{0.3\textwidth}
        \centering
        \includegraphics[width=\textwidth]{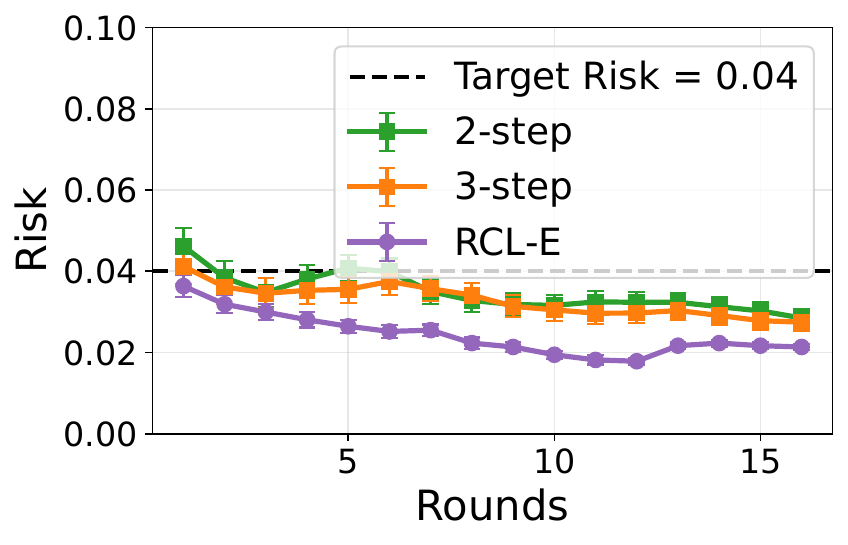}
        \caption{Multistep weight estimation}
    \end{subfigure}
    \caption{ (a)-(b) Experiment on Breast Cancer Diagnosis. Our risk evaluator exhibits lower variance and bias than the importance-weighted method in O-CRC. (c)  We compare against the \emph{multi-step} permutation weight estimator.
    Our full-assignment permutation approximation achieves lower risk.}
    \label{fig:main_risk_evaluator}
    \vspace{-10pt}
\end{figure}

\subsection{Evaluation of High-Probability Risk Control in Factual Medical QA}
\label{sec:medicalQA}
\textbf{Datasets:} We evaluate the RCL-HP on the MedLFQA dataset \citet{jeong2024olaph}. The dataset of \citet{cherian2024large} contains GPT-3.5-Turbo responses, atomic
subclaims extracted via GPT-4o, and factuality annotations
obtained by checking each subclaim against reference answers. Given a confidence score $p(C_{ij} | X_i)$ for each sub-claim, we filter claims below a threshold $\lambda$ and measure the
false discovery rate (FDR), the fraction of retained claims
that are false, as our risk. The reward is recall, the fraction
of total true claims retained. We calibrate the GPT-4o confidence threshold, without sequential data collection or policy updates, to evaluate Gaussian-bootstrap high-probability risk control in the standard fixed-distribution setting. Risk is non-monotonic in $\lambda$. Experimental settings are summarized in Tables~\ref{tab:experiment_metrics} and~\ref{tab:experiment_settings}.

\textbf{Baselines:} In this setting, we mainly compare the control method in \cite{prinster2026conformal} rather than the risk estimator, since there is no policy update and no distribution shift. Generalized conformal policy control (\textbf{gCRC}) extends conformal policy control to the one-step risk control setting without policy updates, using the same policy selection method as CPC. \textbf{Monotonic-risk CRC} addresses non-monotonic risk by monotonizing the risk for each data point in $\lambda$ and then applying standard conformal risk control to the resulting monotonic risk function. Learn-then-Test (\textbf{LTT}) \cite{angelopoulos2021learn} treats each candidate policy as a hypothesis and tests whether its risk is below the target threshold. Details of the baselines and the task setting are in Appendix \ref{app: daatset_medical}.

\textbf{Results:} Figure \ref{fig:medical_qa} shows FDR and recall across risk levels, $\alpha$. RCL-HP controls FDR at or below the target level across all $\alpha$ values while also achieving a high-probability guarantee that FDR stays below the target level. gCRC is less conservative but does not empirically provide a high-probability guarantee. LTT, though, achieves high-probability risk control but is much more conservative because it uses concentration inequalities. Monotonized-loss CRC is conservative due to Jensen’s inequality: monotonizing each loss
is far more conservative than monotonizing the empirical
risk even though it is only considering risk control in expectation. 

Note that the LTT is designed for i.i.d. data and may fail when distribution shift or feedback covariate shift exists. In the medical QA task, the policy is not updated, and no feedback covariate shift exists. In Appendix \ref{sec:highprobcomp}, we provide additional analysis on LTT and another UCB-based method. We show that LTT may fail under covariate shift because it ignores the shift and can be overly conservative. A detailed comparison of our method with other high-probability risk control methods regarding conservatism and feedback covariate shift can be found in Appendix \ref{sec:highprobcomp}. 

\subsection{Comparison of Weighted Risk Estimation Methods}
\label{sec:riskestimatecomp}
\textbf{Outperforming O-CRC with Importance Weights.} O-CRC uses a self-normalized importance-weighting estimator to estimate risk and then applies risk control on top of it. Under a monotonic risk function, the difference between RCL-E and O-CRC is the weighting method, i.e., permutation weight vs. importance weight. The breast cancer results in Figure \ref{fig:main} show that our method achieves lower risk and stays below the target threshold throughout online learning, while O-CRC requires much more calibration data to reach the target risk level. This demonstrates the effectiveness of our permutation-based weight estimator.  
In Figure~\ref{fig:main_risk_evaluator} (a) and (b), we see that our risk evaluator exhibits lower variance and bias than the standard importance-weighted method. This empirically demonstrates the lower-variance property. 

\textbf{Comparison of Weight Estimation Performances.}
We compare RCL-E with alternative approximations of the permutation
marginal weights to show that our Sinkhorn approximation is effective. Specifically, we consider the \emph{multi-step} estimator of
\citet{prinster2024conformal} in Figure~\ref{fig:main_risk_evaluator} (c), which approximates the full permutation marginal
by restricting the assignment space to the most recent positions; details are
provided in Appendix~\ref{app:baseline}. Our method consistently achieves lower risk,
showing the benefit of modeling the full permutation assignment, and our Sinkhorn method is an effective approximation. Moreover, as the
multi-step window increases and better approximates the full assignment, its risk decreases and approaches that of our method. 

\subsection{Additional Ablation Studies}

We analyze the Sinkhorn approximation error in Appendix \ref{app:additional-approximation} and provide empirical and theoretical analysis of how this weight-approximation error affects the theoretical guarantee in Appendix \ref{app:How approximation error affects the guarantee}. We also empirically verify the assumption that the log range does not increase in Appendix \ref{app:additional-log-range} and empirically verify Theorem \ref{thm:main_boot} and its assumptions in Appendix \ref{app: Empirical Evaluation of the Assumption in Theory}.

\section{Conclusion}
\label{sec:discussion_main}
In our work, we present a high-probability risk control method using a permutation-based weighted risk evaluator that employs the Sinkhorn approximation to compute the permutation marginal weight. Theoretically, we show that our method has no larger worst-case variance than the standard importance-weighted method. We also show that our policy selection method achieves asymptotic high-probability risk control. Empirically, we validate our claim that our methods achieve high-probability risk control in safe bandit, safe question answering, and medical QA tasks. We also validated that our efficient risk evaluator outperforms baselines in expected risk control. Future work includes extending the framework to safe LLM reasoning and safe exploration in RL. 

\section*{Acknowledgments}
This project is partially supported by Coefficient Giving, the Gordon
and Betty Moore Foundation grant \#12128, and Schmidt Sciences through
the ``Safely Confident Reasoning'' project in its Science of Trustworthy
AI program. The authors would like to thank the anonymous reviewers
for their helpful feedback. 

\bibliographystyle{unsrtnat}
\bibliography{references}

\clearpage
\newpage
\appendix

\section{Related Work}
\label{app:related_work}

\paragraph{Standard Conformal Prediction and  Risk Control for Exchangeable Data} 
Conformal prediction (CP) is a framework for quantifying uncertainty in machine learning that uses calibration data to turn point predictions into valid, rigorously calibrated prediction sets or intervals. The classic textbook reference for CP and related methods is \citet{vovk2005algorithmic}, but also see \citet{angelopoulos2024theoretical} for a more modern textbook and \citet{angelopoulos2023conformal} for a gentle introduction to CP. Standard CP methods generally assume that the calibration and test data are \textit{exchangeable}, meaning that their joint distribution is invariant to permutations. Informally, exchangeability means the distribution is unchanging across deployment environments and across time; for example, independent and identically distributed (i.i.d.) data are the most common example of exchangeable data. \citet{angelopoulos2024conformal} introduced 
conformal risk control (CRC), an extension of conformal prediction to control the expected value of any loss function that is monotonically parameterized by a control parameter, $\lambda$. CRC recovers standard CP when the loss function in question is a miscoverage indicator and the tuning parameter controls the size of the prediction set. For example, CRC can be used to control the false negative rate in multilabel classification, where the tuning parameter controls the size of the prediction set. 

\paragraph{Weighted Conformal Prediction for Adapting to Fixed Distribution Shifts} 
\citet{tibshirani2019conformal} introduced weighted conformal prediction (WCP) to extend CP to adapt to a fixed covariate shift between a source and target distribution. That is, WCP uses the likelihood ratio between the source and target distribution to reweight calibration data and attain CP's coverage guarantee even under such covariate shift. More broadly, WCP in \citet{tibshirani2019conformal} applies to data that are \textit{weighted exchangeable}, as defined in that paper. \citet{fannjiang2022conformal} extended WCP to allow for a single round of \textit{feedback covariate shift} (FCS), where a single target distribution depends on data from a single source distribution. \citet{prinster2022jaws, prinster2023jaws} developed cross-validation style WCP methods for standard covariate shift and FCS. \citet{nair2023randomization} studied WCP under multiple rounds of FCS and \citet{stanton2023bayesian} applied WCP to multiround Bayesian optimization similar to the online policy learning setting in our paper, but noted gaps in the theory. \citet{prinster2024conformal} demonstrated and explained how WCP can theoretically generalize to \textit{any} data distribution, a practical focus on multiround FCS induced by online policy learning in black-box optimization and active learning. Relevant to the risk evaluation contribution of our paper, several works have leveraged ideas from WCP for off-policy prediction or evaluation in an offline setting \citep{zhang2023conformal, taufiq2022conformal, foffano2023conformal}. However, these works differ from our focus on an online setting with FCS.
Also relevant are the fixed-weight CP methods of \citet{barber2023}, which characterize a worst-case ``coverage gap,'' and the unifying perspective in \citet{barber2025unifying}. We also note that CRC \citep{angelopoulos2024conformal} allows for weighted extensions to adapt to standard covariate shift. This whole line of literature, however, focuses on developing WCP algorithms for adapting to distribution shifts that are \textit{fixed} ahead of time, whether by the environment or a pre-defined learning procedure; in other words, these works do not consider \textit{selecting} a policy to respect a risk constraint.

\paragraph{Conformal Policy Control for Selecting a Risk-Controlling Policy} Most relevant to our current paper, \citet{prinster2026conformal} introduce conformal policy control (CPC) as a framework for \textit{selecting} a policy that is optimized but still controlling a given risk constraint in expectation. 
In contrast to classic conservative optimization methods that generally treat the search for a risk-controlling policy as a hyperparameter tuning problem (e.g., tuning the regularization strength to find an adequate divergence penalty), CPC and our paper circumvent such hyperparameter tuning by directly prescribing a risk-controlling policy to meet a user's declared risk tolerance.
CPC achieves this prescriptive use case  by extending CRC \citep{angelopoulos2024conformal} in two main ways: (i) to allow the tuning parameter to control the policy rather than the loss; and (ii) to allow for non-monotonicity via a more conservative selection criterion. 
As described in the main text, however, CPC has at least two limitations that we aim to address in the current paper. First, CPC's guarantees hold for exact permutation weights (from \citet{prinster2024conformal}). Still, the CPC paper \citep{prinster2026conformal} does not theoretically characterize any relation between its practical weight estimation and exact permutation weights. To address this limitation, we propose an efficient method based on the Sinkhorn algorithm to estimate permutation marginal probability weights, and we show that permutation weights yield a lower worst-case variance than sample-wise importance weights (which can fail in FCS). Second, the CPC guarantee controls risk in \textit{expectation} (as in CRC), which can fail to control risk for a given policy deployment if variance is high. To address this second limitation, we instead focus on (asymptotic) high-probability risk control, which can (asymptotically) assure that a \textit{given} policy deployment is likely safe.

\paragraph{High-Probability Risk Control}
Much of the existing work on high-probability risk control assumes exchangeable or i.i.d. data \citep{bates2021distribution, angelopoulos2021learn}.  However, such assumptions are often violated in adaptive decision-making settings, where the data collection process evolves with the policy itself. \citet{fannjiang2025reliable} study reliable algorithm selection with high-probability guarantees under distribution shift, but in practice they require density-ratio estimation so the guarantees may not directly apply due to estimation error. 
Existing high-probability approaches also typically require multiple testing corrections \citep{bates2021distribution, angelopoulos2021learn, fannjiang2025reliable}, which can become overly conservative. For example, Learn-then-Test achieves an overall error level $\delta$ by requiring each hypothesis test to pass at a much smaller level, e.g.,
$\delta / |\Lambda|$,
which can be prohibitively conservative when the candidate family $|\Lambda|$ is large. 
In contrast, our method develops an asymptotic high-probability risk-control framework based on simultaneous bootstrap correction, which directly accounts for the selection effect while yielding less conservative risk control.
\paragraph{Permutation Marginal Weight Approximation} 
Recent works on conformal prediction and risk control interpret conformal methods
through permutation testing under distribution shift. However,
computing exact permutation marginals quickly becomes intractable as the number
of observations grows. 
\citet{tibshirani2019conformal} acknowledge the computational challenge associated with computing conformal weights for complicated cases of weighted exchangeable data, although 
they limited their practical focus to the relatively simple case of standard covariate shift, where a single source distribution is independent of a single target distribution. 
\citet{prinster2024conformal} use a surrogate approximation that considers only
the permutation assignment at the most recent position, which may introduce
additional approximation bias. \citet{prinster2026conformal} further propose an
averaged weighting approximation for policy control under adaptive data
collection that treats past observations as sampled exchangeably from the mixture distribution of past policies.  \citet{nair2023randomization} and \citet{barber2025unifying} propose using Monte Carlo randomization methods
to estimate permutation weights,
but these randomization approaches generally suffer from high variance.
To obtain a more accurate and scalable approximation, we build on ideas from
permutation marginal inference and entropy-regularized optimization over the
Birkhoff polytope. In particular, \citet{mena2019sinkhorn} propose a Sinkhorn
variational relaxation that replaces the intractable permutation entropy with a
tractable entrywise entropy surrogate, leading to efficient matrix-scaling algorithms based on Sinkhorn iterations \citep{sinkhorn1967concerning}. Inspired by this line of work, we use
a Sinkhorn-based approximation to efficiently estimate permutation marginal
weights induced by adaptive policy assignments under feedback covariate shift.

Also, the quantile-based methods such as \citep{romano2019conformalized} predict the adaptive quantile for the regression and VaR, and CVaR \cite{rockafellar2000optimization,rockafellar2002conditional} predict the statistical confidence bound of the risk. Our work focuses on expected policy risk under FCS, with bootstrap quantiles describing joint estimation error rather than the tail of individual losses. Another related line of work is model-based OPE. OPE includes outcome-model estimators, sequential doubly robust estimators, and model-based bootstrap confidence bounds \cite{dudik2011doubly,yang2023distributionally,guo2024distributionally}. These provide alternative approaches to risk estimation, while our method uses permutation weighting without a separate outcome model and studies calibration for risk-constrained policy selection under a feedback-dependent sampling process. Randomization tests and generalized conformal weighting handle adaptive sampling \cite{nair2023randomization}. Our method shares this likelihood-based foundation but targets expected-loss calibration and uses Sinkhorn approximations for the weights instead of randomization tests or Monte Carlo bootstrapping for permutation estimation.

\section{Algorithm}

\begin{breakablealgorithm}
\caption{High-Probability and Efficient Risk Control for Online Policy Learning}
\label{alg:perm_crc_short}
\begin{algorithmic}[1]
\Require Initial training data $\mathcal D^{\mathrm{tr}}_0$;
calibration data $\mathcal D^{\mathrm{cal}}_0$ with recorded logging policies;
finite ordered candidate set $\Lambda$; risk tolerance $\alpha$;
failure probability $\delta$; loss bound $B$; batch size $m$;
number of bootstrap draws $K$.
\Require A known fallback policy $\pi_{\mathrm{safe}}$
whose deployment risk is at most $\alpha$.
\Ensure Deployed policies $\{\widehat\pi^{t+1}\}_{t\geq 0}$
and updated training and calibration data.

\For{$t=0,1,2,\ldots$}
    \State Update the base policy $\pi_{\theta}^{t+1}$
    using $\mathcal D^{\mathrm{tr}}_t$.
    \State Construct candidate policies
    $\{\pi_{\theta,\lambda}^{t+1}:\lambda\in\Lambda\}$.
    \State Let $n_t=|\mathcal D^{\mathrm{cal}}_t|$.
    Index the calibration records as
    $Z_i=(X_i,A_i,R_i,L_i)$, $i=1,\ldots,n_t$,
    with corresponding logging policies $\rho_i$.
    \State Obtain a provisional context $\widetilde X_{t+1}$,
    shared across candidates.
    \State Set $\widetilde L_i=L_i$ for $i\leq n_t$
    and $\widetilde L_{n_t+1}=B$.

    \For{$\lambda\in\Lambda$}
        \State Sample a provisional action
        $\widetilde A_{t+1}^{(\lambda)}
        \sim\pi_{\theta,\lambda}^{t+1}
        (\cdot\mid\widetilde X_{t+1})$.
        \State Form augmented context--action pairs
        \[
        (\widetilde X_i^{(\lambda)},
         \widetilde A_i^{(\lambda)})
        =
        \begin{cases}
            (X_i,A_i), & i\leq n_t,\\
            (\widetilde X_{t+1},
             \widetilde A_{t+1}^{(\lambda)}), & i=n_t+1.
        \end{cases}
        \]
        \State Build the frozen policy matrix
        $J^{(\lambda)}\in
        \mathbb R_+^{(n_t+1)\times(n_t+1)}$:
        \[
        J_{k,i}^{(\lambda)}
        =
        \begin{cases}
            \rho_k(
                \widetilde A_i^{(\lambda)}
                \mid\widetilde X_i^{(\lambda)}
            ), & k\leq n_t,\\
            \pi_{\theta,\lambda}^{t+1}(
                \widetilde A_i^{(\lambda)}
                \mid\widetilde X_i^{(\lambda)}
            ), & k=n_t+1.
        \end{cases}
        \]
        \State Compute the Sinkhorn approximation and weights:
        \[
        Q^{(\lambda)}
        \gets\operatorname{Sinkhorn}(J^{(\lambda)}),
        \qquad
        p_i(\lambda)\gets Q_{n_t+1,i}^{(\lambda)}.
        \]
        \State Estimate the next-round risk:
        \[
        \hat\ell_{t+1}(\lambda)
        \gets
        \sum_{i=1}^{n_t+1}
        p_i(\lambda)\widetilde L_i.
        \]
    \EndFor

    \If{the risk is monotone and only expected risk control is required}
        \State Set $c_{t+1}\gets 0$.
    \Else
        \State Draw
        $\zeta_i^{(b)}\overset{\mathrm{i.i.d.}}{\sim}N(0,1)$,
        $i=1,\ldots,n_t+1$, $b=1,\ldots,K$,
        sharing each multiplier vector across all candidates.
        \State Compute the bootstrap maxima:
        \[
        T_{t+1}^{*(b)}
        \gets
        \max_{\lambda\in\Lambda}
        \sum_{i=1}^{n_t+1}
        \zeta_i^{(b)}p_i(\lambda)
        \bigl(
            \widetilde L_i-\hat\ell_{t+1}(\lambda)
        \bigr).
        \]
        \State Let $q_{1-\delta,t+1}^{*}$ be the empirical
        $(1-\delta)$-quantile of
        $\{T_{t+1}^{*(b)}\}_{b=1}^{K}$,
        and set $c_{t+1}\gets q_{1-\delta,t+1}^{*}$.
    \EndIf

    \State Construct the feasible set:
    \[
    \widehat\Lambda_{t+1,\mathrm{feas}}
    \gets
    \left\{
        \lambda\in\Lambda:
        \hat\ell_{t+1}(\lambda)+c_{t+1}\leq\alpha
    \right\}.
    \]

    \If{$\widehat\Lambda_{t+1,\mathrm{feas}}\neq\varnothing$}
        \If{the risk is monotone and only expected risk control is required}
            \State Set
            $\hat\lambda_{t+1}
            \gets\min\widehat\Lambda_{t+1,\mathrm{feas}}$.
        \Else
            \State Select
            $\hat\lambda_{t+1}
            \in\widehat\Lambda_{t+1,\mathrm{feas}}$
            using a prespecified selection rule.
        \EndIf
        \State Set
        $\widehat\pi^{t+1}
        \gets\pi_{\theta,\hat\lambda_{t+1}}^{t+1}$.
    \Else
        \State Set
        $\widehat\pi^{t+1}\gets\pi_{\mathrm{safe}}$.
    \EndIf

    \State Initialize
    $\mathcal D^{\mathrm{tr}}_{t+1}
    \gets\mathcal D^{\mathrm{tr}}_t$
    and
    $\mathcal D^{\mathrm{cal}}_{t+1}
    \gets\mathcal D^{\mathrm{cal}}_t$.
    \For{$j=1,\ldots,m$}
        \State Observe a new context $X_{t+1,j}$ and sample
        \[
        A_{t+1,j}
        \sim\widehat\pi^{t+1}(\cdot\mid X_{t+1,j}).
        \]
        \State Observe $(R_{t+1,j},L_{t+1,j})$ and form
        \[
        Z_{t+1,j}
        =
        (X_{t+1,j},A_{t+1,j},R_{t+1,j},L_{t+1,j}).
        \]
        \If{$j$ is even}
            \State Append $Z_{t+1,j}$ to
            $\mathcal D^{\mathrm{tr}}_{t+1}$.
        \Else
            \State Append $Z_{t+1,j}$ to
            $\mathcal D^{\mathrm{cal}}_{t+1}$,
            recording $\widehat\pi^{t+1}$ as its logging policy.
        \EndIf
    \EndFor
\EndFor
\end{algorithmic}
\end{breakablealgorithm}

\section{More Details on Permutation Marginal Weights}

\subsection{Permutation Marginal Weight Calculation}
\label{app:permutation_marginal}

In this section, we derive the permutation marginal weight
$p_i(\lambda)$ using the law of total probability. Let
\[
E_Z := \{Z_1,\ldots,Z_{t+1}\}
\]
denote the unordered bag of observed data, and let $\hat Z_{t+1}$ denote the
random sample assigned to the final position under a random ordering. Then
\begin{align}
\label{eq:total_law}
\mathbb P\bigl(\hat Z_{t+1}=Z_i \mid E_Z\bigr)
&=
\frac{
\mathbb P\bigl(\hat Z_{t+1}=Z_i, E_Z\bigr)
}{
\mathbb P(E_Z)
} \nonumber \\
&=
\frac{
\sum_{\sigma:\sigma(t+1)=i}
\pi\bigl(z_{\sigma(1)},\ldots,z_{\sigma(t+1)}\bigr)
}{
\sum_{\sigma}
\pi\bigl(z_{\sigma(1)},\ldots,z_{\sigma(t+1)}\bigr)
}.
\end{align}
Here, $\sigma$ ranges over all permutations of $\{1,\ldots,t+1\}$, and
$\sigma(t+1)=i$ means that sample $i$ is assigned to the final position.
The term $\pi(z_{\sigma(1)},\ldots,z_{\sigma(t+1)})$ denotes the joint density
of the data under the ordering $\sigma$. Under exchangeability,
\[
\pi(z_{\sigma(1)},\ldots,z_{\sigma(t+1)})
=
\pi(z_1,\ldots,z_{t+1})
\]
for all $\sigma$, and the marginal weights reduce to uniform weights.

Under MFCS for contextual bandits, following \citet{prinster2024conformal},
the joint density can be factorized into time-dependent and time-invariant
terms:
\begin{align}
\label{eq:calculate_f}
\pi(Z_1,\ldots,Z_{t+1})
&=
\prod_{j=1}^{t+1}
\underbrace{
p(A_j,X_j\mid Z_1,\ldots,Z_{j-1})
}_{\text{time-dependent}}
\prod_{j=1}^{t+1}
\underbrace{
p(L_j\mid X_j,A_j)
}_{\text{invariant}}
\nonumber \\
&=
\prod_{j=1}^{t+1}
\underbrace{
p(A_j\mid Z_1,\ldots,Z_{j-1},X_j)
}_{\text{policy-dependent}}
\prod_{j=1}^{t+1}
p_X(X_j)
\prod_{j=1}^{t+1}
\underbrace{
p(L_j\mid X_j,A_j)
}_{\text{invariant}} .
\end{align}
In the permutation ratio in Eq.\eqref{eq:total_law}, the context density
$\prod_{j=1}^{t+1}p_X(X_j)$ and the loss conditional density
$\prod_{j=1}^{t+1}p(L_j\mid X_j,A_j)$ are invariant to the ordering and cancel
between numerator and denominator. Therefore, only the policy-dependent terms
remain. Plugging Eq.\eqref{eq:calculate_f} into Eq.\eqref{eq:total_law} gives
\begin{align}
\label{eq:full_adaptive_weight}
\mathbb P\bigl(\hat Z_{t+1}=Z_i \mid E_Z\bigr)
=
\frac{
\sum_{\sigma:\sigma(t+1)=i}
\prod_{j=1}^{t+1}
p\bigl(
A_{\sigma(j)}
\mid
Z_{\sigma(1)},\ldots,Z_{\sigma(j-1)},X_{\sigma(j)}
\bigr)
}{
\sum_{\sigma}
\prod_{j=1}^{t+1}
p\bigl(
A_{\sigma(j)}
\mid
Z_{\sigma(1)},\ldots,Z_{\sigma(j-1)},X_{\sigma(j)}
\bigr)
}.
\end{align}

Computing Eq.\eqref{eq:full_adaptive_weight} exactly would require re-learning the
policy for every prefix of every permutation, which is infeasible. We therefore
use a fixed learned-policy-sequence approximation: the learned policies are held
fixed, and only the observed data points are permuted. Let
\[
\pi_{\theta,\lambda}^{1},
\ldots,
\pi_{\theta,\lambda}^{t},
\pi_{\theta,\lambda}^{t+1}
\]
denote the fixed policy sequence, where $\pi_{\theta,\lambda}^{t+1}$ is the
candidate deployment policy for the final position. Then the permutation
marginal weight is approximated by
\begin{align}
\label{eq:permutation_weight}
p_i(\lambda)
:=
\mathbb P\bigl(\hat Z_{t+1}=Z_i \mid E_Z,\lambda\bigr)
\approx
\frac{
\sum_{\sigma:\sigma(t+1)=i}
\prod_{j=1}^{t+1}
\pi_{\theta,\lambda}^{j}
\bigl(A_{\sigma(j)}\mid X_{\sigma(j)}\bigr)
}{
\sum_{\sigma}
\prod_{j=1}^{t+1}
\pi_{\theta,\lambda}^{j}
\bigl(A_{\sigma(j)}\mid X_{\sigma(j)}\bigr)
}.
\end{align}
\subsection{Proof of Proposition~\ref{prop:snips}: Reduction to SNIPS}
\label{app:proof_snips}

We show that, under standard covariate shift, the exact permutation marginal
weight reduces to the self-normalized importance sampling (SNIPS) weight.
Suppose all historical data are collected by a fixed behavior policy $\pi_b$,
and the next-round risk is evaluated under the target policy
$\pi_{\theta,\lambda}^{t+1}$.

We first write the probability that the final position is occupied by the
observed data point $Z_i=(X_i,A_i,R_i,L_i)$. Under a fixed behavior policy
$\pi_b$ for the first $t$ positions and the target policy
$\pi_{\theta,\lambda}^{t+1}$ for the $(t+1)$-th position, we have
\begin{align}
\mathbb P(\hat Z_{t+1}=Z_i \mid E_Z,\lambda)
=
\frac{
\sum_{\sigma:\,\sigma(t+1)=i}
\left[
\prod_{k=1}^{t}
\pi_b(A_{\sigma(k)}\mid X_{\sigma(k)})
\right]
\pi_{\theta,\lambda}^{t+1}(A_i\mid X_i)
\prod_{h=1}^{t+1}p_X(X_h)
}{
\mathcal Z
},
\end{align}
where $\mathcal Z$ is the normalizing constant. Since
$\prod_{h=1}^{t+1}p_X(X_h)$ does not depend on the permutation, it cancels
after normalization. Moreover, conditional on $\hat Z_{t+1}=Z_i$, the
remaining $t$ observations can be assigned to the first $t$ positions in
$t!$ possible ways. Therefore,
\begin{align}
\mathbb P(\hat Z_{t+1}=Z_i \mid E_Z,\lambda)
=
\frac{
t!\,
\pi_{\theta,\lambda}^{t+1}(A_i\mid X_i)
\prod_{j\neq i}^{t+1}
\pi_b(A_j\mid X_j)
}{
\mathcal Z
}.
\end{align}

The normalizing constant is obtained by summing over all possible choices of
the observation assigned to the final position:
\begin{align}
\mathcal Z
=
\sum_{\ell=1}^{t+1}
t!\,
\pi_{\theta,\lambda}^{t+1}(A_\ell\mid X_\ell)
\prod_{j\neq \ell}^{t+1}
\pi_b(A_j\mid X_j).
\end{align}
Hence,
\begin{align}
\mathbb P(\hat Z_{t+1}=Z_i \mid E_Z,\lambda)
=
\frac{
\pi_{\theta,\lambda}^{t+1}(A_i\mid X_i)
\prod_{j\neq i}^{t+1}
\pi_b(A_j\mid X_j)
}{
\sum_{\ell=1}^{t+1}
\pi_{\theta,\lambda}^{t+1}(A_\ell\mid X_\ell)
\prod_{j\neq \ell}^{t+1}
\pi_b(A_j\mid X_j)
}.
\end{align}
Dividing both the numerator and denominator by
$\prod_{j=1}^{t+1}\pi_b(A_j\mid X_j)$ gives
\begin{align}
\mathbb P(\hat Z_{t+1}=Z_i \mid E_Z,\lambda)
=
\frac{
\pi_{\theta,\lambda}^{t+1}(A_i\mid X_i)
/\pi_b(A_i\mid X_i)
}{
\sum_{\ell=1}^{t+1}
\pi_{\theta,\lambda}^{t+1}(A_\ell\mid X_\ell)
/\pi_b(A_\ell\mid X_\ell)
}.
\end{align}
Therefore, the permutation marginal probability of assigning $Z_i$ to the
$(t+1)$-th position coincides with the SNIPS weight.

\subsection{Proof of Proposition~\ref{prop:global_iw}: A Global Importance Weight}
\label{app:proof_global_iw}

The proof is straightforward. From Eq.\eqref{eq:permutation_weight}, under the fixed
policy sequence $\pi_{\theta,\lambda}^{1},\ldots,\pi_{\theta,\lambda}^{t+1}$,
we have
\begin{align}
\label{eq:global_iw_start}
p_i(\lambda)
&=
\frac{
\sum_{\sigma:\,\sigma(t+1)=i}
\prod_{h=1}^{t+1}
\pi_{\theta,\lambda}^{h}
\bigl(A_{\sigma(h)}\mid X_{\sigma(h)}\bigr)
}{
\sum_{\sigma}
\prod_{h=1}^{t+1}
\pi_{\theta,\lambda}^{h}
\bigl(A_{\sigma(h)}\mid X_{\sigma(h)}\bigr)
}.
\end{align}
The numerator corresponds to the total likelihood of all permutations in which
sample $i$ is assigned to the $(t+1)$-th position. Therefore, it can be written as
\begin{align}
\label{eq:numerator_decompose}
&\sum_{\sigma:\,\sigma(t+1)=i}
\prod_{h=1}^{t+1}
\pi_{\theta,\lambda}^{h}
\bigl(A_{\sigma(h)}\mid X_{\sigma(h)}\bigr)
\nonumber\\
&\qquad =
\sum_{\sigma:\,\sigma(t+1)=i}
\pi_{\theta,\lambda}^{t+1}
\bigl(A_i\mid X_i\bigr)
\prod_{h=1}^{t}
\pi_{\theta,\lambda}^{h}
\bigl(A_{\sigma(h)}\mid X_{\sigma(h)}\bigr)
\nonumber\\
&\qquad =
\pi_{\theta,\lambda}^{t+1}
\bigl(A_i\mid X_i\bigr)
\sum_{\sigma:\,\sigma(t+1)=i}
\prod_{h=1}^{t}
\pi_{\theta,\lambda}^{h}
\bigl(A_{\sigma(h)}\mid X_{\sigma(h)}\bigr).
\end{align}

Similarly, the denominator is the total likelihood over all permutations. We
decompose it according to the position $k\in\{1,\ldots,t+1\}$ occupied by
sample $i$:
\begin{align}
\label{eq:denominator_decompose}
&\sum_{\sigma}
\prod_{h=1}^{t+1}
\pi_{\theta,\lambda}^{h}
\bigl(A_{\sigma(h)}\mid X_{\sigma(h)}\bigr)
\nonumber\\
&\qquad =
\sum_{k=1}^{t+1}
\sum_{\sigma:\,\sigma(k)=i}
\prod_{h=1}^{t+1}
\pi_{\theta,\lambda}^{h}
\bigl(A_{\sigma(h)}\mid X_{\sigma(h)}\bigr)
\nonumber\\
&\qquad =
\sum_{k=1}^{t+1}
\pi_{\theta,\lambda}^{k}
\bigl(A_i\mid X_i\bigr)
\sum_{\sigma:\,\sigma(k)=i}
\prod_{\substack{h=1\\ h\neq k}}^{t+1}
\pi_{\theta,\lambda}^{h}
\bigl(A_{\sigma(h)}\mid X_{\sigma(h)}\bigr).
\end{align}

Define
\begin{align}
\label{eq:global_context_factor}
C_{i,k}
:=
\sum_{\sigma:\,\sigma(k)=i}
\prod_{\substack{h=1\\ h\neq k}}^{t+1}
\pi_{\theta,\lambda}^{h}
\bigl(A_{\sigma(h)}\mid X_{\sigma(h)}\bigr).
\end{align}
Then the numerator in Eq.\eqref{eq:numerator_decompose} is
\[
\pi_{\theta,\lambda}^{t+1}(A_i\mid X_i) C_{i,t+1},
\]
and the denominator in Eq.\eqref{eq:denominator_decompose} is
\[
\sum_{k=1}^{t+1}
\pi_{\theta,\lambda}^{k}(A_i\mid X_i) C_{i,k}.
\]
Therefore,
\begin{align}
\label{eq:global_iw}
p_i(\lambda)
&=
\frac{
\pi_{\theta,\lambda}^{t+1}(A_i\mid X_i) C_{i,t+1}
}{
\sum_{k=1}^{t+1}
\pi_{\theta,\lambda}^{k}(A_i\mid X_i) C_{i,k}
}.
\end{align}
This shows that the permutation marginal weight can be interpreted as a global
importance-weighting method: the weight assigned to sample $i$ depends not only
on the likelihood of observing $A_i$ under the target policy
$\pi_{\theta,\lambda}^{t+1}$, but also on the global compatibility factors
$C_{i,k}$, which aggregate over all feasible assignments of the remaining samples
to the remaining positions.

\subsection{Proof of Theorem~\ref{thm:sinkhorn}}
\label{app:sinkhorn_proof}

\begin{lemma}[Variational representation of log-sum-exp / Gibbs variational principle]
\label{lemma:log-sum-exp}
Let $\mathcal S$ be a finite set of permutations, and let $f:\mathcal S \to \mathbb R$ be any function. 
Let $\mu(\sigma)$ denote a probability distribution over permutations $\sigma\in\mathcal S$. Then
\[
\log \sum_{\sigma\in\mathcal S} e^{f(\sigma)}
=
\max_{\mu \in \mathcal P(\mathcal S)}
\left\{
\sum_{\sigma\in\mathcal S} \mu(\sigma) f(\sigma) + H(\mu)
\right\},
\]
where 
\[
\mathcal P(\mathcal S) = \left\{\mu:\mu(\sigma)\ge 0,\ \sum_{\sigma\in\mathcal S}\mu(\sigma)=1\right\}
\]
is the probability simplex and
\[
H(\mu) = -\sum_{\sigma\in\mathcal S} \mu(\sigma)\log \mu(\sigma)
\]
is the Shannon entropy. Moreover, the maximizer is
\[
\mu^*(\sigma) 
= 
\frac{e^{f(\sigma)}}{\sum_{\sigma'\in\mathcal S} e^{f(\sigma')}}.
\]
\end{lemma}

\begin{proof}[Proof of Lemma~\ref{lemma:log-sum-exp}]
Let
\[
Z := \sum_{\sigma\in\mathcal S} e^{f(\sigma)}, 
\qquad
\mu^*(\sigma) := \frac{e^{f(\sigma)}}{Z}.
\]
For any $\mu \in \mathcal P(\mathcal S)$, the Kullback--Leibler divergence satisfies
\[
\mathrm{KL}(\mu\|\mu^*) \ge 0,
\]
with equality if and only if $\mu=\mu^*$. Expanding,
\[
\mathrm{KL}(\mu\|\mu^*)
=
\sum_{\sigma\in\mathcal S} 
\mu(\sigma)\log\frac{\mu(\sigma)}{\mu^*(\sigma)}
=
\sum_{\sigma\in\mathcal S}\mu(\sigma)\log \mu(\sigma)
-
\sum_{\sigma\in\mathcal S}\mu(\sigma)\big(f(\sigma)-\log Z\big).
\]
Rearranging gives
\[
\log Z
=
\sum_{\sigma\in\mathcal S}\mu(\sigma)f(\sigma)
-
\sum_{\sigma\in\mathcal S}\mu(\sigma)\log\mu(\sigma)
+
\mathrm{KL}(\mu\|\mu^*).
\]
Since
\[
H(\mu)
=
-\sum_{\sigma\in\mathcal S}\mu(\sigma)\log\mu(\sigma)
\]
and $\mathrm{KL}(\mu\|\mu^*)\ge 0$, we obtain
\[
\sum_{\sigma\in\mathcal S}\mu(\sigma)f(\sigma) + H(\mu) 
\le 
\log Z,
\]
with equality when $\mu=\mu^*$. Therefore,
\[
\log \sum_{\sigma\in\mathcal S} e^{f(\sigma)}
=
\max_{\mu\in\mathcal P(\mathcal S)}
\left\{
\sum_{\sigma\in\mathcal S}\mu(\sigma)f(\sigma) + H(\mu)
\right\}.
\]
\end{proof}

\begin{remark}[Interpretation of the Gibbs variational principle]
\label{rem:gibbs-variational}
The Gibbs variational principle shows that the log-sum-exp over all permutations can be viewed as an optimization over probability distributions on permutations. 
In our setting, each permutation $\sigma$ corresponds to a possible assignment of observed samples to time positions, and $f(\sigma)$ is the log-likelihood score for that ordering. 
The maximizer
\[
\mu^*(\sigma)
=
\frac{e^{f(\sigma)}}{\sum_{\sigma'\in\mathcal S} e^{f(\sigma')}}
\]
is therefore the normalized Gibbs distribution over permutations. 
This provides a probabilistic interpretation of the exact permutation distribution: orderings with larger likelihood scores receive larger probability. At the same time, the entropy term $H(\mu)$ prevents the variational problem from collapsing onto a single highest-scoring permutation. 
The main computational difficulty is that $\mu^*$ is defined over all permutations, whose number grows factorially. 
This motivates replacing the full distribution over permutations by a tractable marginal assignment matrix and approximating its entropy using a marginal-entropy relaxation.
\end{remark}

\begin{lemma}[Expected score expressed by the marginal matrix]
\label{lemma:mu_to_Q}
Let $\mu$ be a probability distribution over permutations $\sigma\in S_n$, and define
\[
f(\sigma)
:=
\sum_{i=1}^n \log J_{i,\sigma(i)},
\]
where $J\in\mathbb R_+^{n\times n}$. 
Define the marginal assignment matrix.
\[
Q_{ij}
:=
\Pr_{\mu}\!\left(\sigma(i)=j\right)
=
\sum_{\sigma:\,\sigma(i)=j}\mu(\sigma).
\]
Then
\[
\mathbb E_{\mu}\!\left[f(\sigma)\right]
=
\sum_{i=1}^n\sum_{j=1}^n
Q_{ij}\log J_{ij}.
\]
\end{lemma}

\begin{proof}[Proof of Lemma~\ref{lemma:mu_to_Q}]
By definition,
\[
\mathbb{E}_{\mu}[f(\sigma)]
=
\sum_{\sigma\in S_n} \mu(\sigma) f(\sigma)
=
\sum_{\sigma\in S_n} \mu(\sigma)
\sum_{i=1}^n \log J_{i,\sigma(i)}.
\]
Interchanging the order of summation gives
\[
\mathbb{E}_{\mu}[f(\sigma)]
=
\sum_{i=1}^n
\sum_{\sigma\in S_n}
\mu(\sigma)\log J_{i,\sigma(i)}.
\]
For a fixed $i$, we group permutations according to the value of $\sigma(i)$:
\[
\sum_{\sigma\in S_n}
\mu(\sigma)\log J_{i,\sigma(i)}
=
\sum_{j=1}^n
\sum_{\sigma:\,\sigma(i)=j}
\mu(\sigma)\log J_{ij}.
\]
Since $\log J_{ij}$ is constant over all permutations satisfying $\sigma(i)=j$, we obtain
\[
\sum_{\sigma\in S_n}
\mu(\sigma)\log J_{i,\sigma(i)}
=
\sum_{j=1}^n
\log J_{ij}
\sum_{\sigma:\,\sigma(i)=j}
\mu(\sigma)
=
\sum_{j=1}^n Q_{ij}\log J_{ij}.
\]
Summing over $i$ yields
\[
\mathbb{E}_{\mu}[f(\sigma)]
=
\sum_{i=1}^n \sum_{j=1}^n Q_{ij}\log J_{ij},
\]
which completes the proof.
\end{proof}

\begin{remark}
Lemma~\ref{lemma:mu_to_Q} shows that the expected log-score of a distribution over permutations depends on $\mu$ only through its marginal assignment matrix $Q$. 
Thus, the likelihood part of the variational objective can be expressed directly in terms of $Q$, even though the entropy term $H(\mu)$ still depends on the full distribution over permutations. 
This key step motivates replacing the intractable optimization over permutation distributions with a tractable relaxation over doubly stochastic matrices.
\end{remark}

Now we prove Theorem~\ref{thm:sinkhorn}.
\begin{proof}
The permanent of $J$ is defined as
\begin{align}
    Z
    =
    \operatorname{perm}(J)
    =
    \sum_{\sigma\in S_n}
    \prod_{i=1}^n J_{i,\sigma(i)},
    \label{eq:logperm}
\end{align}
where $\sigma$ denotes a permutation of $\{1,\ldots,n\}$.

Define the log-score of a permutation $\sigma$ as
\[
f(\sigma)
=
\log\prod_{i=1}^n J_{i,\sigma(i)}
=
\sum_{i=1}^n \log J_{i,\sigma(i)}.
\]
Then the log-permanent can be written as the log-sum-exp over all permutations:
\begin{align}
    \log \operatorname{perm}(J)
    =
    \log \sum_{\sigma\in S_n} e^{f(\sigma)}.
\end{align}

Then, using Lemma~\ref{lemma:log-sum-exp} and Lemma~\ref{lemma:mu_to_Q}, we can rewrite the log-permanent as
\begin{align}
\log Z
=
\log \operatorname{perm}(J)
&=
\log \sum_{\sigma\in S_n} e^{f(\sigma)}
\\
&=
\max_{\mu\in\mathcal P(S_n)}
\left\{
\sum_{\sigma\in S_n}\mu(\sigma)f(\sigma)
+
H(\mu)
\right\}
\\
&=
\max_{\mu\in \mathcal P(S_n)}
\left\{
\sum_{i=1}^n\sum_{j=1}^n
Q_{ij}^{\mu}\log J_{ij}
+
H(\mu)
\right\},
\end{align}
where
\[
Q_{ij}^{\mu}
=
\Pr_{\mu}\!\left(\sigma(i)=j\right)
\]
is the marginal assignment matrix induced by the distribution $\mu$ over permutations.

The entropy term $H(\mu)$ is defined on the full distribution over permutations and cannot, in general, be written exactly as a function of the marginal matrix $Q$ alone. 
We therefore use a marginal-entropy relaxation, replacing $H(\mu)$ by the entrywise entropy of $Q$:
\[
H(\mu)
\approx
H(Q)
=
-\sum_{i=1}^n\sum_{j=1}^n Q_{ij}\log Q_{ij}.
\]
This gives the tractable approximation
\begin{align}
\log Z
=
\log \operatorname{perm}(J)
&\approx
\max_{Q\in B_n}
\left\{
\sum_{i=1}^n\sum_{j=1}^n
Q_{ij}\log J_{ij}
-
\sum_{i=1}^n\sum_{j=1}^n
Q_{ij}\log Q_{ij}
\right\} \\
&=
\max_{Q\in B_n}
\left\{
-\sum_{i=1}^n\sum_{j=1}^n
Q_{ij}\log\frac{Q_{ij}}{J_{ij}}
\right\} \\
&=
-\min_{Q\in B_n}
\mathrm{KL}(Q\|J),
\end{align}
where
\[
B_n
=
\left\{
Q\in\mathbb R_+^{n\times n}:
Q\mathbf 1=\mathbf 1,\ 
Q^\top\mathbf 1=\mathbf 1
\right\}
\]
is the Birkhoff polytope of doubly stochastic matrices.

Equivalently, the approximate marginal assignment matrix is obtained by solving
\[
Q^\star
=
\arg\min_{Q\in B_n}
\mathrm{KL}(Q\|J).
\]
The last row of $Q^\star$ is then used as the approximation to the permutation marginal weights:
\[
p_i(\lambda)
\approx
Q^\star_{n,i},
\qquad i=1,\ldots,n.
\]
\end{proof}

\subsection{Optimization with Sinkhorn algorithm}
\label{app: sinkhorn_optimization}
Consider the entropy-regularized assignment problem
\[
Q^\star
=
\arg\min_{Q\in B_n}
\mathrm{KL}(Q\|J),
\]
where
\[
B_n
=
\left\{
Q\in\mathbb R_+^{n\times n}:
\sum_{j=1}^n Q_{ij}=1,\ 
\sum_{i=1}^n Q_{ij}=1
\right\}
\]
is the set of doubly stochastic matrices, and
\[
\mathrm{KL}(Q\|J)
:=
\sum_{i=1}^n\sum_{j=1}^n
Q_{ij}\log\frac{Q_{ij}}{J_{ij}}.
\]
Here, \(J\in\mathbb R_+^{n\times n}\) is the stage--sample score matrix.

We introduce Lagrange multipliers
\(\{\lambda_i\}_{i=1}^n\) and \(\{\eta_j\}_{j=1}^n\)
for the row- and column-sum constraints, respectively. 
The Lagrangian is
\[
\mathcal L(Q,\lambda,\eta)
=
\sum_{i=1}^n\sum_{j=1}^n
Q_{ij}\log\frac{Q_{ij}}{J_{ij}}
+
\sum_{i=1}^n
\lambda_i
\left(
\sum_{j=1}^n Q_{ij}-1
\right)
+
\sum_{j=1}^n
\eta_j
\left(
\sum_{i=1}^n Q_{ij}-1
\right).
\]
Taking the partial derivative with respect to \(Q_{ij}\) gives
\[
\frac{\partial \mathcal L}{\partial Q_{ij}}
=
\log Q_{ij}+1-\log J_{ij}
+\lambda_i+\eta_j.
\]
At an interior optimum, setting this derivative to zero yields
\[
\log Q_{ij}
=
\log J_{ij}-1-\lambda_i-\eta_j.
\]
Equivalently,
\[
Q_{ij}
=
u_i J_{ij} v_j,
\]
where
\[
u_i
=
e^{-1-\lambda_i},
\qquad
v_j
=
e^{-\eta_j}.
\]
Therefore, any interior stationary point has the matrix-scaling form
\[
Q
=
\operatorname{diag}(u)\,J\,\operatorname{diag}(v).
\]

The scaling vectors \(u\) and \(v\) are chosen so that \(Q\) satisfies the row- and column-sum constraints:
\[
\sum_{j=1}^n u_i J_{ij}v_j=1,
\qquad
\sum_{i=1}^n u_i J_{ij}v_j=1.
\]
Equivalently, they satisfy the fixed-point equations.
\[
u_i
=
\frac{1}{\sum_{j=1}^n J_{ij}v_j},
\qquad
v_j
=
\frac{1}{\sum_{i=1}^n u_i J_{ij}}.
\]
These equations motivate the Sinkhorn updates:
\[
u_i^{(k+1)}
=
\frac{1}{\sum_{j=1}^n J_{ij}v_j^{(k)}},
\qquad
v_j^{(k+1)}
=
\frac{1}{\sum_{i=1}^n u_i^{(k+1)}J_{ij}}.
\]
If \(J\) has strictly positive entries, the Sinkhorn--Knopp theorem guarantees that such positive scaling factors exist and are unique up to a global multiplicative constant. 
Thus, alternating row and column normalization converges to the unique doubly stochastic matrix.
\[
Q^\star
=
\operatorname{diag}(u)\,J\,\operatorname{diag}(v),
\]
which solves the above variational problem.

\begin{algorithm}[t]
\caption{Sinkhorn Scaling / Iterative Proportional Fitting}
\label{alg:sinkhorn}
\begin{algorithmic}[1]
\Require $J\in\mathbb{R}_+^{(t+1)\times (t+1)}$ (ideally $J_{ij}>0$), target marginals $r,c\in\mathbb{R}_+^{t+1}$ with $\sum_i r_i=\sum_j c_j$, tolerance $\varepsilon>0$, max iterations $T$
\Ensure Scaled matrix $Q \approx \mathrm{diag}(u)\,J\,\mathrm{diag}(v)$ such that $Q\mathbf{1}\approx r$ and $Q^\top\mathbf{1}\approx c$
\State $u \gets \mathbf{1}$,\quad $v \gets \mathbf{1}$
\For{$k=1$ \textbf{to} $T$}
    \State $u \gets r \oslash (J v)$ \Comment{elementwise division; enforce row sums}
    \State $v \gets c \oslash (J^\top u)$ \Comment{elementwise division; enforce column sums}
    \State $Q \gets \mathrm{diag}(u)\,J\,\mathrm{diag}(v)$
    \If{$\max\{\|Q\mathbf{1}-r\|_\infty,\ \|Q^\top\mathbf{1}-c\|_\infty\}\le \varepsilon$}
        \State \textbf{break}
    \EndIf
\EndFor
\State \Return $Q$
\end{algorithmic}
\end{algorithm}

\section{Proof of Theorem~\ref{thm:fwer}}
\label{app:fwerproof}
\begin{proof}
By definition of $T$ and $q_{1-\delta}$,
\[
\mathbb P\left(
\forall \lambda\in\Lambda,\;
\ell(\lambda)-\widehat\ell(\lambda)
\le q_{1-\delta}
\right)
\ge 1-\delta.
\]
On this event, for every $\lambda\in\widehat\Lambda_{\mathrm{feas}}$, we have
\[
\ell(\lambda)
\le
\widehat\ell(\lambda)+q_{1-\delta}
\le
\alpha,
\]
where the last inequality follows from the definition
\[
\widehat\Lambda_{\mathrm{feas}}
=
\{\lambda\in\Lambda:
\widehat\ell(\lambda)+q_{1-\delta}\le \alpha\}.
\]
Therefore,
\[
\mathbb P\left(
\forall \lambda\in\widehat\Lambda_{\mathrm{feas}},
\ell(\lambda)\le \alpha
\right)
\ge 1-\delta.
\]
Consequently, any policy selected from
$\widehat\Lambda_{\mathrm{feas}}$ also satisfies the risk constraint with
probability at least $1-\delta$.
\end{proof}

\section{Proof of the worst-case variance}

\subsection{Proof of Proposition~\ref{prop:max_weight_compare}}
\label{app:proof_of_smalle_weight}

\begin{proof}
Since each of the $m$ actions has probability at least $\epsilon$,
every relevant policy probability lies in
$[\epsilon,1-(m-1)\epsilon]$. Hence $M_\epsilon\ge1$,
$\Delta_J\le\log M_\epsilon$, and $\omega_i\le M_\epsilon$
for $i\le t$. The probe ratio $\omega_n=1$ satisfies the same bound,
so $\max_i\omega_i/n\le M_\epsilon/n=a_{\mathrm{IPS}}$.

Let $p_{\max}=\max_i p_i$. By the definition of $\Delta_p$,
$p_i\ge p_{\max}e^{-\Delta_p}$ for every $i$.
Since at least one coordinate equals $p_{\max}$,
\[
1=\sum_{i=1}^n p_i
\ge p_{\max}\bigl[1+(n-1)e^{-\Delta_p}\bigr].
\]
Using $\Delta_p\le\Delta_J\le\log M_\epsilon$ gives
\[
p_{\max}
\le \frac{e^{\Delta_p}}{e^{\Delta_p}+n-1}
\le \frac{M_\epsilon}{M_\epsilon+n-1}
=a_{\mathrm{perm}}.
\]
Finally, $M_\epsilon\ge1$ implies $M_\epsilon+n-1\ge n$,
and therefore $a_{\mathrm{perm}}\le a_{\mathrm{IPS}}$.
\end{proof}

\subsection{Proof of Theorem~\ref{thm:varenv-comparison}}
\label{app:proof_perm_smaller_var}

\begin{proof}
Let $c$ denote either $p$ or $\omega/n$, and put
$c_{\max}=\max_i c_i$. The weights are fixed, and the assumed moment bounds give, almost surely,
\begin{align*}
\operatorname{Var}\!\left(
  \sum_{i=1}^n c_i\widetilde L_i 
\right)
&=\sum_{i=1}^n c_i^2
  \operatorname{Var}(\widetilde L_i)
  +2\sum_{i<j}c_ic_j
  \operatorname{Cov}(\widetilde L_i,\widetilde L_j )\\
&\le \sigma_L^2\left[
  \sum_{i=1}^n c_i^2
  +2C_{\mathrm{mix}}\sum_{i<j}c_ic_j\rho^{j-i}
  \right]\\
&\le \sigma_L^2 c_{\max}^2\left[
  n+2C_{\mathrm{mix}}\sum_{k=1}^{n-1}(n-k)\rho^k
  \right]
=\Gamma_n c_{\max}^2.
\end{align*}
Here there are $n-k$ index pairs at lag $k$.
The constant probe has zero conditional variance and covariance,
so retaining its slot in the bound is valid.

Applying Proposition~\ref{prop:max_weight_compare} to the two
coefficient vectors yields
\[
\operatorname{Var}(\widehat\ell_{\mathrm{perm}} )
\le \Gamma_n a_{\mathrm{perm}}^2=U_{\mathrm{perm}},
\qquad
\operatorname{Var}(\widehat\ell_{\mathrm{IPS}} )
\le \Gamma_n a_{\mathrm{IPS}}^2=U_{\mathrm{IPS}}.
\]
Since $\Gamma_n\ge0$ and
$0\le a_{\mathrm{perm}}\le a_{\mathrm{IPS}}$,
we obtain $U_{\mathrm{perm}}\le U_{\mathrm{IPS}}$.
\end{proof}

\section{Proof of Asymptotic high probability risk control}
\label{app:proof of main thm}
We first list the assumptions. For simplicity, we write $b_i(A|X)$ for the policy at the $i$-th step and $q_j$ for the candidate policy for $\lambda_j$. 

\begin{assumption}
\label{ass:contextual-primitive}
Let $\mathcal X\subseteq\mathbb R^d$ be the context space,
$\mathcal A$ a finite action space, and
$\Lambda_{t+1}=\{\lambda_1,\ldots,\lambda_K\}$
a fixed finite candidate family.
Write $b_i(A|X)$ for the policy at round $i$ and
$q_j$ for the candidate policy indexed by $\lambda_j$.
Assume the following conditions:

\begin{enumerate}

\item \textbf{Stable context and loss distributions.}
Let $\mathcal F_{i-1}$ denote the history available
before round $i$.
Conditional on $\mathcal F_{i-1}$, data are generated
sequentially as
\[
 X_i\sim\nu,\qquad
 A_i\sim b_i(\cdot\mid X_i).\]
The logging policy $b_i$ is
$\mathcal F_{i-1}$-measurable.
The context distribution $\nu$ and the conditional
loss distribution remain fixed, with $0\le L_i\le B$.

\begin{remark} Contexts are i.i.d. and drawn from the same distribution, and the loss distribution remains unchanged for a given context-action pair. The logging policy $b_i$ may depend only on the history
$\mathcal F_{i-1}$. 
\end{remark}
\item \textbf{Finite-dimensional affine policy.}
There exist $\mathcal F_{i-1}$-measurable vectors
$\theta_i\in\mathbb R^r$, feature map
$f:\mathcal X\times\mathcal A\to\mathbb R^r$, where $
 \sup_{x, a}\|f(x, a)\|<\infty$, a reference policy $b(a|x)$, such that the policy at each step $b_i(a|x)$ can be represented as the affine structure of the reference policy:
\[
 b_i(a\mid x)
 =b(a\mid x)\{1+\theta_i^{\mathsf T}f(x,a)\},
 \qquad
 \sum_{a\in\mathcal A}b(a\mid x)f(x,a)=0.
\]
\begin{remark} There exists a reference policy $b(a|x)$, and the policy at each step is an affine transformation of the reference policy. The reference can be interpreted as the 
"averaging" policy that collects the historical data or the policy that the algorithm converges to. We want the policy to change smoothly and converge, so we use this affine structure to capture this pattern and property.
\end{remark}

\item \textbf{Mean-square stabilization.}
There exists a bounded deterministic sequence
$a_i\ge0$ such that, almost surely,
\[
 \|\theta_i\|\le a_i,
 \qquad
 \Delta_t:=\frac1t\sum_{i=1}^{t}a_i^2
 \longrightarrow0.
\]

\begin{remark} We first assume that the $\theta_i$ is bounded, meaning that the deviation of the policy $b_i(A|X)$ to the reference policy $b(A|X)$ is controlled and bounded. $\Delta_t$ converges to $0$ when the policy converges to the reference policy. This assumption bounds the deviation of the logging policy from the reference policy and prevents unbounded, drastic policy changes. 
\end{remark}

\item \textbf{Fixed candidates and uniform positivity.}
The number of candidates $K$ is fixed, and there exists $\varepsilon>0$ such that, almost surely,
\[
 b(a\mid x)\ge\varepsilon,\qquad
 b_i(a\mid x)\ge\varepsilon,\qquad
 q_j(a\mid x)\ge\varepsilon
\]
for all $i,j,x,a$, including possible test-point locations.

\begin{remark} This is a general assumption that every action under every policy is chosen with at least $\epsilon$ probability. 
\item \textbf{Nondegeneracy of the limiting scores.}
For each candidate $\lambda_j$, define
\[
g_j(X,A,L)
:=
\frac{q_j(A\mid X)}{b(A\mid X)}
\bigl(L-\ell_{t+1}(\lambda_j)\bigr).
\]
Let $\mathbb E_b$ denote expectation under
$X\sim\nu$, $A\mid X\sim b(\cdot\mid X)$,
and the fixed conditional loss distribution.
Assume that
\[
\min_{1\le j\le K}
\mathbb E_b\!\left[g_j(X,A,L)^2\right]
>0.
\]
\end{remark}

\end{enumerate}
\end{assumption}

\begin{remark} Assumption \ref{ass:contextual-primitive} states the assumptions of the policy we will use in the following proof. Specifically, the policy depends only on historical data, and it is uniformly positive. The average squared deviation of the logging policies from the reference policy vanishes.  Informally, the policy sequence changes smoothly and converges without extreme or drastic changes. 
\end{remark}

We now present three lemmas and their proofs. Lemma~\ref{lem:wboot-linearization} shows that the risk estimation error can be expanded as the first-order average of each observation's error. Lemma~\ref{lem:wboot-covariance} shows that the covariance of the first-order terms converges to a fixed matrix. Lemma~\ref{lem:wboot-score-consistency} shows that the empirical first-order term used in our bootstrap algorithm converges to the theoretical first-order term in Lemma~\ref{lem:wboot-linearization}. 

\begin{lemma}[First-order expansion of the Sinkhorn risk estimator]
\label{lem:wboot-linearization}
Under the first four conditions of
Assumption~\ref{ass:contextual-primitive}, consider the
Sinkhorn estimator based on $t$ observations and a test point
with assigned loss $\widetilde L_{t+1}=B$, using exact
doubly stochastic scaling. Define a reference policy $b$.

For each $\lambda_j\in\Lambda_{t+1}$, define
\[
 g_j(X,A,L)
 =\frac{q_j(A\mid X)}{b(A\mid X)}
  \bigl(L-\ell_{t+1}(\lambda_j)\bigr),
\]
and, for $i\le t$, let
\[
 d_{i,j}
 =\mathbb E[g_j(X_i,A_i,L_i)\mid\mathcal F_{i-1}],
 \qquad
 \psi_i(\lambda_j)
 =g_j(X_i,A_i,L_i)-d_{i,j}.
\]
These contributions are uniformly bounded and satisfy
\[
 \mathbb E[\psi_i(\lambda_j)\mid\mathcal F_{i-1}]=0.
\]
Moreover,
\begin{equation}
\label{eq:sinkhorn-first-order}
 \widehat\ell_{t+1}(\lambda_j)-\ell_{t+1}(\lambda_j)
 =
 \frac{1}{t+1}\sum_{i=1}^{t}\psi_i(\lambda_j)
 +r_{t+1}(\lambda_j),
\end{equation}
where, for all sufficiently large $t$,
\[
 (t+1)\,
 \mathbb E\!\left[
  \sum_{j=1}^{K}r_{t+1}(\lambda_j)^2
 \right]
 \le C\bigl(\Delta_t+t^{-1}\bigr)
 \longrightarrow0.
\]
Here $C$ is independent of $t$.
The effect of the test point is included in the remainder.
\end{lemma}

\begin{remark}
Lemma~\ref{lem:wboot-linearization} shows that the Sinkhorn
risk estimation error can be represented by the summation of the error of each data point plus an average of negligible remainder. Under the mean-square stabilization assumption, the average
logging policy $\bar b_t = t^{-1}\sum_{i=1}^t b_i$ converges
to the reference policy $b$, which appears in the denominator
of the first-order term. The remaining error is negligible at the
$t^{-1/2}$ scale. This can be interpreted as viewing the historical data collected by an "average" policy, and estimating the risk with an importance-sampling estimator that uses such an "average" policy as the logging policy. Using the "average" policy and the importance-weight estimator induces a negligible error term at the $t^{-1/2}$ scale. We show empirically that the remainder term decreases to zero on a large dataset in Appendix \ref{app: Empirical Evaluation of the Assumption in Theory}.  
\end{remark}

\textbf{Proof Sketch:} The proof begins by writing down the exact weight computed by the permutation and Sinkhorn algorithm. We first show that, replacing the exact weight with Sinkhorn with a simpler form, i.e. $\frac{q(A_k\mid X_k)}{b_{\mathrm{eff}}(A_k\mid X_k)}$, where $b_{\mathrm{eff}}$ is a weighted average of all the logging policies, changes the risk estimate at most at $1/t$ scale. Then, we show that further replacing the weighted-average policy $b_{\mathrm{eff}}$ with the averaging logging policy induces an error of at most $t^{-1/2}$ scale. And finally, by controlling the remainder, we prove the Lemma. 

\begin{proof}
Write $O_i=(X_i,A_i,L_i)$.
All constants below may depend on the fixed policies,
dimensions, feature and loss bounds, and positivity
constant, but not on $t$.

We first identify the centered contributions.
Since $\ell_{t+1}(\lambda_j)=\mathbb E_{q_j}[L]$,
\[
 \mathbb E_b[g_j]=0.
\]
The affine logging condition implies
\[
 \mathbb E[g_j(O_i)\mid\mathcal F_{i-1}]
 =
 \mathbb E_b[(1+\theta_i^{\mathsf T}f)g_j]
 =
 \theta_i^{\mathsf T}h_j,
 \qquad
 h_j:=\mathbb E_b[fg_j].
\]
Consequently,
\[
 \psi_i(\lambda_j)
 =g_j(O_i)-\theta_i^{\mathsf T}h_j
\]
is a martingale difference.
Uniform positivity and bounded losses imply that $g_j$,
its conditional mean, and hence $\psi_i(\lambda_j)$
are uniformly bounded.

\paragraph{A uniform moment bound.}
For a bounded deterministic function $h$, define
\[
 P_i h
 :=\mathbb E[h(O_i)\mid\mathcal F_{i-1}]
 =\mathbb E_b[(1+\theta_i^{\mathsf T}f)h],
 \qquad
 \mathbb M_t h
 :=\frac1t\sum_{i=1}^{t}\{h(O_i)-P_i h\}.
\]
Also write
\[
 \mathbb P_t h:=\frac1t\sum_{i=1}^{t}h(O_i)
\]
for the empirical average over the observed data.

The summands defining $\mathbb M_t h$ are bounded
martingale differences, so
\begin{equation}
\label{eq:linearization-fixed-moment}
 \mathbb E|\mathbb M_t h|^4
 \le \frac{C\|h\|_\infty^4}{t^2}.
\end{equation}
For example, this follows by expanding the fourth power
of the successive partial sums, using the conditional
mean-zero property and the second-moment bound
\[
 \mathbb E\left|
  \sum_{i=1}^{k}\{h(O_i)-P_i h\}
 \right|^2
 \le Ck\|h\|_\infty^2.
\]

Let
\[
 w_j(x,a):=\frac{q_j(a\mid x)}{b(a\mid x)}.
\]
Choose a fixed cube
\[
 \mathcal B=\{z\in\mathbb R^r:\|z\|_\infty\le\rho\}
\]
small enough that
\[
 \rho r\max_s\|f_s\|_\infty<\frac12.
\]
For each function in the finite collection
\[
 \mathcal H
 =\{f_1,\ldots,f_r,w_1,\ldots,w_K,g_1,\ldots,g_K\},
\]
consider
\[
 h_z=\frac{h}{1+z^{\mathsf T}f}.
\]
The series
\[
 h_z=\sum_{k=0}^{\infty}(-1)^k h(z^{\mathsf T}f)^k
\]
converges uniformly on $\mathcal B$.
Expanding each power in coordinates and applying
\eqref{eq:linearization-fixed-moment} and Minkowski's
inequality yields
\[
 \left\|
  \sup_{z\in\mathcal B}|\mathbb M_t h_z|
 \right\|_{L^4}
 \le
 \frac{C\|h\|_\infty}{\sqrt t}
 \sum_{k=0}^{\infty}
 \left(\rho r\max_s\|f_s\|_\infty\right)^k.
\]
The first derivatives with respect to $z$ satisfy the
same order of bound: differentiating the series adds
only a polynomial factor in $k$ to a convergent
geometric series.

Thus there exists a common nonnegative random bound
$U_t$ such that
\[
 \sup_{h\in\mathcal H}\sup_{z\in\mathcal B}
 \left\{
  |\mathbb M_t h_z|
  +\|\nabla_z\mathbb M_t h_z\|
 \right\}
 \le U_t,
 \qquad
 \mathbb E U_t^4\le \frac{C}{t^2},
 \qquad
 \mathbb E U_t^2\le \frac{C}{t}.
\]
All conditional expectations here are first defined
for deterministic $z$.
The resulting uniform bounds allow us to evaluate
these random functions at data-dependent values of $z$.

\paragraph{Step 1: reduce the Sinkhorn estimator to a ratio.}
Fix a candidate $\lambda_j$ and abbreviate
\[
 q=q_j,\qquad
 \ell=\ell_{t+1}(\lambda_j),\qquad
 \widehat\ell=\widehat\ell_{t+1}(\lambda_j),\qquad
 w=q/b,\qquad g=w(L-\ell).
\]
The candidate index is also suppressed for the test point.

The augmented policy matrix is
\[
 C_{ik}=b_i(A_k\mid X_k)\quad(i\le t),
 \qquad
 C_{t+1,k}=q(A_k\mid X_k),
 \qquad k\le t+1.
\]
Write its doubly stochastic scaling as
\[
 P_{ik}=u_iC_{ik}v_k,
 \qquad
 p_k=P_{t+1,k}.
\]
The scaling factors may be normalized so that
\[
 \frac1t\sum_{i=1}^{t}u_i=1.
\]
Every matrix entry lies in $[\varepsilon,1]$.
The row constraints therefore imply
\[
 \frac{u_\ell}{u_i}
 =
 \frac{\sum_k C_{ik}v_k}
      {\sum_k C_{\ell k}v_k}
 \in[\varepsilon,\varepsilon^{-1}].
\]
Averaging over the historical rows gives
\[
 \varepsilon\le u_\ell\le\varepsilon^{-1}
 \qquad(\ell\le t+1).
\]

Define
\[
 b_{\mathrm{eff}}
 :=\frac1t\sum_{i=1}^{t}u_i b_i,
 \qquad
 c:=\frac{u_{t+1}}{t}.
\]
The column constraints give
\begin{equation}
\label{eq:linearization-exact-weight}
 p_k
 =
 \frac{c\,q(A_k\mid X_k)}
      {b_{\mathrm{eff}}(A_k\mid X_k)
       +c\,q(A_k\mid X_k)},
 \qquad
 \frac{\varepsilon}{t}\le c\le\frac{1}{\varepsilon t}.
\end{equation}
Let
\[
 \omega_k
 :=\frac{q(A_k\mid X_k)}
         {b_{\mathrm{eff}}(A_k\mid X_k)},
 \qquad
 \widetilde\omega_k
 :=\frac{\omega_k}{1+c\omega_k}.
\]
Since the target row sums to one,
\[
 p_k
 =\frac{\widetilde\omega_k}
        {\sum_{\ell=1}^{t+1}\widetilde\omega_\ell}.
\]
By the normalization $t^{-1}\sum_{i=1}^t u_i=1$,
$b_{\mathrm{eff}}$ is a convex combination of the logging
policies. Uniform positivity therefore gives
\[
 \varepsilon\le b_{\mathrm{eff}}(a\mid x)\le1,
 \qquad
 \varepsilon\le\omega_k\le\varepsilon^{-1}.
\]
Since $c\le(\varepsilon t)^{-1}$ and $t\ge1$, we also have
\[
 \frac{\varepsilon}{1+\varepsilon^{-2}}
 \le \widetilde\omega_k
 =\frac{\omega_k}{1+c\omega_k}
 \le\varepsilon^{-1}.
\]
Furthermore,
\[
 |\widetilde\omega_k-\omega_k|
 =\frac{c\omega_k^2}{1+c\omega_k}
 \le c\omega_k^2
 \le\frac{\varepsilon^{-3}}{t}.
\]
Finally, the column constraints yield
$p_k=c\widetilde\omega_k$, so
\[
 p_{t+1}
 =\frac{c\omega_{t+1}}{1+c\omega_{t+1}}
 \le c\omega_{t+1}
 \le\frac{\varepsilon^{-2}}{t}.
\]
All these bounds hold uniformly over the data and the
column index, with constants depending only on $\varepsilon$.
Thus, these raw weights are
bounded above and away from zero, and
\[
 |\widetilde\omega_k-\omega_k|\le C/t,
 \qquad
 p_{t+1}\le C/t.
\]
Removing the test-point term from this weighted average
changes its value by at most $C/t$.
Replacing the remaining $\widetilde\omega_k$ by
$\omega_k$ also changes the value by at most $C/t$.

By the affine policy condition,
\[
 b_{\mathrm{eff}}
 =b(1+\beta^{\mathsf T}f),
 \qquad
 \beta:=\frac1t\sum_{i=1}^{t}u_i\theta_i.
\]
It follows that
\begin{equation}
\label{eq:linearization-ratio}
 \widehat\ell
 =\mathcal R_t(\beta)+O(t^{-1}),
 \qquad
 \mathcal R_t(z)
 :=
 \frac{\mathbb P_t[wL/(1+z^{\mathsf T}f)]}
      {\mathbb P_t[w/(1+z^{\mathsf T}f)]}.
\end{equation}
The $O(t^{-1})$ bound is uniform over the data.
The effective policy in this expression is still the
one obtained from the full augmented matrix.

\paragraph{Step 2: replace the effective policy by average logging.}
Put
\[
 \bar\theta:=\frac1t\sum_{i=1}^{t}\theta_i.
\]
The bounds on $u_i$ and the deterministic envelope give
\[
 \|\beta\|+\|\bar\theta\|
 \le C\sqrt{\Delta_t}.
\]
Hence both parameters belong to $\mathcal B$ for all
sufficiently large $t$.
On this cube, $\mathcal R_t$ is uniformly Lipschitz:
its denominator is bounded away from zero, and its
numerator, denominator, and first derivatives are
uniformly bounded.

To compare $\beta$ and $\bar\theta$, write
\[
 f_k=f(X_k,A_k),\qquad
 w_k=w(X_k,A_k),\qquad
 D_k=1+\beta^{\mathsf T}f_k+cw_k,
\]
and define
\[
 A_t^{\mathrm{row}}
 :=\frac1t\sum_{k=1}^{t+1}\frac1{D_k},
 \qquad
 s_t:=\frac1t\sum_{k=1}^{t+1}\frac{f_k}{D_k}.
\]
The column scaling factors satisfy
\[
 v_k=\frac{1}{t\,b(A_k\mid X_k)D_k}.
\]
Thus the historical row constraints imply
\[
 1=u_i\bigl(A_t^{\mathrm{row}}
           +\theta_i^{\mathsf T}s_t\bigr).
\]
The target row constraint gives
\[
 c\sum_{k=1}^{t+1}\frac{w_k}{D_k}=1,
\]
and consequently
\[
 A_t^{\mathrm{row}}+\beta^{\mathsf T}s_t
 =
 \frac{t+1-c\sum_k w_k/D_k}{t}
 =1.
\]
Therefore
\[
 u_i
 =\frac{1}{1+(\theta_i-\beta)^{\mathsf T}s_t}.
\]
Using $\sum_i(u_i-1)=0$, we obtain the exact identity
\begin{align*}
 \beta-\bar\theta
 &=
 \frac1t\sum_{i=1}^{t}(u_i-1)(\theta_i-\beta)\\
 &=-H_t s_t,
 \qquad
 H_t
 :=\frac1t\sum_{i=1}^{t}
 u_i(\theta_i-\beta)(\theta_i-\beta)^{\mathsf T}.
\end{align*}
Moreover,
\[
 \|H_t\|\le C\Delta_t.
\]

The test-point term and the $cw_k$ terms in $s_t$
have total effect $O(t^{-1})$, so
\[
 s_t
 =\mathbb P_t\!\left[
   \frac{f}{1+\beta^{\mathsf T}f}
  \right]+O(t^{-1}).
\]
For a deterministic $z$,
\[
 \frac1t\sum_{i=1}^{t}
 P_i\!\left[\frac{f}{1+z^{\mathsf T}f}\right]
 =
 \mathbb E_b\!\left[
  \frac{(1+\bar\theta^{\mathsf T}f)f}
       {1+z^{\mathsf T}f}
 \right].
\]
Since $\mathbb E_b[f]=0$, evaluating this identity
at $z=\beta$ gives
\[
 s_t
 =
 J_\beta(\bar\theta-\beta)
 +\left.
   \mathbb M_t\!\left[\frac{f}{1+z^{\mathsf T}f}\right]
  \right|_{z=\beta}
 +O(t^{-1}),
 \qquad
 J_\beta
 :=\mathbb E_b\!\left[
    \frac{ff^{\mathsf T}}{1+\beta^{\mathsf T}f}
   \right].
\]
The matrices $J_\beta$ are uniformly bounded.
Combining this identity with $\beta-\bar\theta=-H_t s_t$
and the uniform moment bound yields
\[
 \|\beta-\bar\theta\|
 \le C\Delta_t
       \bigl(\|\beta-\bar\theta\|+U_t+t^{-1}\bigr).
\]
Because $\Delta_t\to0$, the first term on the right
can be absorbed into the left for all sufficiently
large $t$. Hence
\begin{equation}
\label{eq:linearization-beta}
 \|\beta-\bar\theta\|
 \le C\Delta_t(U_t+t^{-1}).
\end{equation}
The Lipschitz property of $\mathcal R_t$ and
\eqref{eq:linearization-ratio} now give
\begin{equation}
\label{eq:linearization-average-ratio}
 \widehat\ell
 =\mathcal R_t(\bar\theta)+e_t,
 \qquad
 |e_t|\le C(\Delta_t U_t+t^{-1}).
\end{equation}

\paragraph{Step 3: expand the ratio and control the remainder.}
Define
\[
 N_t
 :=\mathbb P_t\!\left[
   \frac{g}{1+\bar\theta^{\mathsf T}f}
  \right],
 \qquad
 D_t
 :=\mathbb P_t\!\left[
   \frac{w}{1+\bar\theta^{\mathsf T}f}
  \right].
\]
For any deterministic $z$ and either $h=g$ or $h=w$,
\[
 \frac1t\sum_{i=1}^{t}P_i h_z
 =
 \mathbb E_b\!\left[
  \frac{(1+\bar\theta^{\mathsf T}f)h}
       {1+z^{\mathsf T}f}
 \right].
\]
Evaluating these conditional-mean functions at
$z=\bar\theta$ cancels the numerator and denominator.
Since $\mathbb E_b[g]=0$ and $\mathbb E_b[w]=1$,
\[
 N_t
 =\left.\mathbb M_t g_z\right|_{z=\bar\theta},
 \qquad
 D_t
 =1+\left.\mathbb M_t w_z\right|_{z=\bar\theta}.
\]
Here the substitution is made after the functions of
deterministic $z$ have been formed; it does not treat
$\bar\theta$ as predictable at every round.

Let
\[
 S_t:=\mathbb M_t g
     =\frac1t\sum_{i=1}^{t}\psi_i(\lambda_j),
 \qquad
 a_t:=N_t-S_t.
\]
The uniform derivative bound and
$\|\bar\theta\|\le\sqrt{\Delta_t}$ imply
\[
 |a_t|\le C\sqrt{\Delta_t}\,U_t,
 \qquad
 |S_t|+|D_t-1|\le C U_t.
\]
Also, positivity of the weights gives $D_t\ge c_0>0$.

By \eqref{eq:linearization-average-ratio},
\begin{align*}
 \widehat\ell-\ell
 &=\frac{N_t}{D_t}+e_t\\
 &=S_t+\widetilde r_t,
 \qquad
 \widetilde r_t
 =\frac{a_t}{D_t}
  -\frac{S_t(D_t-1)}{D_t}
  +e_t.
\end{align*}
For all sufficiently large $t$, $\Delta_t\le1$, and thus
\[
 |\widetilde r_t|
 \le C\left(
   \sqrt{\Delta_t}\,U_t+U_t^2+t^{-1}
  \right).
\]
Using the second- and fourth-moment bounds on $U_t$,
\begin{align*}
 t\,\mathbb E|\widetilde r_t|^2
 &\le
 C\left(
   t\Delta_t\,\mathbb E U_t^2
   +t\,\mathbb E U_t^4+t^{-1}
  \right)\\
 &\le C(\Delta_t+t^{-1})
 \longrightarrow0.
\end{align*}

Finally, the desired normalization is $t+1$.
Since
\[
 S_t
 =\frac{1}{t+1}\sum_{i=1}^{t}\psi_i(\lambda_j)
  +\frac{S_t}{t+1},
\]
define
\[
 r_{t+1}(\lambda_j)
 :=\widetilde r_t+\frac{S_t}{t+1}.
\]
Martingale orthogonality and boundedness imply
$\mathbb E S_t^2\le C/t$, so
\[
 (t+1)\,
 \mathbb E\left|\frac{S_t}{t+1}\right|^2
 \le C/t^2.
\]
Combining the bounds proves
\[
 (t+1)\mathbb E[r_{t+1}(\lambda_j)^2]
 \le C(\Delta_t+t^{-1}).
\]
Summing over the fixed number $K$ of candidates
establishes the stated vector remainder bound.
\end{proof}

\begin{lemma}[Stabilization of the conditional covariance]
\label{lem:wboot-covariance}
Under Assumption~\ref{ass:contextual-primitive}, let
\[
 \psi_i
 =\bigl(
   \psi_i(\lambda_1),\ldots,\psi_i(\lambda_K)
  \bigr)^{\mathsf T}
\]
be the contributions defined in
Lemma~\ref{lem:wboot-linearization}.
Define
\[
 V_{jk}:=\mathbb E_b[g_jg_k],
 \qquad
 V_{t+1}
 :=\frac{1}{t+1}\sum_{i=1}^{t}
   \mathbb E[\psi_i\psi_i^{\mathsf T}
             \mid\mathcal F_{i-1}].
\]
Then $V$ is a deterministic positive semidefinite matrix
with $V_{jj}>0$ for every $j$, and
\[
 \|V_{t+1}-V\|
 \le C\bigl(\sqrt{\Delta_t}+\Delta_t+t^{-1}\bigr)
 \longrightarrow0
 \qquad\text{almost surely}.
\]
In particular, $V_{t+1}\xrightarrow{p}V$.
Here $\|\cdot\|$ denotes the operator norm and $C$
is independent of $t$.
\end{lemma}

\begin{remark}
Lemma~\ref{lem:wboot-covariance} shows that the average covariance of the first-order terms described in Lemma~\ref{lem:wboot-linearization} converges
to a fixed matrix $V$, even though the policy changes
over time. We empirically show that the covariance discrepancy decreases to zero with a large dataset in Appendix \ref{app: Empirical Evaluation of the Assumption in Theory}.  
\end{remark}

\textbf{Proof sketch:}
We can bound the difference of the $V$, from reference policy, and $V_{t+1}$, from data, by
$C(\|\theta_i\|+\|\theta_i\|^2)$, which is how far the policy at each step deviates from the reference policy. 
Although the policy at each step differs from the reference policy, the covariance difference is bounded by how far the policy deviates from the reference policy. Then, by averaging over all steps and using the mean-square stabilization assumption that bounds the deviation of the policy sequence, we obtain an entry-wise bound on the covariance.

\begin{proof}
Recall that
\[
 h_j=\mathbb E_b[fg_j],
 \qquad
 d_{i,j}=\theta_i^{\mathsf T}h_j,
 \qquad
 \psi_i(\lambda_j)=g_j(O_i)-d_{i,j},
\]
and
\[
 \mathbb E[g_j(O_i)\mid\mathcal F_{i-1}]=d_{i,j}.
\]
Therefore, for every pair of candidates $j,k$,
\begin{align*}
 &\mathbb E[
   \psi_i(\lambda_j)\psi_i(\lambda_k)
   \mid\mathcal F_{i-1}]\\
 &\qquad=
 \mathbb E[g_j(O_i)g_k(O_i)\mid\mathcal F_{i-1}]
 -d_{i,j}d_{i,k}.
\end{align*}

By the affine logging condition and the fixed context
and conditional loss distributions,
\[
 \mathbb E[g_j(O_i)g_k(O_i)\mid\mathcal F_{i-1}]
 =
 \mathbb E_b[(1+\theta_i^{\mathsf T}f)g_jg_k].
\]
Consequently,
\begin{align*}
 &\mathbb E[
   \psi_i(\lambda_j)\psi_i(\lambda_k)
   \mid\mathcal F_{i-1}]
 -V_{jk}\\
 &\qquad=
 \theta_i^{\mathsf T}\mathbb E_b[fg_jg_k]
 -(\theta_i^{\mathsf T}h_j)
  (\theta_i^{\mathsf T}h_k).
\end{align*}
The functions $f,g_j,g_k$ are uniformly bounded.
Hence
\[
 \left|
  \mathbb E[
   \psi_i(\lambda_j)\psi_i(\lambda_k)
   \mid\mathcal F_{i-1}]
  -V_{jk}
 \right|
 \le C\bigl(\|\theta_i\|+\|\theta_i\|^2\bigr).
\]

The stabilization condition gives
\[
 \frac1t\sum_{i=1}^{t}\|\theta_i\|^2
 \le\Delta_t,
 \qquad
 \frac1t\sum_{i=1}^{t}\|\theta_i\|
 \le
 \left(\frac1t\sum_{i=1}^{t}\|\theta_i\|^2\right)^{1/2}
 \le\sqrt{\Delta_t}.
\]
Thus
\[
 \left|
  \frac1t\sum_{i=1}^{t}
  \mathbb E[
   \psi_i(\lambda_j)\psi_i(\lambda_k)
   \mid\mathcal F_{i-1}]
  -V_{jk}
 \right|
 \le C(\sqrt{\Delta_t}+\Delta_t).
\]
Since $K$ is fixed, the same bound holds in operator
norm for the corresponding matrices.

Finally, changing the normalization from $t$ to $t+1$
gives
\[
 V_{t+1}-V
 =
 \frac{t}{t+1}
 \left\{
  \frac1t\sum_{i=1}^{t}
  \mathbb E[\psi_i\psi_i^{\mathsf T}
            \mid\mathcal F_{i-1}]
  -V
 \right\}
 -\frac{V}{t+1}.
\]
It follows that
\[
 \|V_{t+1}-V\|
 \le C(\sqrt{\Delta_t}+\Delta_t+t^{-1})
 \longrightarrow0
 \qquad\text{almost surely}.
\]

The matrix $V$ is deterministic because the reference
policy, candidate policies, and data-generating
distributions are fixed.
For every $a\in\mathbb R^K$,
\[
 a^{\mathsf T}Va
 =
 \mathbb E_b\!\left[
  \left(\sum_{j=1}^{K}a_jg_j\right)^2
 \right]\ge0.
\]
Thus $V$ is positive semidefinite.
Its positive diagonal follows from the nondegeneracy
condition in Assumption~\ref{ass:contextual-primitive}.
\end{proof}

\begin{lemma}[Consistency of the bootstrap scores]
\label{lem:wboot-score-consistency}
Under the first four conditions of
Assumption~\ref{ass:contextual-primitive}, define
\[
 \widehat\psi_i(\lambda_j)
 :=(t+1)\widehat\phi_i(\lambda_j)
 =(t+1)p_i(\lambda_j)
   \bigl(\widetilde L_i-\widehat\ell_{t+1}(\lambda_j)\bigr),
\]
where $\widetilde L_i=L_i$ for $i\le t$ and
$\widetilde L_{t+1}=B$ for the test point.
Let
\[
 \widehat\psi_i
 =\bigl(
   \widehat\psi_i(\lambda_1),\ldots,
   \widehat\psi_i(\lambda_K)
  \bigr)^{\mathsf T},
\]
and let $\psi_i$ be the real-data contributions from
Lemma~\ref{lem:wboot-linearization}.

Assigning zero ideal contribution to the test point,
define
\[
 e_{t+1}^{\,2}
 :=
 \frac{1}{t+1}\sum_{i=1}^{t}
   \|\widehat\psi_i-\psi_i\|^2
 +\frac{1}{t+1}\|\widehat\psi_{t+1}\|^2.
\]
Then, for all sufficiently large $t$,
\[
 \mathbb E[e_{t+1}^{\,2}]
 \le C(\Delta_t+t^{-1})
 \longrightarrow0.
\]
In particular, $e_{t+1}^{\,2}\xrightarrow{p}0$.
\end{lemma}

\begin{remark}
In practice, we do not explicitly have the reference policy $b$. So instead, we use $\widehat{\psi}_i(\lambda_j)
= (t+1)p_i(\lambda_j)
  \bigl(\widetilde L_i-\widehat\ell_{t+1}(\lambda_j)\bigr)$ to approximate the first-order linear expansion of the error.  
Lemma~\ref{lem:wboot-score-consistency} shows that 
the implementation approximates the
true first-order error contributions in average squared
error. We show empirically that the average squared discrepancy between the implemented and ideal scores decreases to zero on a large dataset in Appendix \ref{app: Empirical Evaluation of the Assumption in Theory}. This lemma connects the theoretical first-order expansion with our implementation.
\end{remark}

\textbf{Proof Sketch:} The implementation and the theoretical first-order term differ in weight and in the base risk estimate. For the weight, the implementation uses the Sinkhorn weight, while the theoretical first-order term uses the importance weight with the "average" logging policy. We first show that the error is negligible as $t$ increases. For the base risk estimate, where the theoretical first-order term uses the true risk while the implementation uses the estimated risk, we show that the risk estimation error is also negligible. Combining these two terms, we show that the averaging error between $\widehat\psi_i$ and $\psi_i$ vanishes.

\begin{proof}
Fix a candidate $\lambda_j$ and abbreviate
\[
 q=q_j,\qquad
 \ell=\ell_{t+1}(\lambda_j),\qquad
 \widehat\ell=\widehat\ell_{t+1}(\lambda_j),\qquad
 w_i=\frac{q(A_i\mid X_i)}{b(A_i\mid X_i)}.
\]
All constants below are independent of $t$.

\paragraph{Step 1: the normalized weights approach the reference ratios.}
The proof of Lemma~\ref{lem:wboot-linearization} gives
\[
 p_i
 =\frac{\widetilde\omega_i}
        {\sum_{k=1}^{t+1}\widetilde\omega_k},
 \qquad
 \widetilde\omega_i
 =\frac{\omega_i}{1+c\omega_i},
 \qquad
 \omega_i
 =\frac{q(A_i\mid X_i)}
        {b_{\mathrm{eff}}(A_i\mid X_i)},
\]
where
\[
 b_{\mathrm{eff}}=b(1+\beta^{\mathsf T}f),
 \qquad
 \|\beta\|\le C\sqrt{\Delta_t},
 \qquad
 0<c\le C/t.
\]
Uniform positivity and boundedness of $f$ imply
\[
 \max_{i\le t+1}|\omega_i-w_i|
 \le C\sqrt{\Delta_t},
 \qquad
 \max_{i\le t+1}|\widetilde\omega_i-\omega_i|
 \le C/t.
\]
Consequently,
\begin{equation}
\label{eq:score-raw-weight}
 \max_{i\le t+1}|\widetilde\omega_i-w_i|
 \le C(\sqrt{\Delta_t}+t^{-1}).
\end{equation}
The quantities $w_i$ and $\widetilde\omega_i$ are
uniformly bounded above and away from zero.

For the real observations, define
\[
 M_t
 :=\frac1t\sum_{i=1}^{t}
   \bigl\{w_i-\mathbb E[w_i\mid\mathcal F_{i-1}]\bigr\}.
\]
The summands are bounded martingale differences.
Thus martingale orthogonality gives
\[
 \mathbb E M_t^2\le C/t.
\]
Writing $w(X,A)=q(A\mid X)/b(A\mid X)$, the affine
logging condition implies
\[
 \mathbb E[w_i\mid\mathcal F_{i-1}]
 =\mathbb E_b[(1+\theta_i^{\mathsf T}f)w]
 =1+\theta_i^{\mathsf T}\mathbb E_b[fw].
\]
Therefore, with $\bar\theta=t^{-1}\sum_{i=1}^{t}\theta_i$,
\[
 \frac1t\sum_{i=1}^{t}w_i
 =1+\bar\theta^{\mathsf T}\mathbb E_b[fw]+M_t,
 \qquad
 \|\bar\theta\|\le\sqrt{\Delta_t}.
\]

Let
\[
 \overline W_t
 :=\frac{1}{t+1}\sum_{i=1}^{t+1}\widetilde\omega_i.
\]
Using \eqref{eq:score-raw-weight}, the preceding identity,
and the boundedness of the test-point weight gives
\[
 |\overline W_t-1|
 \le C(\sqrt{\Delta_t}+|M_t|+t^{-1}).
\]
Moreover, $\overline W_t$ is bounded away from zero.
Since
\[
 (t+1)p_i=\frac{\widetilde\omega_i}{\overline W_t},
\]
we obtain
\[
 \max_{i\le t+1}|(t+1)p_i-w_i|
 \le C(\sqrt{\Delta_t}+|M_t|+t^{-1}).
\]
In particular,
\begin{equation}
\label{eq:score-normalized-weight}
 \mathbb E\!\left[
  \max_{i\le t+1}|(t+1)p_i-w_i|^2
 \right]
 \le C(\Delta_t+t^{-1}),
 \qquad
 \max_{i\le t+1}(t+1)p_i\le C.
\end{equation}

\paragraph{Step 2: centering by the estimated risk is negligible.}
Lemma~\ref{lem:wboot-linearization} gives
\[
 \widehat\ell-\ell
 =\frac{1}{t+1}\sum_{i=1}^{t}\psi_i(\lambda_j)
  +r_{t+1}(\lambda_j).
\]
The contributions are bounded martingale differences,
so
\[
 \mathbb E\left[
  \left(\sum_{i=1}^{t}\psi_i(\lambda_j)\right)^2
 \right]\le Ct.
\]
Together with the remainder bound and boundedness of
$\Delta_t$, this yields
\begin{equation}
\label{eq:score-risk-mse}
 \mathbb E|\widehat\ell-\ell|^2\le C/t.
\end{equation}

\paragraph{Step 3: compare the implemented and true contributions.}
For $i\le t$, recall that
\[
 \psi_i(\lambda_j)
 =w_i(L_i-\ell)-d_{i,j},
 \qquad
 d_{i,j}=\theta_i^{\mathsf T}h_j,
 \qquad
 h_j=\mathbb E_b[fg_j].
\]
The difference between the two contributions has the
exact decomposition
\begin{align*}
 \widehat\psi_i(\lambda_j)-\psi_i(\lambda_j)
 &=
 \bigl((t+1)p_i-w_i\bigr)(L_i-\ell)\\
 &\quad +(t+1)p_i(\ell-\widehat\ell)
       +d_{i,j}.
\end{align*}
The first term is controlled by
\eqref{eq:score-normalized-weight}, the second by
\eqref{eq:score-risk-mse}, and the third satisfies
\[
 \frac1t\sum_{i=1}^{t}d_{i,j}^2
 \le C\frac1t\sum_{i=1}^{t}\|\theta_i\|^2
 \le C\Delta_t.
\]
Using bounded losses and $(t+1)p_i\le C$, we conclude
\[
 \mathbb E\!\left[
  \frac{1}{t+1}\sum_{i=1}^{t}
  |\widehat\psi_i(\lambda_j)-\psi_i(\lambda_j)|^2
 \right]
 \le C(\Delta_t+t^{-1}).
\]

For the test point, $\widetilde L_{t+1}=B$ and
$0\le\widehat\ell\le B$, so
\[
 |\widehat\psi_{t+1}(\lambda_j)|
 =(t+1)p_{t+1}|B-\widehat\ell|
 \le C.
\]
Its contribution to the average squared discrepancy
is therefore at most $C/(t+1)$.

Summing over the fixed number $K$ of candidates gives
\[
 \mathbb E[e_{t+1}^{\,2}]
 \le C(\Delta_t+t^{-1})
 \longrightarrow0.
\]
Convergence in probability follows from Markov's
inequality.
\end{proof}

\subsection{Proof of Theorem \ref{thm:main_boot}}

We prove Theorem~\ref{thm:main_boot} by combining the three
lemmas. Lemma~\ref{lem:wboot-linearization} writes the true
estimation error as an average of mean-zero contributions plus a negligible remainder. Together with the covariance
stabilization in Lemma~\ref{lem:wboot-covariance}, we then show that the estimation error of each candidate policy induces a Gaussian limit. 
Lemma~\ref{lem:wboot-score-consistency} then ensures that
the bootstrap score construction recovers the same limiting covariance, demonstrating the effectiveness of the bootstrap score in approximating the estimation error. Finally, as the distribution of the bootstrap estimation matches the true error distribution, the
maximum-error distribution matches, yielding consistent bootstrap
quantiles.

\begin{proof}[Proof of Theorem~\ref{thm:main_boot}]
Throughout the proof, vectors collect the coordinates
indexed by $\lambda_1,\ldots,\lambda_K$.
Write
\[
 \boldsymbol\ell_{t+1}
 =\bigl(\ell_{t+1}(\lambda_j)\bigr)_{j=1}^{K},
 \qquad
 \widehat{\boldsymbol\ell}_{t+1}
 =\bigl(\widehat\ell_{t+1}(\lambda_j)\bigr)_{j=1}^{K}.
\]

\paragraph{Step 1: the true error has a Gaussian limit.}
By Lemma~\ref{lem:wboot-linearization},
\[
 \sqrt{t+1}\,
 \bigl(\widehat{\boldsymbol\ell}_{t+1}
       -\boldsymbol\ell_{t+1}\bigr)
 =
 S_t+\sqrt{t+1}\,r_{t+1},
 \qquad
 S_t:=\frac{1}{\sqrt{t+1}}\sum_{i=1}^{t}\psi_i,
\]
where
\[
 \mathbb E\|\sqrt{t+1}\,r_{t+1}\|^2\longrightarrow0.
\]
The vectors $\psi_i$ are uniformly bounded martingale
differences. In particular, for every $\eta>0$,
\[
 \frac{1}{t+1}\sum_{i=1}^{t}
 \mathbb E\!\left[
  \|\psi_i\|^2
  \mathbf 1\{\|\psi_i\|>\eta\sqrt{t+1}\}
  \,\middle|\,\mathcal F_{i-1}
 \right]=0
\]
for all sufficiently large $t$.
Thus the conditional Lindeberg condition holds.

Lemma~\ref{lem:wboot-covariance} gives
\[
 V_{t+1}
 :=
 \frac{1}{t+1}\sum_{i=1}^{t}
 \mathbb E[\psi_i\psi_i^{\mathsf T}
           \mid\mathcal F_{i-1}]
 \xrightarrow{p}V.
\]
The martingale central limit theorem therefore yields
\[
 S_t\Rightarrow Z,
 \qquad Z\sim N(0,V).
\]
The remainder converges to zero in probability after
scaling by $\sqrt{t+1}$.
Since $-Z$ has the same distribution as $Z$,
\begin{equation}
\label{eq:bootstrap-proof-true-limit}
 \sqrt{t+1}\,
 \bigl(\boldsymbol\ell_{t+1}
       -\widehat{\boldsymbol\ell}_{t+1}\bigr)
 \Rightarrow N(0,V).
\end{equation}

The $L^2$ remainder also implies convergence of the
actual error covariance.
Indeed, martingale orthogonality gives
\[
 \mathbb E S_t=0,
 \qquad
 \mathbb E[S_tS_t^{\mathsf T}]
 =\mathbb E V_{t+1}\longrightarrow V,
\]
where convergence of expectations follows from
uniform boundedness of $V_{t+1}$.
The remainder covariance and its cross terms with
$S_t$ vanish by Cauchy--Schwarz. Consequently,
\begin{equation}
\label{eq:bootstrap-proof-true-covariance}
 (t+1)\operatorname{Cov}
 \bigl(\boldsymbol\ell_{t+1}
       -\widehat{\boldsymbol\ell}_{t+1}\bigr)
 \longrightarrow V.
\end{equation}

\paragraph{Step 2: the bootstrap estimates the same covariance.}
Define
\[
 \widetilde V_{t+1}
 :=\frac{1}{t+1}\sum_{i=1}^{t}
   \psi_i\psi_i^{\mathsf T}.
\]
Each entry of
\[
 \psi_i\psi_i^{\mathsf T}
 -
 \mathbb E[\psi_i\psi_i^{\mathsf T}
           \mid\mathcal F_{i-1}]
\]
is a bounded martingale difference.
Since $K$ is fixed, martingale orthogonality gives
\[
 \mathbb E\|
  \widetilde V_{t+1}-V_{t+1}
 \|_F^2\le C/t.
\]
Together with Lemma~\ref{lem:wboot-covariance}, this implies
\[
 \widetilde V_{t+1}\xrightarrow{p}V.
\]

Now define the covariance estimate from the actual
bootstrap scores:
\[
 \widehat V_{t+1}
 :=\frac{1}{t+1}\sum_{i=1}^{t+1}
   \widehat\psi_i\widehat\psi_i^{\mathsf T}
 =(t+1)\sum_{i=1}^{t+1}
   \widehat\phi_i\widehat\phi_i^{\mathsf T},
\]
where $\widehat\psi_i=(t+1)\widehat\phi_i$.
Let $e_{t+1}$ be the square root of the discrepancy
defined in Lemma~\ref{lem:wboot-score-consistency}.
Expanding the outer products and applying
Cauchy--Schwarz yields
\[
 \|\widehat V_{t+1}-\widetilde V_{t+1}\|
 \le 2C e_{t+1}+e_{t+1}^{\,2}
 \xrightarrow{p}0.
\]
The test-point term is included in $e_{t+1}^{\,2}$
through its zero ideal contribution.
It follows that
\begin{equation}
\label{eq:bootstrap-proof-estimated-covariance}
 \widehat V_{t+1}\xrightarrow{p}V.
\end{equation}

Conditional on $\mathcal D_t$, the scores are fixed
and the multipliers are independent standard Gaussian
variables. Hence the bootstrap vector satisfies the
exact conditional identity
\begin{equation}
\label{eq:bootstrap-proof-conditional-gaussian}
 \sqrt{t+1}\,
 \widehat{\boldsymbol G}_{t+1}^{*}
 \mid\mathcal D_t
 \sim N(0,\widehat V_{t+1}),
\end{equation}
where
\[
 \widehat{\boldsymbol G}_{t+1}^{*}
 :=\bigl(
   \widehat G_{t+1}^{*}(\lambda_1),\ldots,
   \widehat G_{t+1}^{*}(\lambda_K)
  \bigr)^{\mathsf T}.
\]
Equations
\eqref{eq:bootstrap-proof-true-covariance}--%
\eqref{eq:bootstrap-proof-conditional-gaussian}
also establish asymptotic matching between the
conditional bootstrap covariance and the actual
error covariance.

\paragraph{Step 3: the distributions of the maxima are close.}
Let
\[
 W:=\max_{j\le K}Z_j,
 \qquad Z\sim N(0,V),
 \qquad F(x):=\mathbb P(W\le x).
\]
By Lemma~\ref{lem:wboot-covariance}, $V_{jj}>0$
for every $j$, so each $Z_j$ has a continuous distribution.
For every $x\in\mathbb R$,
\[
 \mathbb P(W=x)
 \le\sum_{j=1}^{K}\mathbb P(Z_j=x)=0.
\]
Thus $F$ is continuous; independence between the
coordinates is not required.

The maximum is a continuous function.
Applying the continuous mapping theorem to
\eqref{eq:bootstrap-proof-true-limit} gives
\[
 \sqrt{t+1}\,T_{t+1}\Rightarrow W.
\]
Because the limiting distribution function is continuous,
this convergence is uniform at the level of
distribution functions:
\begin{equation}
\label{eq:bootstrap-proof-true-cdf}
 \sup_{x\in\mathbb R}
 \left|
  \mathbb P(\sqrt{t+1}\,T_{t+1}\le x)-F(x)
 \right|
 \longrightarrow0.
\end{equation}

To handle the conditional bootstrap distribution,
take $U\sim N(0,I_K)$ independently of $\mathcal D_t$
and couple the two Gaussian vectors as
\[
 \widehat V_{t+1}^{1/2}U
 \quad\text{and}\quad
 V^{1/2}U.
\]
The positive semidefinite square-root map is continuous,
including at singular matrices. Therefore,
\[
 \|\widehat V_{t+1}^{1/2}-V^{1/2}\|_F
 \xrightarrow{p}0.
\]
Define the coupled maxima
\[
 \widehat W_t
 :=\max_j(\widehat V_{t+1}^{1/2}U)_j,
 \qquad
 W_0:=\max_j(V^{1/2}U)_j.
\]
Conditional on $\mathcal D_t$, $\widehat W_t$ has the
same distribution as $\sqrt{t+1}\,T_{t+1}^{*}$,
while $W_0$ has distribution function $F$.

Using
\[
 |\max_j x_j-\max_j y_j|
 \le\|x-y\|_2
\]
and conditional Markov's inequality, for any $\eta>0$,
\[
 \mathbb P(
  |\widehat W_t-W_0|>\eta
  \mid\mathcal D_t)
 \le
 \frac{
  \|\widehat V_{t+1}^{1/2}-V^{1/2}\|_F^2
 }{\eta^2}.
\]
A distribution-function sandwich then gives
\begin{align*}
 &\sup_x
 \left|
  \mathbb P(
   \sqrt{t+1}\,T_{t+1}^{*}\le x
   \mid\mathcal D_t)
  -F(x)
 \right|\\
 &\qquad\le
 \frac{
  \|\widehat V_{t+1}^{1/2}-V^{1/2}\|_F^2
 }{\eta^2}
 +\sup_x\{F(x+\eta)-F(x-\eta)\}.
\end{align*}
For fixed $\eta$, the first term converges to zero
in probability as $t\to\infty$.
The second tends to zero as $\eta\downarrow0$,
because a continuous distribution function is
uniformly continuous.
Consequently,
\begin{equation}
\label{eq:bootstrap-proof-bootstrap-cdf}
 \sup_x
 \left|
  \mathbb P(
   \sqrt{t+1}\,T_{t+1}^{*}\le x
   \mid\mathcal D_t)
  -F(x)
 \right|
 \xrightarrow{p}0.
\end{equation}

Combining
\eqref{eq:bootstrap-proof-true-cdf} and
\eqref{eq:bootstrap-proof-bootstrap-cdf}
by the triangle inequality, and changing variables
$x=\sqrt{t+1}\,u$, proves
\[
 \sup_{u\in\mathbb R}
 \left|
  \mathbb P(T_{t+1}\le u)
  -\mathbb P(T_{t+1}^{*}\le u\mid\mathcal D_t)
 \right|
 \xrightarrow{p}0.
\]

\paragraph{Step 4: quantile consistency gives simultaneous coverage.}
We first verify that the limiting quantile is unique.
The Gaussian vector $Z$ has positive density relative
to its support subspace
\[
 S=\operatorname{range}(V^{1/2}).
\]
The image of $S$ under $z\mapsto\max_j z_j$
is either $\mathbb R$ or $[0,\infty)$:
a positive value exists, positive scaling gives
every positive value, and if a negative value exists,
scaling also gives every negative value.
Every open interval intersecting the interior of
this image has a nonempty relatively open preimage
in $S$, and therefore positive Gaussian probability.
Thus $F$ is strictly increasing on the interior
of its support.

For fixed $\delta\in(0,1)$, let
\[
 c:=F^{-1}(1-\delta).
\]
Continuity and strict increase imply
\[
 F(c)=1-\delta,
 \qquad
 F(c-\eta)<1-\delta<F(c+\eta)
 \quad\text{for every }\eta>0.
\]
Define the scaled bootstrap critical value
\[
 Q_t:=\sqrt{t+1}\,q_{1-\delta,t+1}^{*}.
\]
Uniform conditional distribution-function convergence
in \eqref{eq:bootstrap-proof-bootstrap-cdf}
implies
\[
 Q_t\xrightarrow{p}c.
\]
Indeed, with probability tending to one, the
conditional bootstrap distribution function is
below $1-\delta$ at $c-\eta$ and above $1-\delta$
at $c+\eta$, placing its quantile between these points.

The critical value and the true error may depend
on the same data. To obtain coverage without an
independence assumption, use the inequalities
\begin{align*}
 &\mathbb P(
  \sqrt{t+1}\,T_{t+1}\le c-\eta)
  -\mathbb P(|Q_t-c|>\eta)\\
 &\qquad\le
 \mathbb P(T_{t+1}\le q_{1-\delta,t+1}^{*})\\
 &\qquad\le
 \mathbb P(
  \sqrt{t+1}\,T_{t+1}\le c+\eta)
  +\mathbb P(|Q_t-c|>\eta).
\end{align*}
Letting $t\to\infty$ and then $\eta\downarrow0$ gives
\[
 \mathbb P(T_{t+1}\le q_{1-\delta,t+1}^{*})
 \longrightarrow F(c)=1-\delta.
\]
Finally, by the definition of $T_{t+1}$,
\[
 \{T_{t+1}\le q_{1-\delta,t+1}^{*}\}
 =
 \left\{
  \ell_{t+1}(\lambda)
  \le\widehat\ell_{t+1}(\lambda)
     +q_{1-\delta,t+1}^{*}
  \text{ for all }\lambda\in\Lambda_{t+1}
 \right\}.
\]
This proves the simultaneous coverage statement.
\end{proof}

\textbf{Now we prove the asymptotic high-probability risk control in Corollary~\ref{cor:main_fwer}.}
\begin{proof}
Let \(q_{1-\delta,t+1}^*\) be the conditional \((1-\delta)\)-quantile of
\(T_{t+1}^*\):
\[
q_{1-\delta,t+1}^*
:=
\inf
\left\{
q\in\mathbb R:
\mathbb P(T_{t+1}^*\le q\mid Z_{1:t+1})
\ge
1-\delta
\right\}.
\]
By the bootstrap consistency above and the continuity of the limiting Gaussian
maximum at the \((1-\delta)\)-quantile,
\[
\mathbb P(T_{t+1}\le q_{1-\delta,t+1}^*)
\to
1-\delta.
\]
By definition of \(T_{t+1}\), the event
\[
T_{t+1}\le q_{1-\delta,t+1}^*
\]
is equivalent to
\[
\max_{\lambda\in\Lambda_{t+1}}
\bigl(
\ell_{t+1}(\lambda)-\widehat\ell_{t+1}(\lambda)
\bigr)
\le
q_{1-\delta,t+1}^*.
\]
Equivalently,
\[
\ell_{t+1}(\lambda)
\le
\widehat\ell_{t+1}(\lambda)
+
q_{1-\delta,t+1}^*,
\qquad
\forall \lambda\in\Lambda_{t+1}.
\]
Therefore,
\[
\mathbb P\!\left(
\ell_{t+1}(\lambda)
\le
\widehat\ell_{t+1}(\lambda)
+
q_{1-\delta,t+1}^*
\ \text{for all }\lambda\in\Lambda_{t+1}
\right)
\to
1-\delta.
\]
This completes the proof.
\end{proof}

\section{Experiment Details and Additional Experiments}

\subsection{Datasets and Experimental Setup}
\label{app: dataset}
\subsubsection{Breast cancer diagnosis.}
\label{app: daatset_breast}
We convert the breast cancer classification dataset into a binary contextual bandit problem following \cite{yang2023distributionally, guo2024distributionally}. 
Each feature vector is treated as a context $x$, and the action space is $\mathcal A=\{0,1\}$, 
where $a=1$ denotes predicting malignant and $a=0$ denotes predicting benign. 
The reward is the classification accuracy reward $r(x,a,y)=\mathbf 1\{a=y\}$.
The risk of interest is the false-negative rate $R(\pi)=\Pr_{\pi}(a=0\mid y=1)$,
and the goal is to select a policy at each round satisfying
\[
R(\pi)\le \alpha.
\]

The base policy is a neural network classifier with logits $f_\theta(x)\in\mathbb R^2$. 
The control parameter is a positive-class bias $\lambda=b$, and the controlled policy is
\[
\pi_{\theta,\lambda}(a\mid x)
=
\mathrm{softmax}\!\left(
f_\theta(x)+
\begin{bmatrix}
0\\ \lambda
\end{bmatrix}
\right)_a .
\]
Larger $b$ makes the policy more likely to select the positive class, thereby reducing false negatives but potentially increasing false positives. The $\pi_\theta =\mathrm{softmax}(f_\theta(x))$ is optimized every round with importance weights. Here, the FNR risk function is monotonic in $\lambda$. We then train the policy with a policy gradient using self-normalized importance sampling. We use accuracy as the reward. 

\subsubsection{Stock investment.}
\label{app: daatset_stock}
We also evaluate on a six-action stock investment bandit. 
Each context $x$ contains market features, and each action $a\in\{1,\ldots,6\}$ corresponds to selecting one asset. 
The realized reward is the normalized profit of the selected asset. 
The risk is the probability of selecting the worst-performing action,
\[
L(\pi)
=
\Pr_{\pi}\!\left(a=\arg\min_{a'\in\mathcal A} r(x,a')\right).
\]
The goal is to select a policy whose worst-action selection risk is below a target level $\alpha$.

The policy is parameterized by a neural network: $\pi_{\theta,\lambda}(a|x) = \mathrm{softmax}(\hat{r}_\theta(x,a) - \lambda \hat{l}(x,a))$, where $\hat{r}$ and $\hat{l}$ are the estimated reward and risk functions respectively. They are trained with MSE loss.
The control parameter $\lambda$ adjusts the policy's decision rule over candidate actions, producing a candidate family $\{\pi_\lambda:\lambda\in\Lambda\}$. 
Unlike the breast cancer task, the induced risk curve over $\lambda$ can be non-monotone. 
Therefore, in addition to comparing different risk evaluators, we use this task to test the proposed Gaussian multiplier bootstrap correction. 
The bootstrap estimates the maximum risk-estimation error over $\Lambda$ and adds a uniform inflation term to all candidate risk estimates before selection.

\subsubsection{Safe Question Answering with SafetyBench.}
\label{app: daatset_safe}
We evaluate our method on SafetyBench, a benchmark
containing 11,435 multiple-choice questions covering seven categories
of safety concerns. We use its English version and formulate the task
as selective single-answer question answering: the system either
returns one answer option or abstains. Our objective is to answer as
many questions correctly as possible while keeping the error rate
among answered questions below a target level $\alpha$.
We use Qwen2.5-1.5B-Instruct as the base language model.

\textbf{Gate.}
For each checkpoint $t$, we independently train a gate
$g_t(x,a)\in[0,1]$ to predict whether a proposed answer $a$ to question
$x$ is correct. The gate is an $\ell_2$-regularized logistic classifier
with 47 input features: 15 features describing model confidence,
answer-choice statistics, and text lengths, together with 16 principal
components of the prompt representation and 16 principal components
of the prompt-with-answer representation.
All representations are extracted from the current checkpoint.
The PCA transformations, feature standardization, and classifier are
fitted separately at each checkpoint using 2,000 gate-training
questions. For each question, one answer is sampled from the current
model, and its correctness provides the supervision.
A separate set of 512 development questions defines 41 candidate
thresholds: 39 gate-score quantiles and the two endpoints
$-\infty$ and $+\infty$. Development labels are not used to select
these thresholds.

\textbf{Policy.}
Let $\mathcal A(x)$ denote the valid answer options for question $x$.
The model induces an answer distribution
\[
q_t(a\mid x)
=
\frac{\exp\{z_t(x,a)/T_{\mathrm{temp}}\}}
{\sum_{a'\in\mathcal A(x)}
 \exp\{z_t(x,a')/T_{\mathrm{temp}}\}},
\qquad T_{\mathrm{temp}}=0.8,
\]
where $z_t(x,a)$ is the logit of the corresponding answer-option token.
Given a threshold $\beta$, we first sample a proposed answer from
$q_t(\cdot\mid x)$ and return it only when $g_t(x, a)\geq\beta$.
Otherwise, the system abstains. Thus, the controlled policy is
\[
\pi_{t,\beta}(a\mid x)
=
q_t(a\mid x)\mathbf{1}\{g_t(x,a)\geq\beta\},
\qquad a\in\mathcal A(x),
\]
\[
\pi_{t,\beta}(\bot\mid x)
=
1-\sum_{a\in\mathcal A(x)}\pi_{t,\beta}(a\mid x),
\]
where $\bot$ denotes abstention. The control method selects $\beta$
at each round.

\begin{figure}[t]
    \centering
    \begin{tabular}{@{}c@{}}
        \includegraphics[width=\columnwidth]{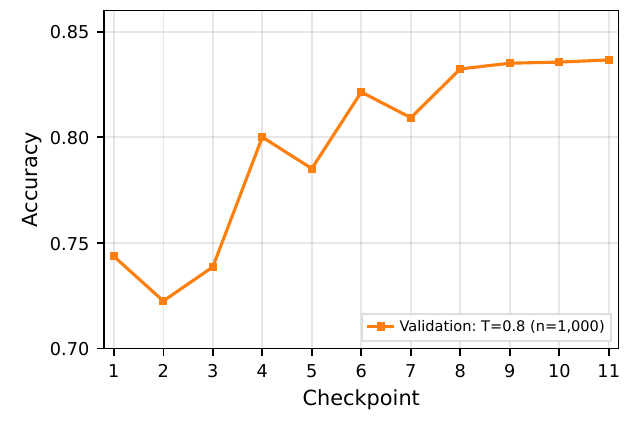}
    \end{tabular}
    \caption{Validation accuracy at $T=0.8$ on 1,000 SafetyBench questions. Qwen2.5-1.5B-Instruct is fine-tuned with LoRA (rank 64) on 4,409 questions for five epochs. Checkpoints 1--11 correspond to optimizer steps $0,69,\ldots,690$, respectively, with checkpoint 1 denoting the pretrained model. We use the checkpoint sequence as the policy sequence.}
    \label{fig:safetybench-training}
\end{figure}

\textbf{Policy sequence.}
After removing duplicate or ambiguously labeled records, we retain
11,409 questions and construct internal splits of 4,409 training
questions, 1,000 validation questions, and 6,000 questions reserved
for gate fitting and risk-control evaluation.
We fine-tune the base model using LoRA with rank 64 and scaling
parameter 128 on the attention and MLP projections.
Training uses supervised cross-entropy over valid answer options,
an effective batch size of 32, and AdamW with a peak learning rate
of $2\times10^{-5}$ followed by cosine decay.
Five training epochs produce 690 optimizer updates.
We save a checkpoint every 69 updates, obtaining eleven models at
steps $0,69,\ldots,690$, including the initial model.
These checkpoints, paired with their respective gates, define the
sequence of candidate policy families.
The experiment replays this fixed training trajectory:
Calibration feedback does not update the language model. Figure \ref{fig:safetybench-training} reports each checkpoint's accuracy during full-data fine-tuning. We then use each checkpoint as the policy for online calibration. 

\textbf{Metrics.}
For calibration repetition $b$ and checkpoint $t$, let $C_{bt}$
denote the probability of answering and $E_{bt}$ the probability of
answering incorrectly under the selected policy.
Its conditional risk is $R_{bt}=E_{bt}/C_{bt}$ when $C_{bt}>0$.
Across $B=200$ repetitions, we report the mean refusal rate and
pooled conditional risk:
\[
\mathrm{Refusal}_t
=
1-\frac{1}{B}\sum_{b=1}^{B}C_{bt},
\qquad
\mathrm{Risk}_t
=
\frac{\sum_{b=1}^{B}E_{bt}}
     {\sum_{b=1}^{B}C_{bt}}.
\]
We also report cumulative expected wrong answers,
\[
\mathrm{CumWrong}_t
=
\frac{M}{B}
\sum_{b=1}^{B}\sum_{s=0}^{t} E_{bs},
\qquad M=2464.
\]
and the risk satisfaction rate,
\[
\mathrm{Satisfaction}_t
=
\frac{1}{B}\sum_{b=1}^{B}
\mathbf{1}\{E_{bt}\leq\alpha C_{bt}\}.
\]
A policy that always abstains has undefined conditional risk;
it is counted as nonviolating in the satisfaction rate, and
its occurrence is reported separately.
Cumulative reward includes all evaluated checkpoints up to $t$
and excludes the calibration warmup.

\textbf{Procedure.}
The 6,000 reserved questions are partitioned into 2,000 gate-training
questions, 512 gate-development questions, a calibration pool of
2,464 questions, and a separate audit pool of 1,024 questions.
Each calibration trajectory starts with $m_0=512$ labeled
observations sampled with replacement from the calibration pool
using the initial answer policy $q_0$.
At checkpoint $t$, the current gate rescores all historical
question--answer pairs while preserving their original answers,
error labels, and logging probabilities.
The applicable controllers also receive an unlabeled probe sampled
from $q_t$.
After threshold selection, the probe label is revealed, and the
record is appended together with 128 additional labeled observations
from $q_t$. Consequently, the history available at checkpoint $t$
contains $m_0+129t$ records.
These observations are collected from the base answer policy,
irrespective of the selected abstention threshold.

We set $\alpha=0.1$ and use 2,000 Gaussian multiplier bootstrap draws
with $\delta=0.2$ for the high-probability variant, corresponding to
a nominal confidence level of $1-\delta=0.8$.
The methods share calibration histories and candidate threshold
grids within each repetition.
We also include a historical LTT--Bonferroni baseline, which tests
the 40 nontrivial candidates using one-sided exact binomial bounds
at level $\delta/40$.
This baseline ignores policy shift; its usual i.i.d.\ guarantee
is not assumed to hold for histories generated by changing policies.

We perform 200 calibration repetitions with the model checkpoints
and gates held fixed.
After recording the threshold selections, we compute the main
metrics by exactly integrating over all valid answer options and
the uniform distribution on the 2,464 question calibration pool.
Thus, the main evaluation concerns this fixed finite reference
distribution; the separate audit pool provides a secondary evaluation.
Risk satisfaction is measured separately at each checkpoint and
does not imply a simultaneous guarantee over the entire trajectory.

\subsubsection{Factual Medical Question Answering}
\label{app: daatset_medical}

We use the MedLFQA dataset. 
Each record contains a medical question, a generated response,
atomic claims extracted from the response, and binary labels
indicating whether the reference answer supports each claim.
Following the original preprocessing, we obtain 4,805 records
containing 81,196 atomic claims and use the cached log-probability
scores for claim filtering.
We repeat the experiment over 1,000 random record-level splits,
each containing 3,363 calibration records and 1,442 held-out test
records. Original records, including repeated question texts,
are retained.

\textbf{Control:}
The policy is a threshold-based filter: a claim is retained if
its score is at least $\lambda$.
For each response, the loss is the false discovery proportion
(FDP), defined as the fraction of retained claims that are
unsupported, with loss zero when no claims are retained.
Risk is the expected response-level FDP, while utility is the
recall of supported claims.
This is a static post-processing task without feedback-induced
covariate shift, so all importance weights equal one.
For each split, we construct approximately 200 candidate thresholds
from the calibration scores using the original implementation.
Our method uses Gaussian bootstrap calibration with nominal
failure probability $\delta=0.2$ and 2,000 joint Gaussian draws
to estimate a simultaneous risk upper bound over the candidates.
We select the smallest threshold whose upper bound does not exceed
$\alpha$, with an all-reject policy as a fallback.
We report
mean held-out FDP, supported-claim recall, and held-out risk
satisfaction rate, defined as the fraction of splits for which
the mean test FDP is at most $\alpha$.

\textbf{Baselines.}
We compare against gCRC, Monotonized CRC, and Learn Then Test (LTT).
gCRC scans thresholds from high to low and returns the preceding
threshold when the CRC-corrected empirical risk first exceeds
$\alpha$.
Monotonized CRC first replaces each response's loss by
$\widetilde L_i(\lambda)=\max_{\lambda'\geq\lambda}L_i(\lambda')$ over the
candidate grid, then applies standard CRC to these losses.
Both CRC methods use the correction
$(\sum_{i=1}^{n}L_i+1)/(n+1)$, with the appropriate original or
monotonized losses.
LTT follows the original implementation's fixed-sequence
Hoeffding--Bentkus tests, using $\delta_{\mathrm{LTT}}=\delta$.
All methods select thresholds using calibration data and are
evaluated using the original, unmodified FDP on held-out records.

\textbf{Summary:} Table \ref{tab:experiment_metrics} summarizes the control parameters, their effects on the distribution, and the risk metrics. 
\begin{table}[t]

\centering
\small
\renewcommand{\arraystretch}{1.2}
\caption{Control parameters and risk metrics.}
\label{tab:experiment_metrics}
\begin{tabularx}{\linewidth}{@{}lXX@{}}
\toprule
Experiment & Effect of $\lambda$ & Risk metric \\
\midrule
Breast cancer
& Probability of selecting malignancy
& False negative rate \\

Stock
& Probability of selecting each asset
& Probability of selecting the worst-performing asset \\

SafetyBench
& Deployment distribution over answers and abstention
& Error rate among answered questions \\

Medical QA
& Which claims are retained in fixed responses
& Expected false discovery proportion of retained claims \\
\bottomrule
\end{tabularx}
\end{table}

\begin{table}[t]
\label{tab:controlsetting}
\centering
\small
\renewcommand{\arraystretch}{1.2}
\caption{Distribution shifts and control settings.}
\label{tab:experiment_settings}
\begin{tabularx}{\linewidth}{@{}lXcc@{}}
\toprule
Experiment
& Shift type
& \shortstack{Monotonic\\risk function?}
& Control type \\
\midrule
Breast cancer
& Feedback covariate shift
& Yes
& Policy control \\

Stock
& Feedback covariate shift
& No
& Policy control \\

SafetyBench
& Non-adaptive policy-induced covariate shift
& No
& Policy control \\

Medical QA
& No distribution shift
& No
& Risk control \\
\bottomrule
\end{tabularx}
\end{table}

\subsection{Distribution Shift Type and Control Problem}
\label{app:comparisonofdefinition}
\textbf{Risk Control vs. Policy Control}
Our primary setting controls deployment risk through policy calibration: the parameter $\lambda$ modifies the action distribution of $\pi_\lambda$, thereby changing its expected deployment risk or the expected risk of the collected data. Policy control is thus both a mechanism for achieving risk control and a specific problem within the broader risk control framework. In the standard conformal risk control formulation \cite{angelopoulos2024conformal}, different from "policy" control, $\lambda$ instead parameterizes a prediction rule (e.g., classification threshold) and its associated risk $L_\lambda$. Although primarily motivated by policy control, our framework accommodates both policy-based and risk-based risk control. In settings without distribution shift, our high-probability control procedure can also be applied to standard risk estimators for high-probability guarantees. When distribution shift exists, our risk estimator can be applied to risk-based risk control alongside the high-probability control method, provided the shift can be quantified for reweighting. This includes both changes in the action distribution induced by policy updates and changes in the covariate distribution, as in the covariate-shift case in CRC. Our experiments include both policy calibration and direct calibration of parameterized loss functions.

\textbf{Feedback Covariate Shift vs. Non-Adaptive Covariate Shift} Our method is primarily motivated by feedback covariate shift,
where policies learned from past observations affect subsequent future data collection, inducing dependence across data observations. However, the permutation-weighting method is also applicable to covariate shift without feedback, including changes in the covariate distribution or data collected under multiple
non-adaptive policies, and performs well compared with baselines. Such settings do not require the
history-dependent policy updates that characterize feedback
covariate shift. Our framework therefore accommodates both
feedback and non-adaptive covariate shift, provided that the
relevant sampling distributions can be characterized to construct the weights. 

\subsection{Baselines}
\label{app:baseline}
\paragraph{Online CRC (O-CRC)}
Online CRC extends conformal risk control under distribution shift to the online policy-learning setting. 
At each round, it evaluates each candidate policy using a sample-wise importance-weighted risk estimate,
where the weight corrects the mismatch between the candidate policy and the historical data-collection policy. 
It then selects a candidate whose estimated risk is below the target threshold. 
This baseline corresponds to using local importance weights without the proposed permutation-based correction for feedback covariate shift.

\paragraph{RollingRC}RollingRC extends adaptive conformal inference to general online risk control by sequentially updating a calibration parameter according to the observed risk loss. 
Specifically, after observing the realized loss at each round, it updates the calibration parameter in the direction that would reduce future deviations from the target risk level. 

\paragraph{Conformal Policy Control (CPC)}
CPC uses a policy-control rule that orders candidate policies from safest to most aggressive and selects the most aggressive candidate whose conservatively adjusted weighted empirical risk remains below the target threshold. In our experiments, we use only this policy-selection rule as a baseline, without the safe-policy interpolation mechanism. 

\paragraph{Learn-then-Test (LTT)}
Learn-then-Test treats each candidate policy as a hypothesis and tests whether its risk is below the target threshold. 
It selects policies that pass a finite-sample risk test, typically using a multiple-testing correction to account for searching over many candidates. 
While LTT can provide high-probability guarantees under its assumptions, it is often conservative and designed for i.i.d. data.

\paragraph{Upper Confidence Bound Risk Control (UCB-RC).}
For each candidate $\lambda\in\Lambda$, we compute a weighted empirical risk estimate
\[
\widehat \ell_t(\lambda)
=
\sum_{i=1}^{t} w_i(\lambda)L_i,
\qquad 
\sum_{i=1}^{t}w_i(\lambda)=1.
\]
Assuming $L_i\in[0,1]$, we construct a Hoeffding-style upper confidence bound
\[
\mathrm{UCB}_t(\lambda)
=
\widehat \ell_t(\lambda)
+
\sqrt{
\frac{\log(|\Lambda|/\delta)}{2\,n_{\mathrm{eff}}(\lambda)}
},
\qquad
n_{\mathrm{eff}}(\lambda)
=
\frac{1}{\sum_{i=1}^{t} w_i(\lambda)^2}.
\]
The selected policy is restricted to the feasible set
\[
\widehat\Lambda_{\mathrm{UCB}}
=
\left\{
\lambda\in\Lambda:
\mathrm{UCB}_t(\lambda)\le \alpha
\right\}.
\]
We can then select the $\lambda$ with the highest estimated reward. 

\paragraph{Multi-step permutation weight}
\citet{prinster2024conformal} approximate the exact permutation weight by keeping only the dynamic factors from the most recent $d$ rounds. For each candidate point $Z_i$, its $d$-step weight is defined as the normalized total likelihood of all permutations that place $Z_i$ at the target position $t+1$:
\begin{align*}
p^{(d)}_{t+1}(Z_i \mid E_z^{(t)})
&=
\frac{
\sum_{\sigma:\,\sigma(n+t)=i}
\prod_{j=t+1-d}^{t+1}
p\!\left(
Z_{\sigma(j)}
\mid
Z_{\sigma(1)},\ldots,Z_{\sigma(j-1)}
\right)
}{
\sum_{\sigma}
\prod_{j=t+1-d}^{t+1}
p\!\left(
Z_{\sigma(j)}
\mid
Z_{\sigma(1)},\ldots,Z_{\sigma(j-1)}
\right)
}\\
& = \frac{
\sum_{\sigma:\,\sigma(n+t)=i}
\prod_{j=t+1-d}^{t+1}
\pi_{\theta,\lambda}^j\!\left(
Z_{\sigma(j)}
\mid
X_{\sigma(j)}
\right)
}{
\sum_{\sigma}
\prod_{j=t+1-d}^{t+1}
\pi_{\theta,\lambda}^j\!\left(
Z_{\sigma(j)}
\mid
X_{\sigma(j)}
\right)
}
\end{align*}
Here, the numerator sums the truncated likelihood over all orderings in which
$z_i$ is assigned to the final target timestep, while the denominator normalizes
over all possible orderings. Thus, $ p^{(d)}_{t+1}(z_i \mid E_z^{(t)})$
forms a probability distribution over the observed points.

\subsection{Reward of the risk control experiments} 

\begin{figure}[t]
    \centering
    \begin{subfigure}[b]{0.3\textwidth}
        \centering
        \includegraphics[width=\textwidth]{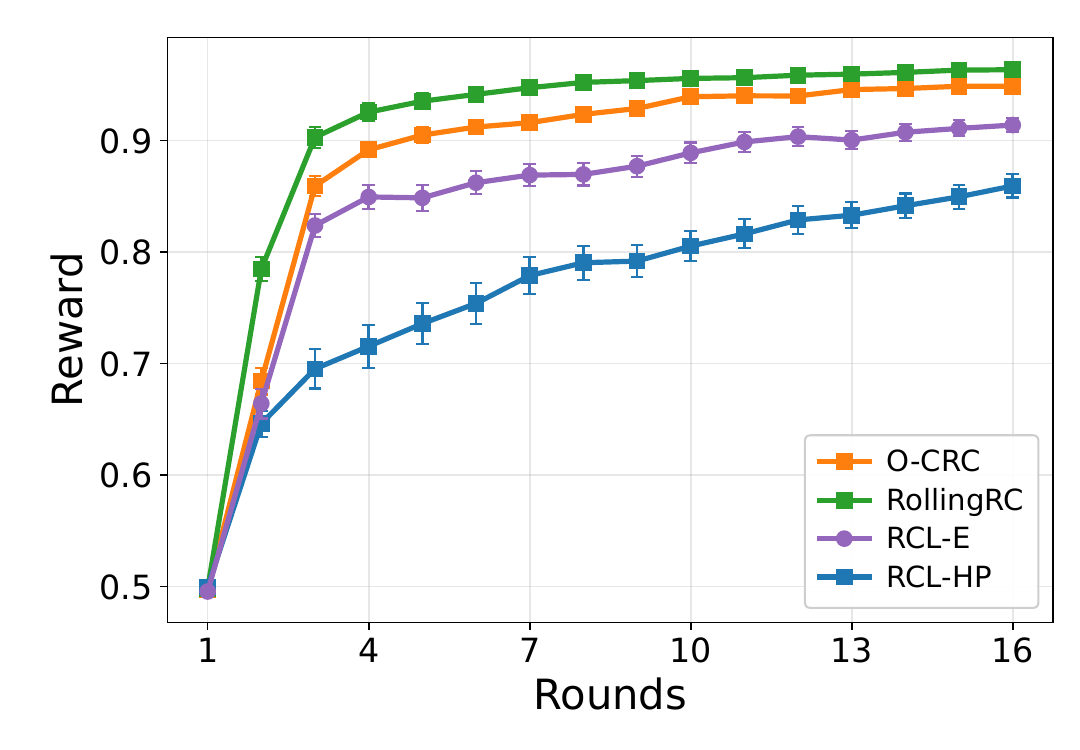}
        \caption{Breast Cancer}
    \end{subfigure}
    \begin{subfigure}[b]{0.3\textwidth}
        \centering
        \includegraphics[width=\textwidth]{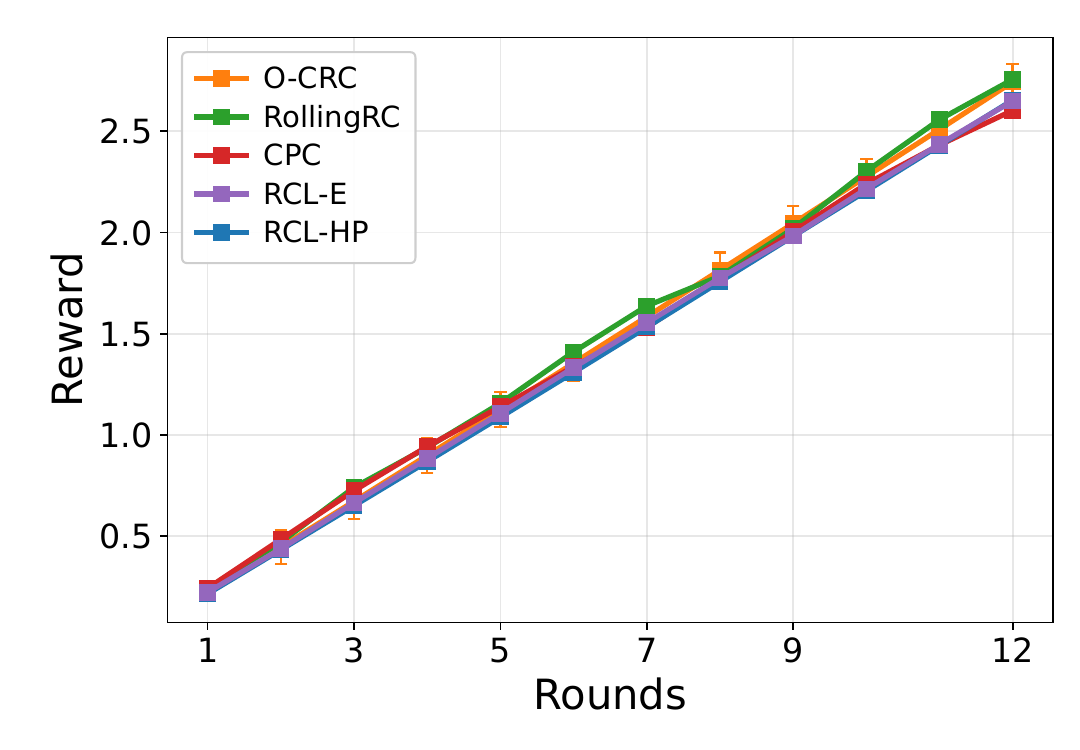}
        \caption{Stock Investment}
    \end{subfigure}
    \begin{subfigure}[b]{0.3\textwidth}
        \centering
        \includegraphics[width=\textwidth]{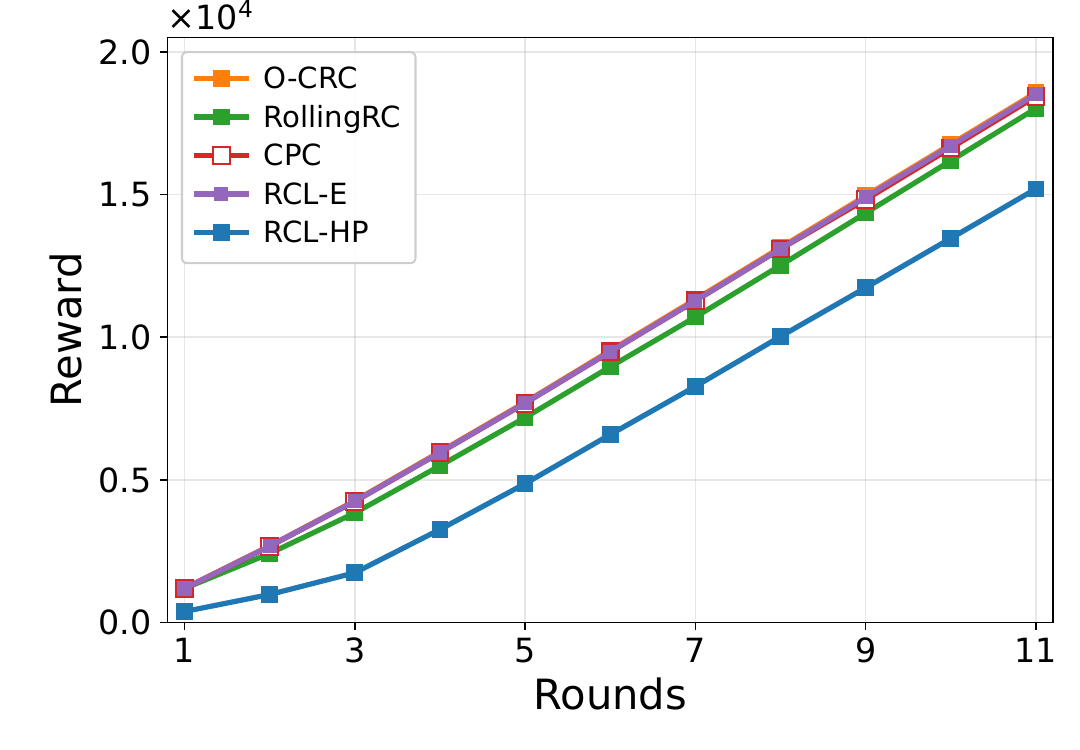}
        \caption{SafetyBench}
    \end{subfigure}
    \caption{Reward comparisons across the three experiments.}
    \label{fig:reward}
    \vspace{-10pt}
\end{figure}

In Figure \ref{fig:reward}, we show our method's reward compared with the baselines. We see that in Stock Investment, we receives a similar reward with lower risk. In the other two cases, our method receives a lower reward than the baselines because we satisfy the risk constraint.

\subsection{High-Probability Risk Control: Conservatism and Robustness to Distribution Shift}
\label{sec:highprobcomp}

Section~\ref{sec:medicalQA} compares RCL-HP with several high-probability risk-control
methods, including LTT. Here, we further analyze two aspects of
high-probability risk control: conservatism across nominal confidence
levels and robustness to distribution shift. We compare RCL-HP with
LTT and UCB-RC, where UCB-RC applies a concentration-based correction
to our risk estimates. Details of these baselines are provided in
Appendix~\ref{app:baseline}.

\begin{figure}[t]
    \centering
    \begin{subfigure}[b]{0.3\textwidth}
        \centering
        \includegraphics[width=\textwidth]{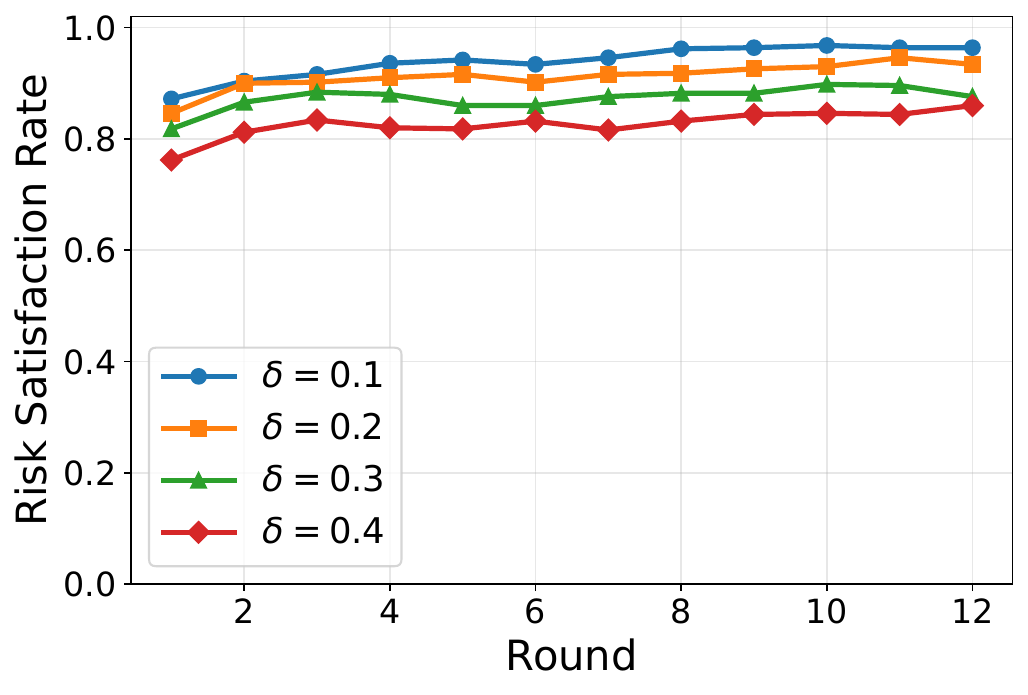}
        \caption{RCL-HP}
    \end{subfigure}
    \begin{subfigure}[b]{0.3\textwidth}
        \centering
        \includegraphics[width=\textwidth]{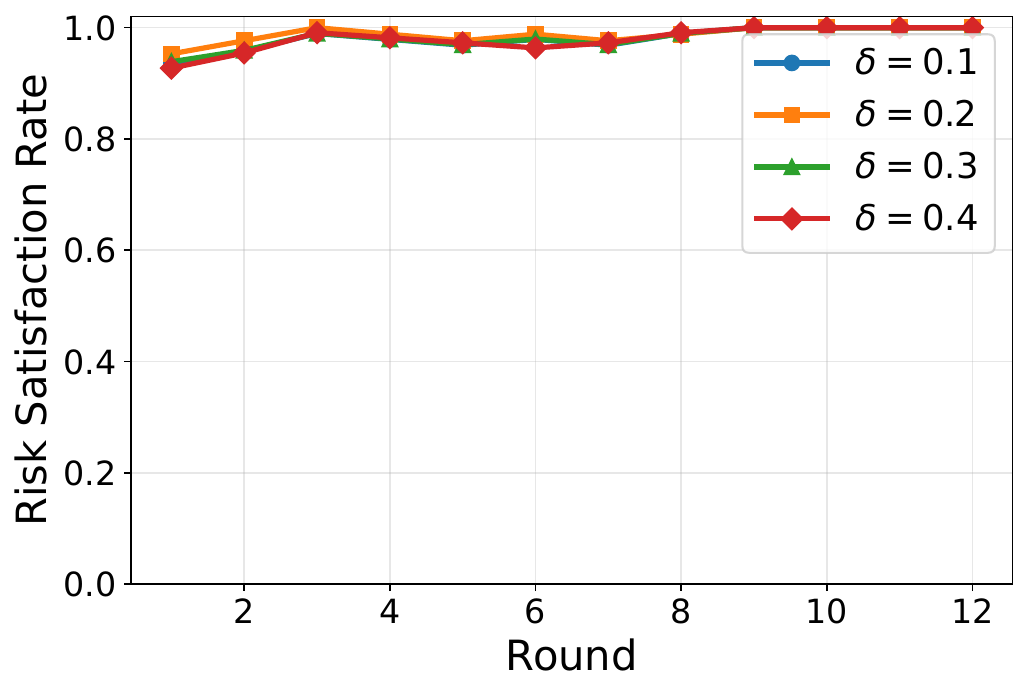}
        \caption{UCB-RC}
    \end{subfigure}
    \begin{subfigure}[b]{0.3\textwidth}
        \centering
        \includegraphics[width=\textwidth]{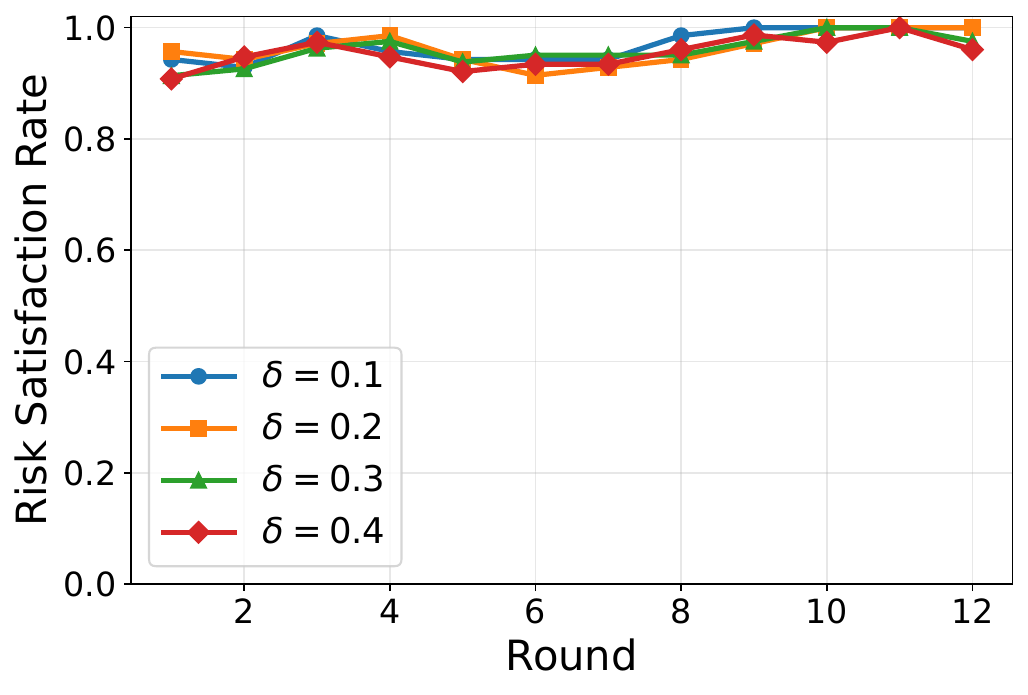}
        \caption{LTT}
    \end{subfigure}
    \caption{Risk control probability under different $1-\delta$  quantile/confidence. The baselines remain close to perfect risk satisfaction even for large $\delta$, indicating substantial conservativeness rather than adaptively adjusting the degree of conservativeness to the target high-probability risk control level.}
    \label{fig:risk_satisfaction}
    \vspace{-10pt}
\end{figure}

\textbf{Conservatism and sensitivity to $\delta$.}
Figure~\ref{fig:risk_satisfaction} compares empirical risk satisfaction
across different failure levels $\delta$.
UCB-RC and LTT retain very high satisfaction rates (close to $100\%$) across different $\delta$, including $\delta=0.4$, for which the
requirement is only $1-\delta=0.6$.
Although these rates satisfy the nominal requirement, they remain very conservative regardless of how we choose $\delta$. 
RCL-HP responds more strongly to the nominal level.

\begin{figure}[h]
    \centering
    \begin{subfigure}[b]{0.38\textwidth}
        \centering
        \includegraphics[width=\textwidth]{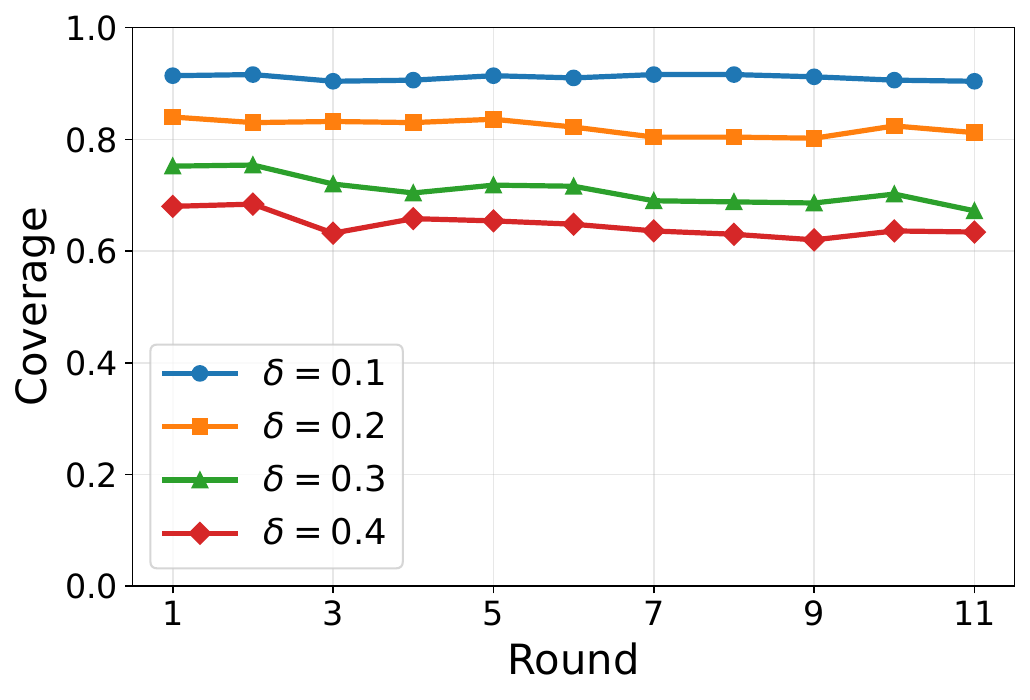}
        \caption{RCL-HP}
    \end{subfigure}
    \hfill
    \begin{subfigure}[b]{0.38\textwidth}
        \centering
        \includegraphics[width=\textwidth]{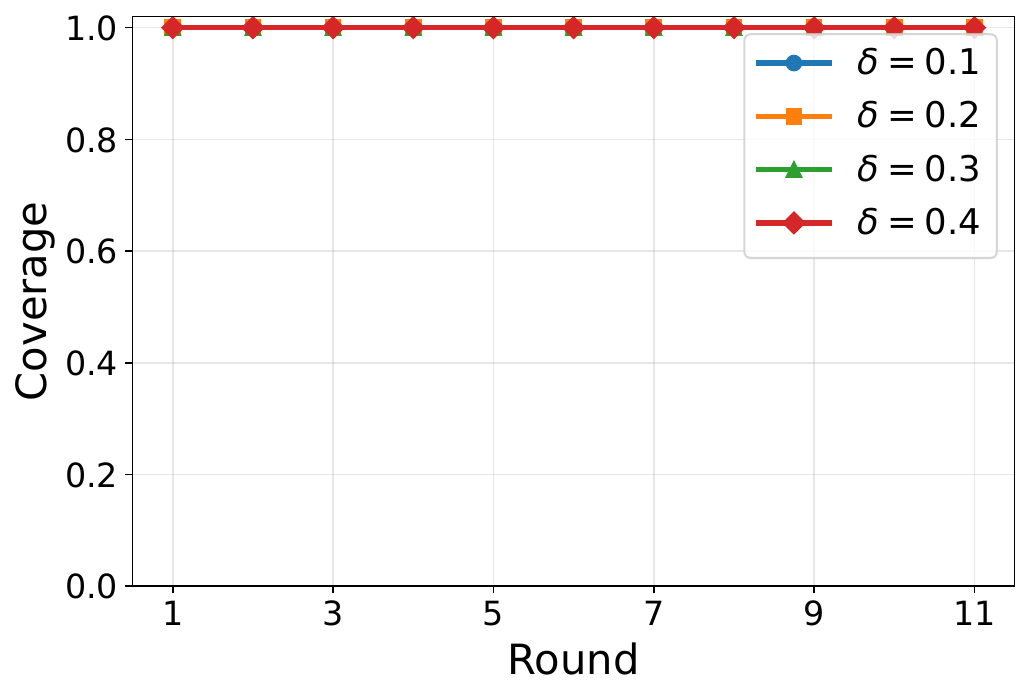}
        \caption{UCB-RC}
    \end{subfigure}
    \caption{
Empirical $P(\ell < \hat{\ell} + q_{1-\delta})$ under different $\delta$. The $P(\ell < \hat{\ell} + q_{1-\delta})$ of our method shows that the Gaussian bootstrap method achieves $1-\delta$ coverage, while for UCB $P(\ell < \hat{\ell} + q_{1-\delta}) = 1$ for all $\delta$.
These results suggest that the bootstrap correction provides a more adaptive uncertainty estimate for FCS data, while the UCB method can be overly conservative no matter how we choose $\delta$.
}
    \label{fig:coerage}
\end{figure}

To further examine the calibration of the uncertainty corrections,
Figure~\ref{fig:coerage} reports the empirical coverage of the risk
upper bounds.
UCB-RC maintains coverage close to one across the evaluated values
of $\delta$, indicating substantial overcoverage, particularly at
lower nominal confidence levels.
By comparison, the coverage of RCL-HP varies more closely with
$1-\delta$.
Together with the risk satisfaction results, this provides evidence
that the bootstrap correction is more responsive to the
confidence level in these experiments.

These differences are consistent with how the methods construct
their uncertainty corrections.
LTT combines individual risk tests with a family-wise testing
procedure, whose conservatism depends on the tests and the
multiple-testing rule employed.
For UCB-RC, the Hoeffding-style correction includes a multiplicity
term involving $\log(|\Lambda|/\delta)$ and depends on the effective
sample size
$
n_{\mathrm{eff}}(\lambda)
=
\frac{1}{\sum_i w_i(\lambda)^2},
$
where the weights are normalized to sum to one.
More concentrated weights reduce this effective sample size and
increase the correction uncertainty and upper bounds.
Together, weight concentration and the multiplicity adjustment can
produce conservative upper bounds.
RCL-HP instead calibrates the maximum fluctuation \emph{jointly} over the
candidate family using a Gaussian multiplier bootstrap, accounting
for dependence among the candidate-wise risk estimates.

\textbf{Robustness to distribution shift.}
Conservative uncertainty correction does not by itself ensure
robustness or risk satisfaction under distribution shift.
In particular, the original i.i.d. guarantee of LTT does not directly
apply to our settings.

\begin{figure*}[t]
\centering
\setlength{\tabcolsep}{1pt}
\begin{tabular}{cccc}
\includegraphics[width=0.48\textwidth]{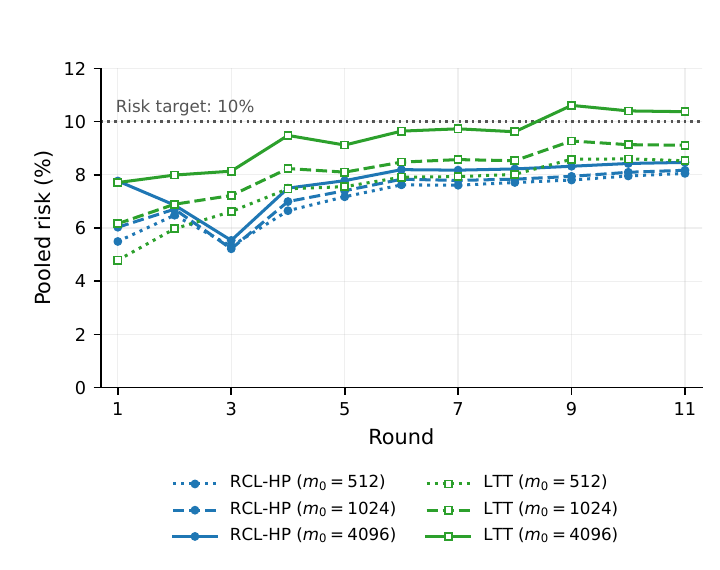} &
\includegraphics[width=0.48\textwidth]{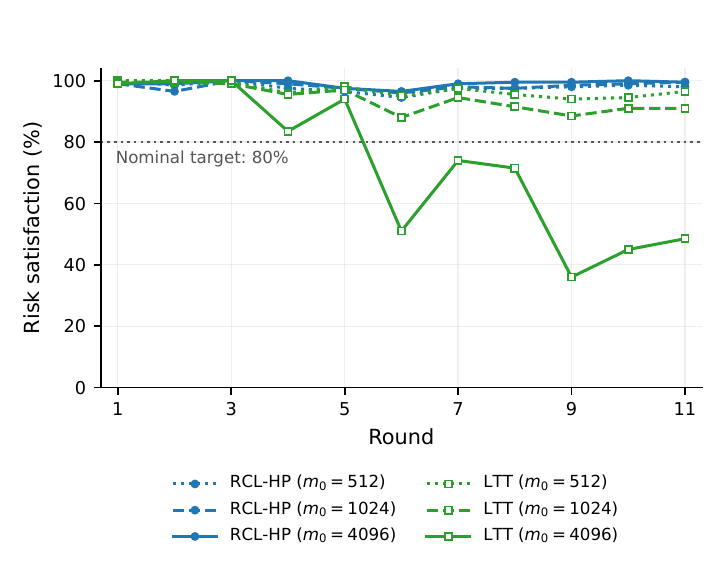}
\end{tabular}
\caption{SafetyBench results with warmup calibration sizes
$m_0\in\{512,1024,4096\}$.
LTT ignores policy shift.
The target risk is $\alpha=0.1$, with $\delta=0.2$ for HP calibration. Repeat 200 times. LTT may fail as it ignores the shift. Also, when the warmup size is larger, the failure is more significant even though it is more conservative.}
\label{fig:safetybench_ltt}
\end{figure*}

Figure~\ref{fig:safetybench_ltt} compares LTT and RCL-HP on SafetyBench
with warm-up sizes $m_0\in\{512,1024,4096\}$, where warm-up observations
are collected under the initial policy before online calibration
begins.
For $m_0=512$ and $1024$, LTT maintains the risk below the target
and empirical risk satisfaction above the nominal level throughout
the evaluated trajectory.
For $m_0=4096$, however, LTT exhibits violations in later rounds:
the risk exceeds the target, and empirical risk satisfaction
falls below the nominal level.
RCL-HP maintains the risk below the target and empirical
satisfaction above the nominal level at every evaluated round
across all three settings.

This pattern is consistent with a mismatch between the historical
calibration distribution and the current policy.
At a given online round, a larger $m_0$ increases the fraction of
calibration observations collected under the initial policy.
When subsequent policies induce different distributions, evaluating
current-policy risk using the historical observations, collected by the initial policy, without
adjusting for this mismatch can lead to unreliable risk estimates.
The results illustrate that the LTT baseline can fail to
maintain the nominal risk satisfaction level in this setting,
despite its conservative behavior.

\subsection{Experiments to Validate Assumptions and Lemmas in Theorem \ref{thm:main_boot}}
\label{app:additional-experimental-results}

\subsubsection{Approximation Error}
\label{app:additional-approximation}

\paragraph{Dataset and evaluation.}
We use the Wisconsin Diagnostic Breast Cancer dataset, containing
569 observations and 30 features. Malignant cases are treated as
the positive class, and the evaluated risk is the false-negative
rate (FNR). The sample sizes below refer to the observations used
to construct the permutation weights, rather than the entire dataset. We distinguish two sources of approximation error: fixing the learned policy sequence under permutation and
replacing exact permutation marginals with estimation from the Sinkhorn algorithm. Let $p^{\mathrm{ad}}$ denote the
exact adaptive weights, $p^{\mathrm{fr}}$ the exact weights for the frozen
policy sequence, and $q$ the computed Sinkhorn weights. We compare normalized
weight vectors using total variation,
$\operatorname{TV}(u,v)=\frac12\sum_i|u_i-v_i|$.

\paragraph{Exact calculation using fixed policy sequence versus Sinkhorn.}
We compare exact permutation marginals with Sinkhorn weights
computed from the same frozen policy matrix. The two checkpoints
contain 70 and 80 observations.
We compute exact weights using grouped permanent dynamic programming.

\begin{table}[htbp]
\centering
\small
\caption{Exact frozen-policy versus Sinkhorn weights. Relative gap: $|\mathrm{FNR}_{\mathrm{SH}}-\mathrm{FNR}_{\mathrm{exact}}|/\mathrm{FNR}_{\mathrm{SH}}$.}
\label{tab:sinkhorn-exact}
\begin{tabular}{@{}rrrrrr@{}}
\toprule
$N$ & TV & Exact FNR & Sinkhorn FNR
& Absolute gap & Relative gap \\
\midrule
70 & 0.001100 & 0.024359 & 0.026338 & 0.001979 & 7.5\% \\
80 & 0.001018 & 0.023472 & 0.025277 & 0.001805 & 7.1\% \\
\bottomrule
\end{tabular}
\label{tab: exact vs sinkhorn}
\end{table}

In Table \ref{tab: exact vs sinkhorn}, we see that the TV of the two weights is approximately $0.001$,
and the reported FNR differences are approximately
$0.20$ and $0.18$ percentage points with only $7\%$ gap. These results indicate
small relaxation errors for the evaluated frozen policy sequence. We believe that this gap will be even smaller with a larger dataset.

\paragraph{Adaptive exact weights versus Sinkhorn.}
We run experiments across five data collection rounds, with 10 data points per round. The exact adaptive weights account for policy updates under alternative observation orderings, rather than fixing the observed policy sequence. This comparison therefore includes both
policy-freezing and Sinkhorn-relaxation errors. We show the results in Table \ref{tab: exact weight vs sinkhorn}. The total variation is 0.00132. 

\begin{table}[htbp]
\centering
\small
\caption{Adaptive exact (retrain policy sequence for different permutations) versus Sinkhorn.}
\label{tab:adaptive-sinkhorn}
\begin{tabular}{@{}lr@{}}
\toprule
Comparison & TV \\
\midrule
Adaptive exact permutation weights versus Sinkhorn & 0.00132 \\
\bottomrule
\end{tabular}
\label{tab: exact weight vs sinkhorn}
\end{table}

\subsubsection{How approximation error affects the guarantee.}
\label{app:How approximation error affects the guarantee}
We first bound the risk-estimation error given the weight approximation error. For normalized weights
$u,v$ and losses $L_i\in[0,B]$,
\begin{equation}
\left|\sum_i u_iL_i-\sum_i v_iL_i\right|
\leq B\operatorname{TV}(u,v).
\label{eq:ae-risk-tv}
\end{equation}

\emph{Proof.}
Let $d_i=u_i-v_i$. Since both weight vectors sum to one,
$\sum_i d_i=0$, and hence
\[
\sum_{d_i>0}d_i
=
-\sum_{d_i<0}d_i
=
\frac12\sum_i|d_i|
=
\operatorname{TV}(u,v).
\]
Using $0\leq L_i\leq B$, we obtain
\[
-B\operatorname{TV}(u,v)
\leq \sum_i d_iL_i
\leq B\operatorname{TV}(u,v),
\]
which proves Eq.~\eqref{eq:ae-risk-tv}.

To connect this estimator bound to risk control, assume that the following
bounds hold simultaneously for all $\lambda\in\Lambda$:
\[
\operatorname{TV}(p^{\mathrm{ad}}_\lambda,q_\lambda)\leq\gamma,
\]
where $p^{\mathrm{ad}}_\lambda,q_\lambda$ are adaptive exact weight and the sinkhorn permutation weight and $\gamma\geq0$.

For RCL-E, suppose the 
exact adaptive weight provides a risk estimate $\widehat R_{\mathrm{ad}}$,  where $\mathrm{ad}$ stands for adaptive exact weight. Then, we can account for the weight-estimation error induced by the approximation weight by selecting parameters such that $\widehat R_{\mathrm{SH}} + B\gamma \leq \alpha$, preserving the original risk guarantee. 

For RCL-HP, we must also account for the error in the quantile estimation, i.e., $q_{1-\delta}^{\mathrm{SH}}$ and $q_{1-\delta}^{\mathrm{ad}}$. We now discuss the error.

Both quantiles are computed empirically using the same
Gaussian multiplier draws. Write
$p_i^{\mathrm{SH}}(\lambda):=q_{\lambda,i}$ and, for
$s\in\{\mathrm{ad},\mathrm{SH}\}$, define
$\phi_i^s(\lambda)
=p_i^s(\lambda)(L_i-\widehat R_s(\lambda))$.
Let $n=t+1$ and
$M_R=\max_{r\le R,\,i\le n}|\zeta_i^{(r)}|$, where $\zeta$ is the Gaussian multiplier.
On the event
$\sup_{\lambda\in\Lambda}
\operatorname{TV}(p_\lambda^{\mathrm{ad}},
p_\lambda^{\mathrm{SH}})\le\gamma$,
we have
\begin{equation}
\left|q_{1-\delta}^{\mathrm{ad}}
-q_{1-\delta}^{\mathrm{SH}}\right|
\le 3B\gamma M_R.
\label{eq:bootstrap-quantile-weight-error}
\end{equation}

\begin{proof}
The bounded-loss TV inequality gives
$|\widehat R_{\mathrm{ad}}-\widehat R_{\mathrm{SH}}|
\le B\gamma$.
Since
$\phi_i^{\mathrm{ad}}-\phi_i^{\mathrm{SH}}
=(p_i^{\mathrm{ad}}-p_i^{\mathrm{SH}})
(L_i-\widehat R_{\mathrm{ad}})
+p_i^{\mathrm{SH}}
(\widehat R_{\mathrm{SH}}-\widehat R_{\mathrm{ad}})$,
normalization and $L_i,\widehat R_{\mathrm{ad}}\in[0,B]$
imply
\[
\sum_i|\phi_i^{\mathrm{ad}}-\phi_i^{\mathrm{SH}}|
\le B\sum_i|p_i^{\mathrm{ad}}-p_i^{\mathrm{SH}}|
+|\widehat R_{\mathrm{SH}}-\widehat R_{\mathrm{ad}}|
\le 3B\gamma.
\]
For draw $r$, let
$T_s^{*(r)}
=\max_{\lambda\in\Lambda}\sum_i
\zeta_i^{(r)}\phi_i^s(\lambda)$.
Using the common multipliers,
\[
\begin{aligned}
|T_{\mathrm{ad}}^{*(r)}-T_{\mathrm{SH}}^{*(r)}|
&\le
\sup_{\lambda\in\Lambda}
\left|\sum_i\zeta_i^{(r)}
\bigl(\phi_i^{\mathrm{ad}}(\lambda)
-\phi_i^{\mathrm{SH}}(\lambda)\bigr)\right|\\
&\le 3B\gamma\max_i|\zeta_i^{(r)}|
\le 3B\gamma M_R.
\end{aligned}
\]
Sorting preserves this uniform bound: corresponding order
statistics of the two bootstrap samples differ by at most
$3B\gamma M_R$. 
\end{proof}

So we have:
\[
|q_{1-\delta}^{\text{SH}} - q_{1-\delta}^{\text{ad}}| \leq 3B\gamma M_R.
\]

Then, 
\begin{align*}
R(\lambda)
&\leq \widehat R_{\mathrm{ad}}(\lambda)
       +q_{1-\delta}^{\mathrm{ad}}\\
&\leq \widehat R_{\mathrm{SH}}(\lambda)
       + q_{1-\delta}^{\mathrm{SH}}+B\gamma (3M_R + 1) 
\qquad\text{for all }\lambda\in\Lambda.
\end{align*}
Thus, screening with
$\widehat R_{\mathrm{SH}}(\lambda)
       + q_{1-\delta}^{\mathrm{SH}}+B\gamma (3M_R + 1)\leq\alpha$
preserves the original risk threshold $\alpha$ whenever a feasible
candidate is selected.

\subsubsection{Assumption of the Log Range Not Increasing}
\label{app:additional-log-range}

For a positive weight vector or matrix $p$, define the log range of the row as:
$\Delta(p)=\max_i\log p_i-\min_i\log p_i$.
We test whether the assumption that the Sinkhorn weight row $Q$ has a smaller log range than the original policy matrix $J$ holds in practice, leading to smaller worst-case variance than IPS. 

Intuitively, the bounded log-range assumption means that after permutation, the weight is not too "sharp" and is smoother. We interpret this as minimizing KL while maximizing entropy. We directly evaluated the log-range of the permutation weights of the policy selected at every round.  $\Delta J$ and $\Delta Q$  are the log-range values before and after the Sinkhorn approximation. Table \ref{tab:satisfaction-rates} shows the satisfaction rate, i.e., $\Delta Q < \Delta J$ is $97\%$, indicating that the assumption holds in most cases. We also see that the average log range of the permutation weight is much smaller than that of the original policy. 

\begin{table}[t]
    \centering
    \caption{Satisfaction rates and the mean log range of the $Q$ and $J$.}
    \begin{tabular}{lccc}
        \toprule
        Dataset / setting & Satisfaction rate & Mean $\Delta J$ & Mean $\Delta Q$ \\
        \midrule
        Breast Cancer & 97.20\% & 4.595 & 2.854\\
        \bottomrule
    \end{tabular}
    \label{tab:satisfaction-rates}
\end{table}

\subsubsection{Empirical Evaluation of the Assumption in Theory}
\label{app: Empirical Evaluation of the Assumption in Theory}

We first evaluate whether the variance of the bootstrap estimator asymptotically converges to the variance of the true error using the breast cancer dataset. We also evaluate the Lemma derived directly from the assumption to show the validity of our assumptions. 

\paragraph{Variance Matching}
We evaluate variance matching for our Sinkhorn-based Gaussian
bootstrap using the breast cancer dataset.

\begin{figure}[t]
    \centering
    \includegraphics[width=0.5\textwidth]{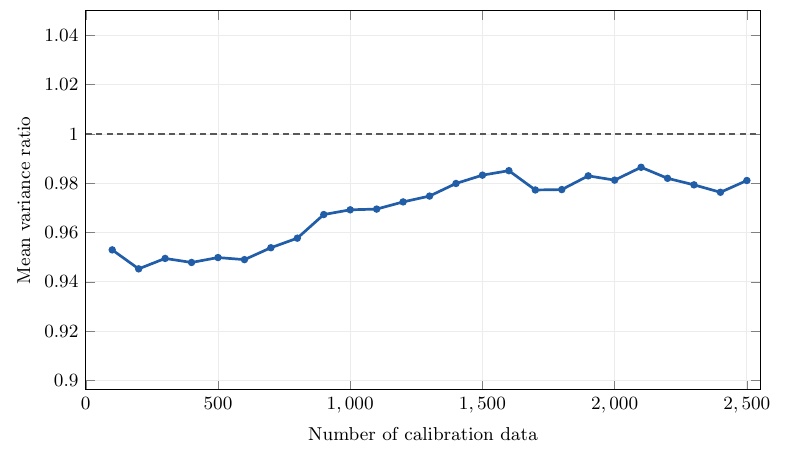}
    \caption{$\frac{\text{Var}_{\text{Estimated}}}{\text{Var}_{\text{GT}}}$. The ratio is close to 1 as the number of data increases.}
    \label{fig:variance-ratio}
\end{figure}

We sample 100 data points with replacement at each step until we have 2,500 data points for calibration and estimate the risk. We estimate the variance of the risk estimate with respect to the true risk by repeating the experiments 500 times. We also run 500 seeds, report the variance from bootstrapping each run, and compare it with the ground-truth variance. We report the results with the ratio $\frac{\text{Var}_{\text{Estimated}}}{\text{Var}_{\text{GT}}}$ in the Figure \ref{fig:variance-ratio}.

As the number of data points increases, the variance ratio approaches 1, indicating that the variances match. 

\paragraph{Verification of Linear Expansion in Lemma \ref{lem:wboot-linearization}}

As stated in the Lemma \ref{lem:wboot-linearization}, we have:  
\begin{equation}
 \widehat\ell_{t+1}(\lambda_j)-\ell_{t+1}(\lambda_j)
 =
 \frac{1}{t+1}\sum_{i=1}^{t}\psi_i(\lambda_j)
 +r_{t+1}(\lambda_j),
\end{equation}
where, for all sufficiently large $t$,
\[
 (t+1)\,
 \mathbb E\!\left[
  \sum_{j=1}^{K}r_{t+1}(\lambda_j)^2
 \right]
 \le C\bigl(\Delta_t+t^{-1}\bigr)
 \longrightarrow0.
\]
We now empirically evaluate $r_{t+1}(\lambda_j)$ and show that it goes to 0 as $t$ increases. We average the last three checkpoints as the reference policy $b(a\mid x)$ or $b_\infty(a\mid x)$. We compute the remainder as follows:
\begin{align}
g_j(x,a)
&=
\frac{q_j(a\mid x)}{b_\infty(a\mid x)}
\bigl(L(x,a)-\ell_j\bigr), \notag\\
d_{i,j}
&=
\frac{1}{H}\sum_{h=1}^{H}\sum_{a\in\{0,1\}}
b_i(a\mid x_h)g_j(x_h,a), \notag\\
\psi_{i,j}
&=
g_j(X_i,A_i)-d_{i,j},
\qquad i\le t,
\qquad \psi_{n,j}=0.
\label{eq:empirical-sh-contribution}
\end{align}
The conditional mean $d_{i,j}$ is evaluated exactly over the
$H=200$ contexts and two possible actions.
For each independent history $m$, the remainder is
\begin{equation}
r_{n,j}^{(m)}
=
\widehat\ell_{n,j}^{(m)}-\ell_j
-\frac{1}{n}\sum_{i=1}^{t}\psi_{i,j}^{(m)},
\qquad n=t+1.
\label{eq:empirical-sh-residual}
\end{equation}
We then estimate $n\mathbb E\|\boldsymbol r_n\|_2^2$ by
\begin{equation}
\widehat D_{\mathrm{rem}}(n)
=
\frac{n}{M}\sum_{m=1}^{M}
\|\boldsymbol r_n^{(m)}\|_2^2,
\label{eq:empirical-sh-remainder-statistic}
\end{equation}
using $M=2{,}500$ independent histories.

\begin{figure*}[t]
    \centering

    \begin{subfigure}[t]{0.32\textwidth}
        \centering
        \includegraphics[width=\linewidth]
        {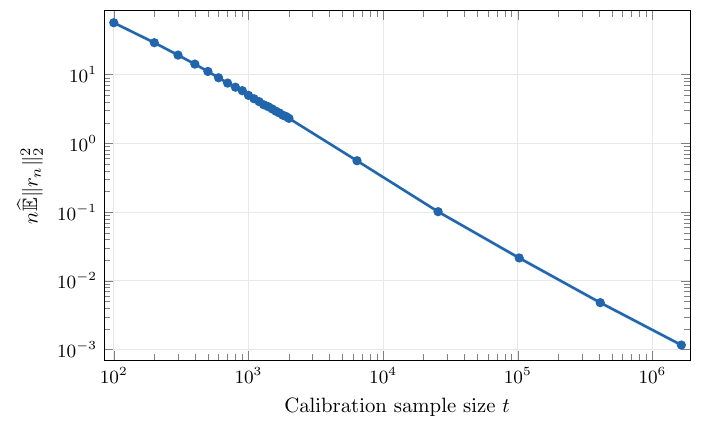}
        \caption{First-order remainder.}
        \label{fig:empirical-sh-remainder}
    \end{subfigure}
    \hfill
    \begin{subfigure}[t]{0.32\textwidth}
        \centering
        \includegraphics[width=\linewidth]
        {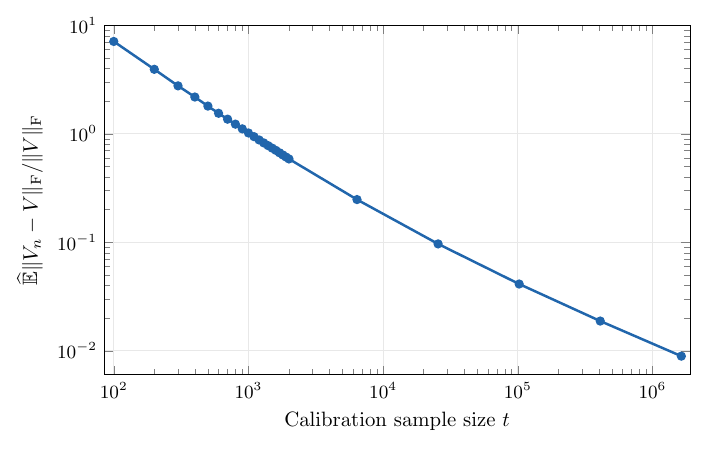}
        \caption{Covariance stability.}
        \label{fig:empirical-sh-covariance}
    \end{subfigure}
    \hfill
    \begin{subfigure}[t]{0.32\textwidth}
        \centering
        \includegraphics[width=\linewidth]
        {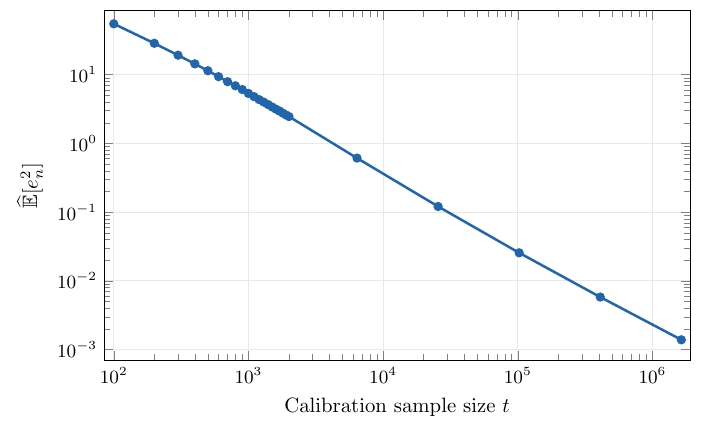}
        \caption{Score consistency.}
        \label{fig:empirical-sh-scores}
    \end{subfigure}

    \caption{
        Empirical diagnostics for the three conclusions of
        Lemma~\ref{lem:wboot-linearization}, \ref{lem:wboot-covariance},\ref{lem:wboot-score-consistency}.
        Results use the Sinkhorn estimator under Breast Cancer
        configuration B, with 200 fixed empirical contexts,
        three frozen candidate policies, and 2,500 independent
        histories.
    }
    \label{fig:empirical-sh-diagnostics}
\end{figure*}

Figure~\ref{fig:empirical-sh-remainder} shows that
$\widehat D_{\mathrm{rem}}(n)$ decreases as number of data increases. 
This sustained decay toward zero provides empirical support
for the $L^2$ remainder convergence in
Lemma~\ref{lem:wboot-linearization}, namely
$n\mathbb E\|\boldsymbol r_n\|_2^2\to0$.

\paragraph{Verification of the convergence of the covariance in Lemma \ref{lem:wboot-covariance}}

We now empirically evaluate the conditional covariance in Lemma \ref{lem:wboot-covariance}. As mentioned, define 
\[
V_{t+1}
 :=\frac{1}{t+1}\sum_{i=1}^{t}
   \mathbb E[\psi_i\psi_i^{\mathsf T}
             \mid\mathcal F_{i-1}],
\]
where $\psi_i$ is the linear expansion term defined in Lemma \ref{lem:wboot-linearization}.
And $V_{t+1}\xrightarrow{p}V$.
Here $\|\cdot\|$ denotes the operator norm and $C$
is independent of $t$.

We conduct experiments on breast cancer with three values of $\lambda$: $0, 6, 12$. We calculate $V_{t+1}$ as follows.
For each simulated history, let $H=200$ denote the size of
the fixed empirical context pool, and define
\begin{equation}
g_j(x,a)
=
\frac{q_j(a\mid x)}{b_\infty(a\mid x)}
\bigl(L(x,a)-\ell_j\bigr),
\qquad
\boldsymbol g(x,a)
=
\bigl(g_1(x,a),g_2(x,a),g_3(x,a)\bigr)^{\mathsf T},
\end{equation}
where $q_j$ is the $j$th frozen target policy,
$b_\infty$ is the fixed logging policy, which is the average of the last three checkpoint policies, and
$\ell_j$ is the true target risk over the empirical pool.

At observation $i$, the logging policy $b_i$ is determined
by the preceding history $\mathcal F_{i-1}$.
Since contexts are sampled uniformly from the pool, the
conditional mean of $\boldsymbol g(X_i,A_i)$ is
\begin{equation}
\boldsymbol d_i
=
\frac{1}{H}
\sum_{h=1}^{H}
\sum_{a\in\{0,1\}}
b_i(a\mid x_h)\boldsymbol g(x_h,a).
\end{equation}
The centered first-order contribution is therefore
$\boldsymbol\psi_i
=\boldsymbol g(X_i,A_i)-\boldsymbol d_i$, satisfying
$\mathbb E[\boldsymbol\psi_i\mid\mathcal F_{i-1}]=0$.
Its conditional covariance is
\begin{equation}
C_i
=
\frac{1}{H}
\sum_{h=1}^{H}
\sum_{a\in\{0,1\}}
b_i(a\mid x_h)
\bigl(\boldsymbol g(x_h,a)-\boldsymbol d_i\bigr)
\bigl(\boldsymbol g(x_h,a)-\boldsymbol d_i\bigr)^{\mathsf T}.
\end{equation}
Consequently,
\begin{equation}
\boxed{
V_{t+1}
=
\frac{1}{t+1}
\sum_{i=1}^{t}
\mathbb E\!\left[
\boldsymbol\psi_i\boldsymbol\psi_i^{\mathsf T}
\mid\mathcal F_{i-1}
\right]
=
\frac{1}{t+1}\sum_{i=1}^{t}C_i.
}
\end{equation}

We also calculate $V$ as follows.
Using the fixed limiting logging policy $b_\infty$, we obtain
\begin{equation}
V
=
\mathbb E_{b_\infty}\!\left[
\boldsymbol g(X,A)\boldsymbol g(X,A)^{\mathsf T}
\right]
=
\frac{1}{H}
\sum_{h=1}^{H}
\sum_{a\in\{0,1\}}
b_\infty(a\mid x_h)
\boldsymbol g(x_h,a)\boldsymbol g(x_h,a)^{\mathsf T},
\end{equation}
where $H=200$, $X$ is uniform over the fixed empirical
context pool, and $A\mid X\sim b_\infty(\cdot\mid X)$.
Since $\mathbb E_{b_\infty}[\boldsymbol g(X,A)]
=\boldsymbol 0$, this second moment equals the covariance
matrix of $\boldsymbol g(X,A)$ under the limiting policy.

The resulting $3\times3$ matrix $V$ is evaluated exactly
over all context-action pairs and serves as the same
deterministic reference for every sample size and
simulated history.

Figure~\ref{fig:empirical-sh-covariance}.
We see that the average relative discrepancy between
$V_{t+1}$ and $V$ decreases as number of data increases,
This provides empirical
support for the covariance stabilization
$V_{t+1}\xrightarrow{p}V$ in the second conclusion of
Lemma~\ref{lem:wboot-covariance}.

\paragraph{Verification of the Lemma \ref{lem:wboot-score-consistency}}

We further show the consistency of the bootstrap in Lemma \ref{lem:wboot-score-consistency}. 

We assess the consistency of the estimated first-order
contributions:
\begin{equation}
\boxed{
\frac{1}{t+1}\sum_{i=1}^{t+1}
\left\|
\widehat{\boldsymbol\psi}_{t+1,i}
-\boldsymbol\psi_i
\right\|_2^2
\xrightarrow{p}0.
}
\end{equation}

The estimated contribution for candidate $j$ is
\begin{equation}
\widehat\psi_{t+1,i,j}
=
(t+1)w_{t+1,i,j}
\bigl(L_i-\widehat\ell_{t+1,j}\bigr),
\qquad i=1,\ldots,t+1,
\end{equation}
where $w_{t+1,i,j}$ is the normalized Sinkhorn weight and
\begin{equation}
\widehat\ell_{t+1,j}
=
\sum_{i=1}^{t+1}w_{t+1,i,j}L_i.
\end{equation}

The true first-order contribution is defined using
\begin{align}
g_j(x,a)
&=
\frac{q_j(a\mid x)}{b_\infty(a\mid x)}
\bigl(L(x,a)-\ell_j\bigr),\\
d_{i,j}
&=
\frac{1}{H}
\sum_{h=1}^{H}
\sum_{a\in\{0,1\}}
b_i(a\mid x_h)g_j(x_h,a),\\
\psi_{i,j}
&=
g_j(X_i,A_i)-d_{i,j},
\qquad i=1,\ldots,t.
\end{align}
Here, $H=200$, $q_j$ is the frozen target policy,
$\ell_j$ is its exact risk over the fixed empirical pool,
$b_i$ is the actual logging policy at observation $i$,
and $b_\infty$ is the fixed limiting logging policy.
The centering term $d_{i,j}$ is evaluated exactly over
all context-action pairs, ensuring
$\mathbb E[\psi_{i,j}\mid\mathcal F_{i-1}]=0$.

For the test point, we set
$\boldsymbol\psi_{t+1}=\boldsymbol 0$.
Its estimated contribution is retained in the sum,
using the corresponding Sinkhorn weight and the
assigned worst-case loss $L_{t+1}=1$.

The experimental results are shown in
Figure~\ref{fig:empirical-sh-scores}.
We observe that the mean squared discrepancy between
the estimated and true first-order contributions decreases.
This provides empirical support for 
Lemma~\ref{lem:wboot-score-consistency}.

\subsection{Implementation Details}
We conduct experiments on one A5000 GPU for Breast Cancer and Stock Investment. One seed experiment takes ~2-5 minutes. We use RTX PRO 6000 for SafetyBench. Fine-tuning takes ~20 minutes, and the calibration process takes 1.5 hours. MedicalQA doesn't involve any GPU usage and takes ~3 minute.

\section*{Broader Impact}
Our work aims to ensure the safe deployment of online policy learning algorithms, which is crucial for reliable online and sequential decision-making. Providing principled risk-control guarantees during learning and deployment is essential to make online policy learning more trustworthy and practically usable. However, high-probability guarantees are asymptotic and depend on the stated assumptions and a sufficiently large dataset; they should not be interpreted as unconditional finite-sample safety guarantees. If these assumptions fail or estimation errors are underestimated, a deployed policy may exceed the target risk. 

\end{document}